\documentclass[journal,twoside,web]{ieeecolor}
\usepackage{tmi}
\usepackage{cite}
\usepackage{amsmath,amssymb,amsfonts}
\usepackage{graphicx}
\usepackage{textcomp}
\usepackage{hyperref}
\usepackage{pifont}
\usepackage{soul}
\usepackage{dsfont}
\usepackage{caption}
\usepackage{subcaption}
\usepackage{booktabs}
\usepackage{wrapfig}
\usepackage[normalem]{ulem}
\usepackage{algpseudocode}
\usepackage{algcompatible}
\usepackage{multirow}
\usepackage{booktabs}
\usepackage{makecell}
\usepackage{xcolor, colortbl}

\definecolor{LightCyan}{rgb}{0.88,1,1}
\definecolor{darkcyan}{rgb}{0,.79,.75}
\newcommand{\cmark}{\ding{51}}%
\newcommand{\xmark}{\ding{55}}%
\def\BibTeX{{\rm B\kern-.05em{\sc i\kern-.025em b}\kern-.08em
    T\kern-.1667em\lower.7ex\hbox{E}\kern-.125emX}}
\definecolor{darkred}{rgb}{0.6148, 0., 0.}
\definecolor{moredarkred}{rgb}{0.5, 0., 0.}
\definecolor{darkgreen}{rgb}{0., 0.6148, 0.}
\definecolor{darkblue}{rgb}{0., 0., 0.6148}
\definecolor{darkgold}{rgb}{0.9, 0.7, 0.}
\definecolor{darkmagenta}{rgb}{.5, 0, .5}

\definecolor{LightCyan}{rgb}{0.88,1,1}
\definecolor{DarkCyan}{rgb}{0,.79,.75}
\definecolor{LightGray}{rgb}{0.9, 0.9, 0.9}

\newcommand{\new}[1]{{\color{black}#1}}

\begin{document}
\title{Translation Consistent Semi-supervised Segmentation for 3D Medical Images}
\author{Yuyuan Liu, Yu Tian, Chong Wang, Yuanhong Chen, Fengbei Liu, Vasileios Belagiannis, Gustavo Carneiro
\thanks{This work was supported in part by funding from the Australian Research Council through grant FT190100525. (Corresponding author: Chong Wang and Gustavo Carneiro).}
\thanks{Yuyuan Liu and Yuanhong Chen are with the Australian Institute for Machine Learning, University of Adelaide, SA 5000, AU. 
(e-mail: {yuyuan.liu, yuanhong.chen}@adelaide.edu.au).}
\thanks{Yu Tian is with the Harvard Ophtalmology AI Lab of Schepens Eye Research Institute, Harvard Medical School, Boston 02114ß, US.
(e-mail: ytian11@meei.harvard.edu).}
\thanks{Chong Wang is with the College of Electronic and Information Engineering, Tongji University, Shanghai 201804, CN. 
(e-mail: chongwangsmu@gmail.com)}
\thanks{Fengbei Liu is with the School of Electrical and Computer Engineering, Cornell University and Cornell Tech, New York, US. (e-mail: fl453@cornell.edu)}
\thanks{Vasileios Belagiannis is with the the Faculty of Engineering of the Friedrich-Alexander-Universität Erlangen-Nürnberg, 91058 Erlangen, GER.
(e-mail: vasileios.belagiannis@fau.de).}
\thanks{Gustavo Carneiro is with the Centre for Vision, Speech and Signal Processing (CVSSP), University of Surrey, UK. (email: g.carneiro@surrey.ac.uk).}
}

\maketitle
\begin{abstract}
3D medical image segmentation methods have been successful, but their dependence on large amounts of voxel-level annotated data is a disadvantage that needs to be addressed given the high cost to obtain such annotation. 
Semi-supervised learning (SSL) solves this issue by training models with a large unlabelled and a small labelled dataset. The most successful SSL approaches are based on consistency learning that minimises the distance between model responses obtained from perturbed views of the unlabelled data. These perturbations usually keep the spatial input context between views fairly consistent, which may cause the model to learn segmentation patterns from the spatial input contexts instead of the foreground objects. 
In this paper, we introduce the \underline{Tra}nslation \underline{Co}nsistent \underline{Co}-training (TraCoCo) which is a consistency learning SSL method that perturbs the input data views by varying their spatial input context, allowing the model to learn segmentation patterns from foreground objects. Furthermore, we propose a new Confident Regional Cross entropy (CRC) loss, which improves training convergence and keeps the robustness to co-training pseudo-labelling mistakes.
Our method yields state-of-the-art (SOTA) results for several 3D data benchmarks, such as the Left Atrium (LA), Pancreas-CT (Pancreas), and Brain Tumor Segmentation (BraTS19). Our method also attains best results on a 2D-slice benchmark, namely the Automated Cardiac Diagnosis Challenge (ACDC), further demonstrating its effectiveness.
Our code, training logs and checkpoints are available at \url{https://github.com/yyliu01/TraCoCo}. 
\end{abstract}


\section{Introduction}
\label{sec:introduction}
\IEEEPARstart{T}{he} training of 3D medical image segmentation neural network methods requires large sets of voxel-wise annotated samples.  These sets are obtained using a laborious and expensive slice-by-slice annotation process, so alternative methods based on small labelled set training methods have been sought.
 One example is semi-supervised learning (SSL) that relies on a large unlabelled set and a small labelled set to train the model,
and a particularly effective SSL approach is based on the consistency learning that minimises the distance between model responses obtained from different views of the unlabelled data~\cite{ouali2020semi, chen2021-CPS}.

The different views of consistency learning methods can be obtained via data augmentation~\cite{berthelot2019mixmatch} or from the outputs of differently initialized networks~\cite{tarvainen2017mean,chen2021-CPS,ke2020guided}. 
Mean teacher (MT)~\cite{tarvainen2017mean,yu2019uncertainty,hang2020local,wang2021tripled,liu2021perturbed} combines these two perturbations and averages the network parameters during training, yielding reliable pseudo labels for the unlabelled data. 
Various schemes in 3D medical image segmentation are introduced to improve the generalisation of teacher-student methods, including uncertainty guided threshold~\cite{yu2019uncertainty, hang2020local} or multi-task assistance~\cite{wang2021tripled,luo2021dtc}. However, the domain-specific transfer~\cite{berthelot2019mixmatch} of the teacher-student scheme can cause both networks to converge to a similar local minimum, reducing the network perturbation effectiveness. Moreover, some hard medical segmentation cases are consistently segmented similarly by the teacher and student models, potentially resulting in confirmation bias during training.
This issue motivated the introduction of the co-training framework that involves two models that are initialized with different parameters and mutually supervise each other by generating pseudo-labels for the unlabelled data during the training phase. Those two independent models have less chance of converging to the same local minima than the teacher-student model.
Recent approaches~\cite{chen2021-CPS,ke2020guided} show that co-training
provides an effective consistency regularization with the cross-supervision between two independent networks. 

\begin{figure*}[t!]
    \centering
     \begin{subfigure}[b]{.192\textwidth}
     \includegraphics[width=1.\textwidth]{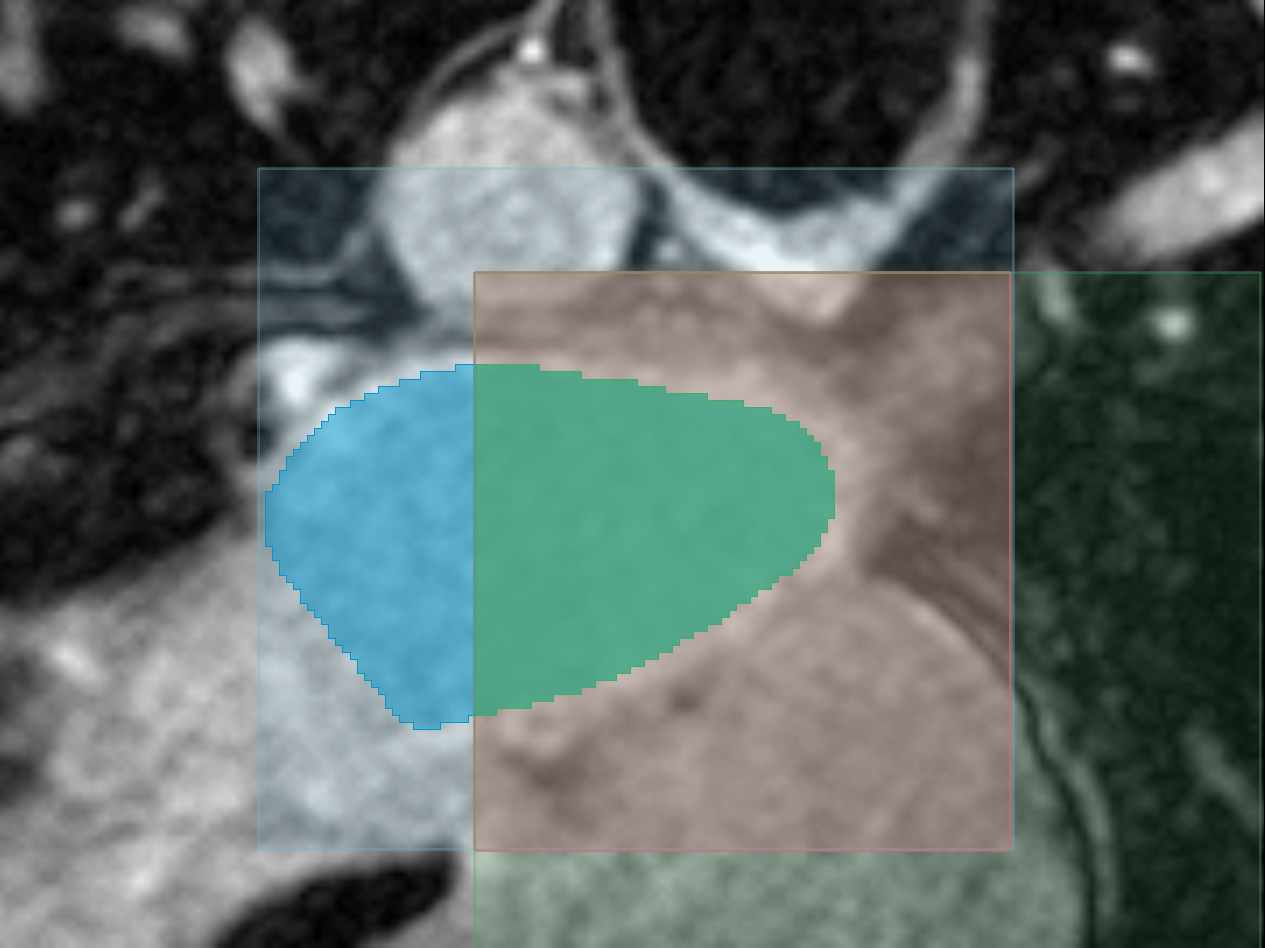}
     \includegraphics[width=1.\textwidth]{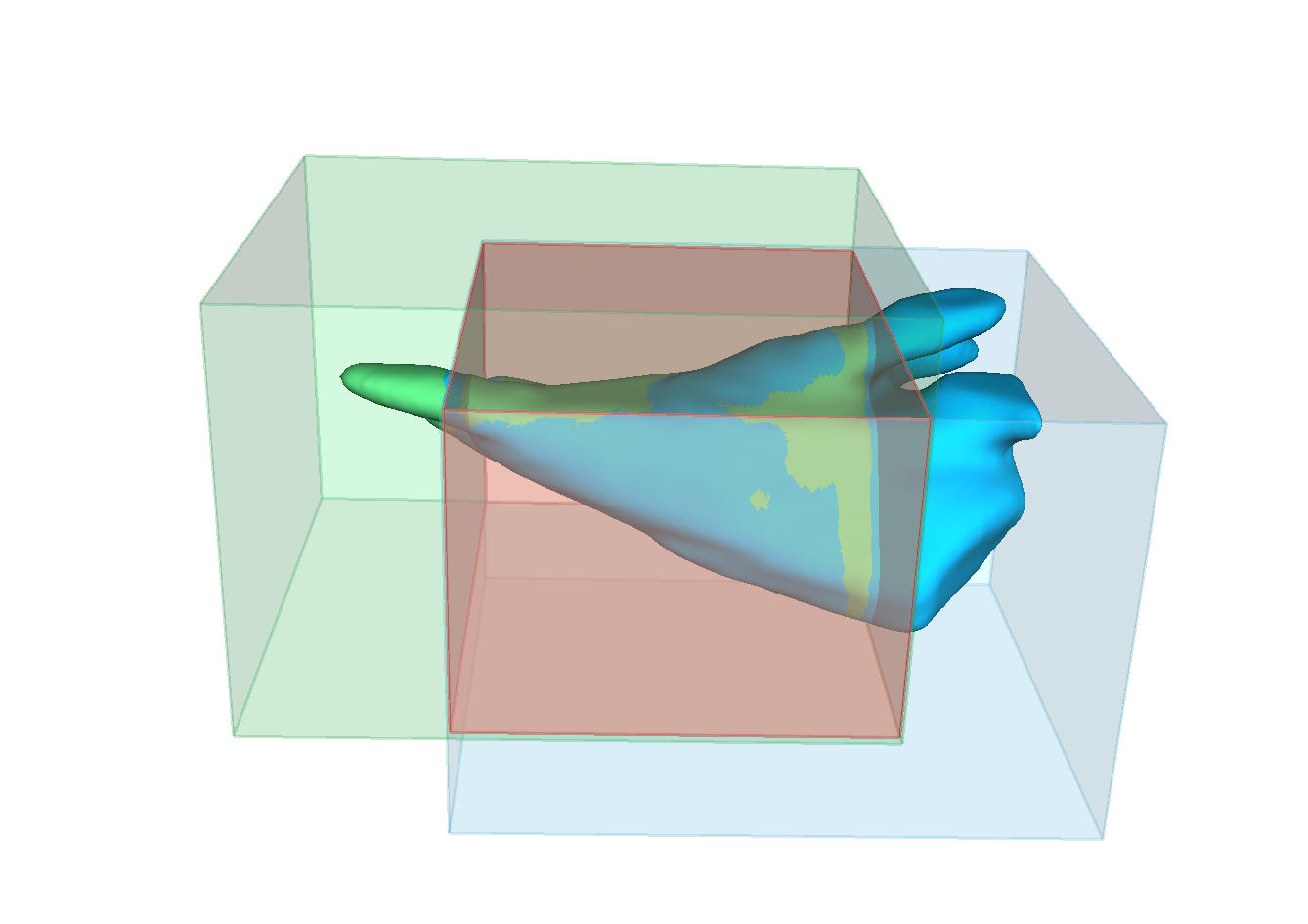}
     \caption*{Ground truth}
     \end{subfigure}
     \begin{subfigure}[b]{.192\textwidth}
     \includegraphics[width=1.\textwidth]{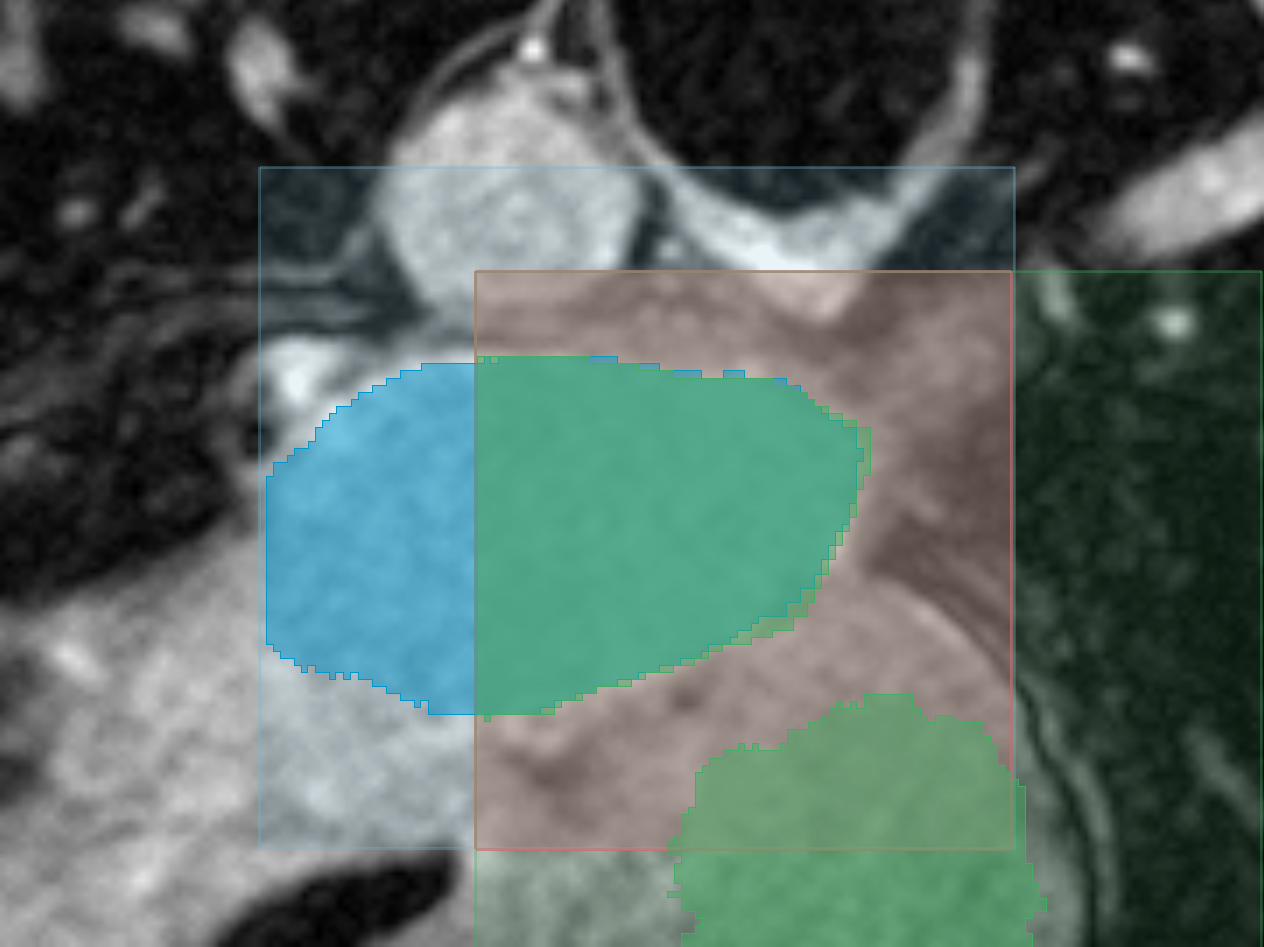}
     \includegraphics[width=1.\textwidth]{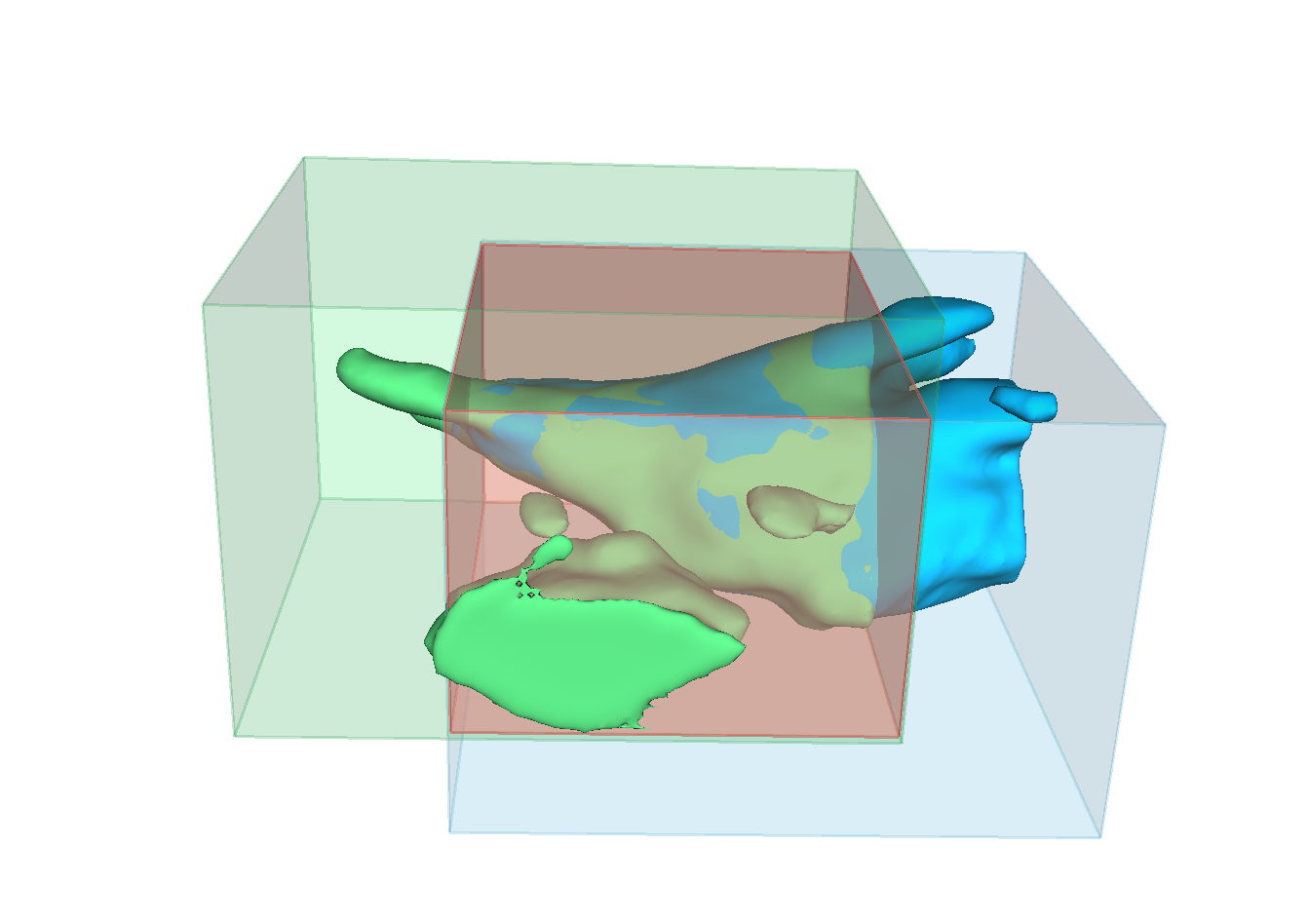}
     \caption*{SASSNet~\cite{li2020shape}}
     \end{subfigure}
     \begin{subfigure}[b]{.192\textwidth}
     \includegraphics[width=1.\textwidth]{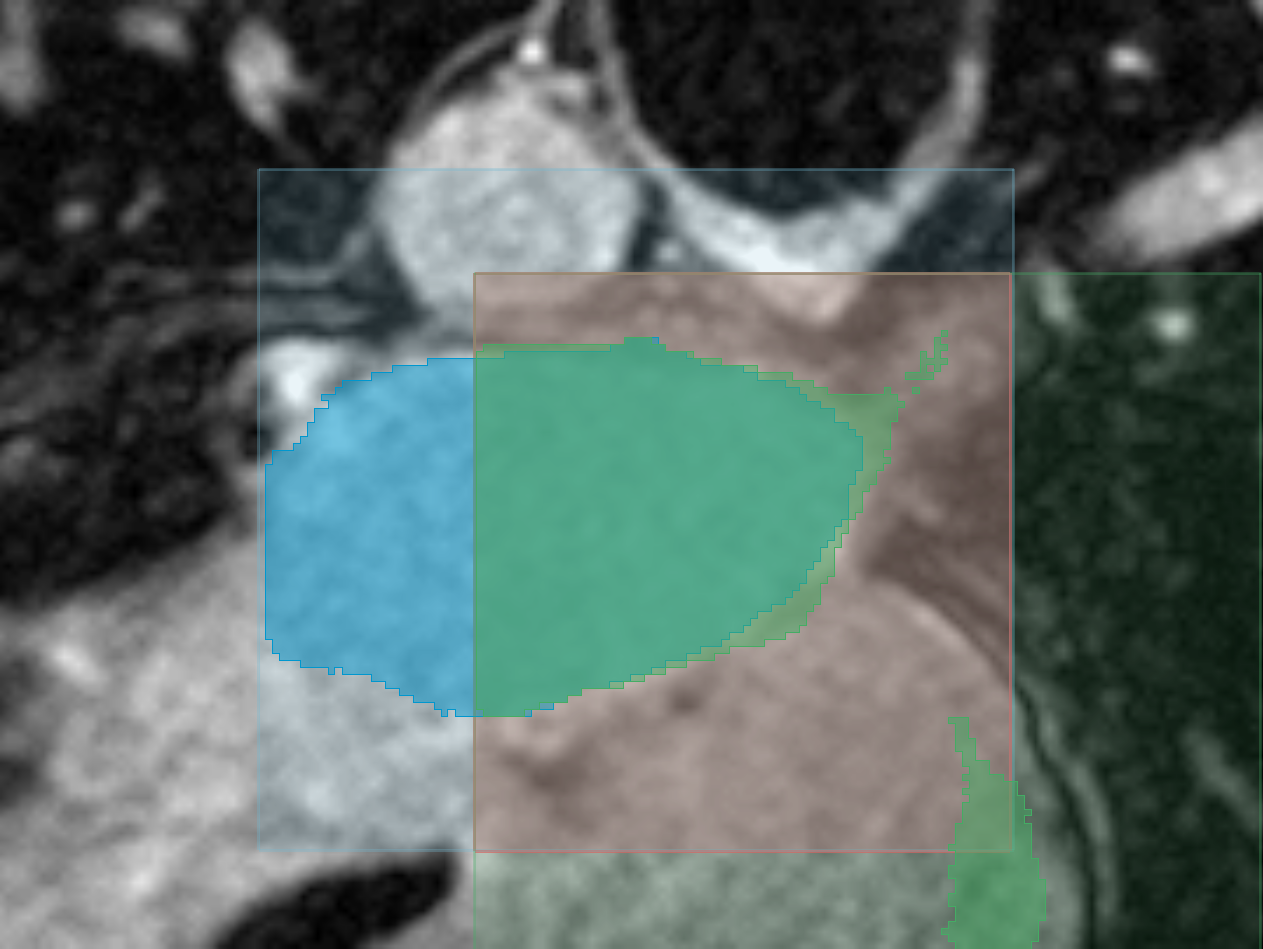}
     \includegraphics[width=1.\textwidth]{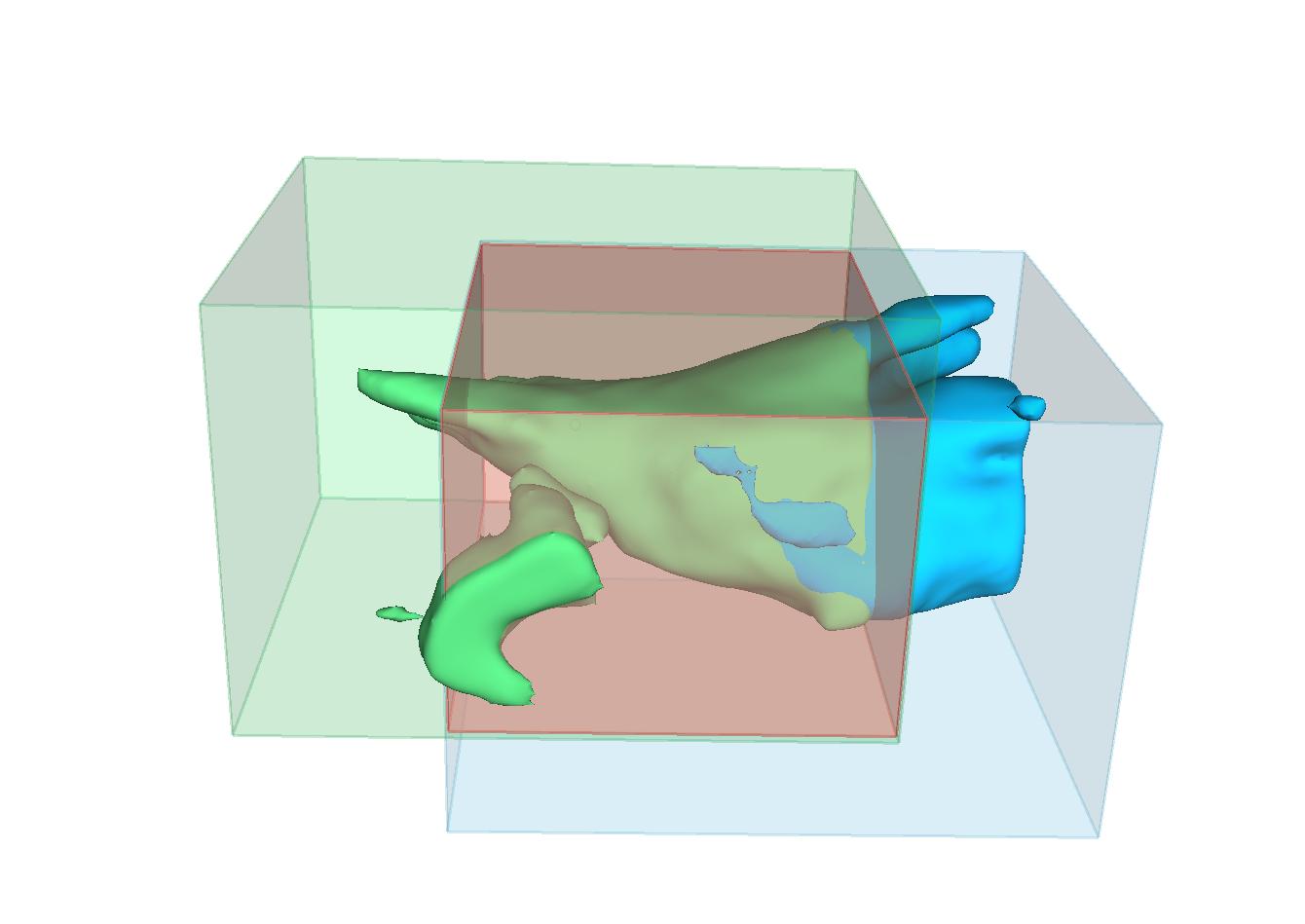}
     \caption*{UA-MT~\cite{yu2019uncertainty}}
     \end{subfigure}
     \begin{subfigure}[b]{.192\textwidth}
     \includegraphics[width=1.\textwidth]{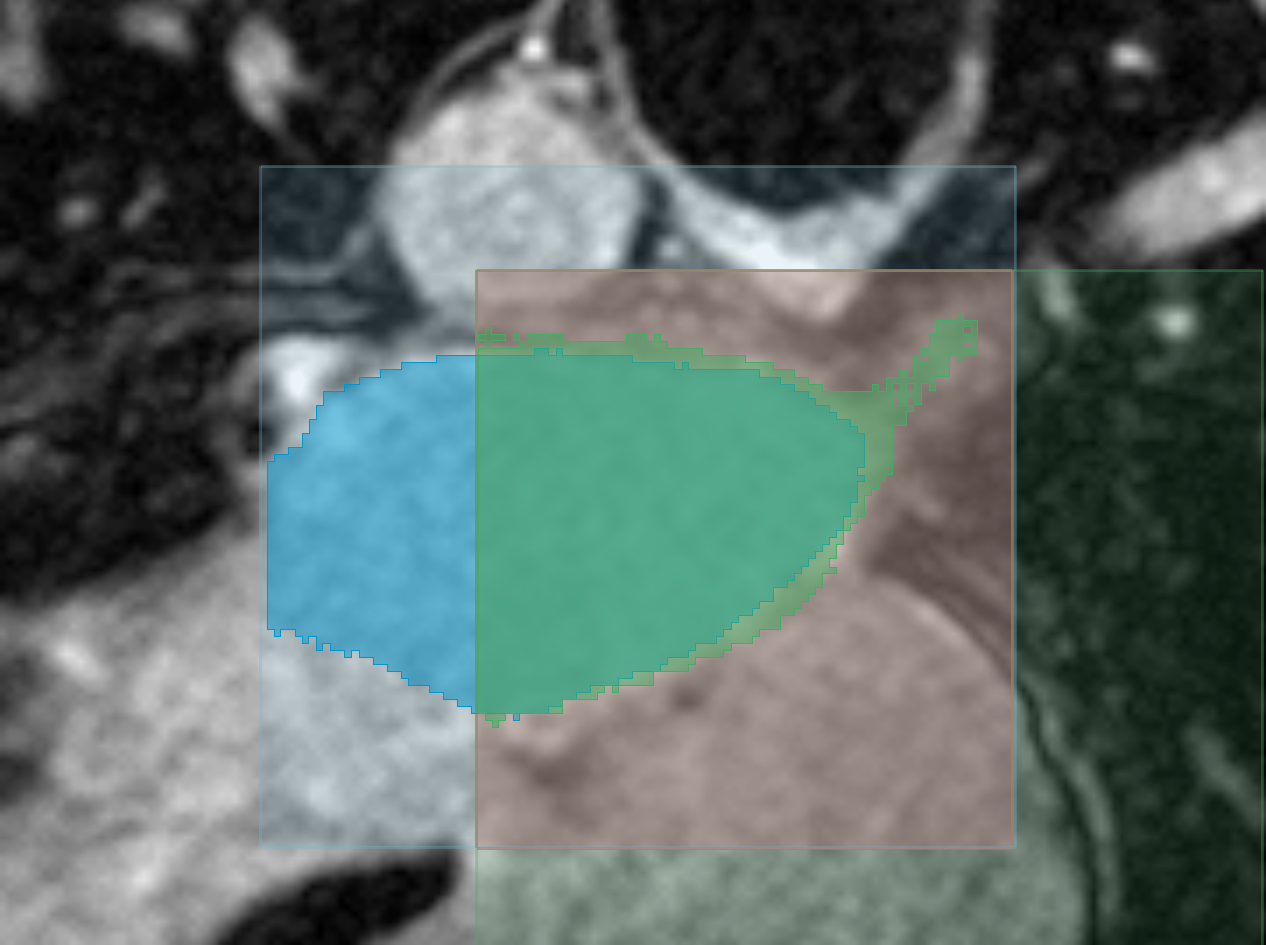}
     \includegraphics[width=1.\textwidth]{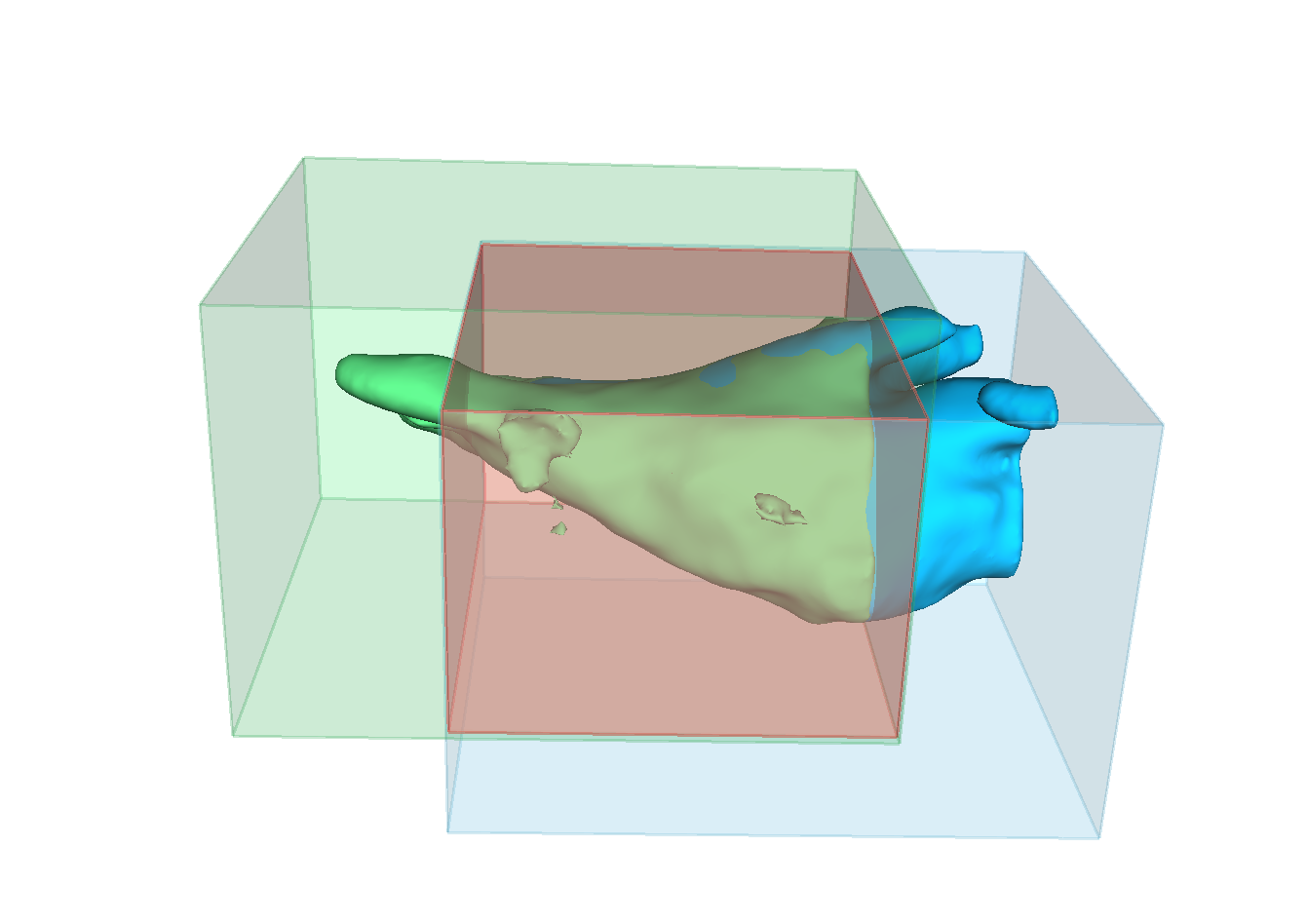}
     \caption*{CAC~\cite{lai2021semi}}
     \end{subfigure}
     \begin{subfigure}[b]{.192\textwidth}
     \includegraphics[width=1.\textwidth]{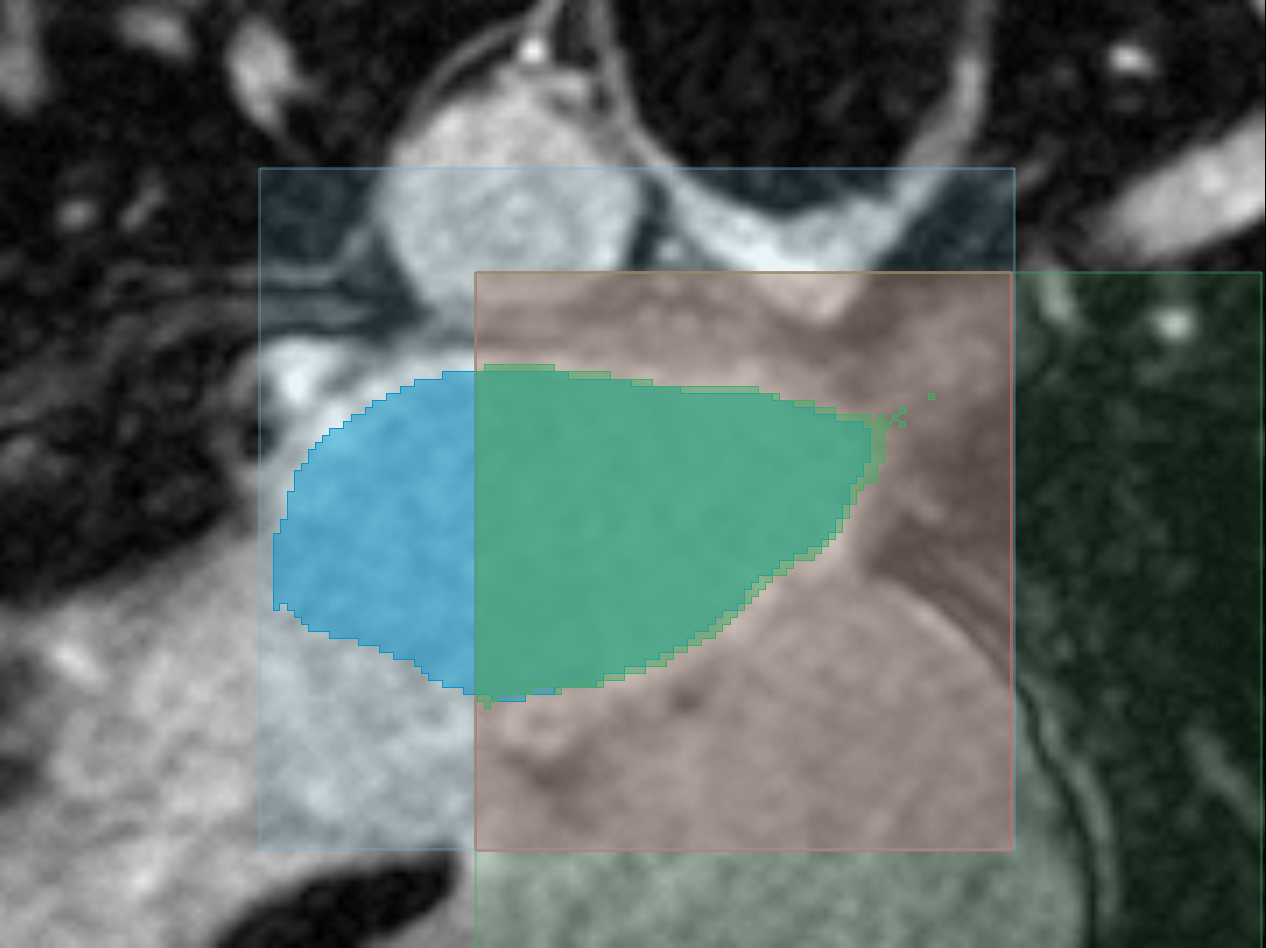}
     \includegraphics[width=1.\textwidth]{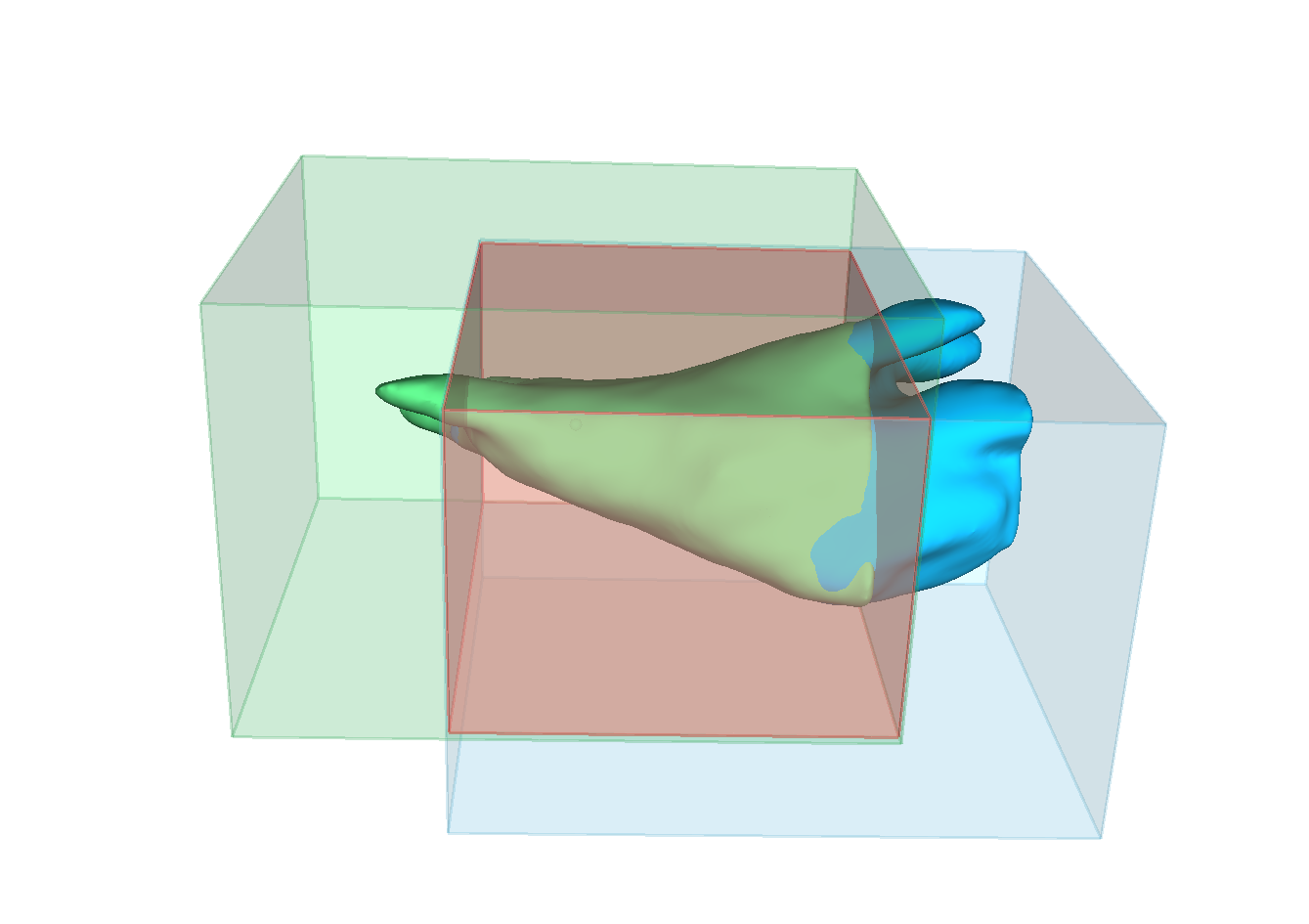}
     \caption*{Ours}
     \end{subfigure}
     \caption{Visualisation of segmentation results by SASSNet~\cite{li2020shape}, UA-MT~\cite{yu2019uncertainty}, CAC~\cite{lai2021semi}, and our TraCoCo (Ours) on a volume of the LA dataset~\cite{xiong2021global} using the original \textcolor{blue}{blue} volume and its \textcolor{green}{green} translated version, with the \textcolor{red}{red} box denoting the overlapping region. Notice how the segmentation is affected when the background changes with the translated volume.}
     \label{fig:traco_tmi_v2}
     \vspace{-10pt}
\end{figure*}

\indent 
Even though successful, the approaches above can inadvertently learn the segmentation pattern from the spatial input context of the training data rather than from the foreground objects to be segmented~\cite{lai2021semi}, which causes overfitting of \textbf{contextual patterns} in the labelled dataset of natural images. The contextual patterns in natural images can be represented by many types of objects, such as a `dog' standing nearby a `person', or a `road' that is localised under a `car'. 
In order to focus on the objects of interest (i.e., the objects to be segmented) instead of the context, CAC~\cite{lai2021semi} utilises contrastive learning to cluster the embeddings of pseudo-labelled data in the feature space based on an additional multi-layer perceptron (MLP). 
Such dense contrastive learning strategy is well known to be computationally expensive~\cite{wang2021dense, wang2021exploring} and limited by the quantity of positive and negative pairs that are selected for training, potentially leading to incomplete optimisation.
Furthermore, this sample selection is guided by potentially noisy pseudo labels produced from a fixed decoder. With the absence of network perturbation, the confirmation bias will inevitably be introduced during such semi-supervised training.  

Those limitations are more pronounced in the challenging 3D medical semi-supervised segmentation task, where the segmented object is likely to be surrounded by `background' organs distributed in a relatively stable topology. For example, the `Left Atrium' is always surrounded by `Pulmonary Vein' and `Mitral Valve'. The strong signals from these `background' organs increase the risk of `memorising' the training samples, leading to noisy prediction when incoming unlabelled data exhibit small variations in the organization of background organs. As depicted in Fig.~\ref{fig:traco_tmi_v2}, the \textcolor{blue}{blue} and \textcolor{green}{green} volumes contain different background organs, which can lead to  inconsistent segmentation inside the \textcolor{red}{overlapping} region, as shown by the segmentation results from SASSNet~\cite{yu2019uncertainty} (column 2) and UA-MT~\cite{li2020shape} (column 3).  
Note that CAC~\cite{lai2021semi} (column 4) still shows an inconsistent segmentation inside the red volume, but to a lesser degree compared with other methods~\cite{yu2019uncertainty, li2020shape}. 

Such inconsistent segmentation motivated us to
propose \ul{Tra}nslation \ul{Co}nsistent \ul{Co}-training (TraCoCo), which relies on more effective perturbations to make our approach less susceptible
to the `memorisation' of background patterns. TraCoCo employs \textbf{translation perturbation} to alleviate the `memorisation' impact, which is achieved by enforcing the segmentation result inside the intersection region to be similar between the volumes containing different backgrounds. 
Instead of relying on contrastive learning that is used to penalise \ul{a limited number of} positive/negative samples with a high computational cost training strategy, 
translation perturbation can effectively constrain all samples in the overlapping region, leading to better optimisation than CAC~\cite{lai2021semi}. 
Such perturbation is built upon the \textbf{co-training network perturbation}, that employs two models trained with  the Confident Regional Cross-Entropy (CRC) loss that operates directly in the voxel space, mitigating the confirmation bias observed in CAC~\cite{lai2021semi}. Our approach yields a result that is of higher quality than the other SOTA methods depicted in Fig.~\ref{fig:traco_tmi_v2}.
\new{Moreover, CutMix~\cite{yun2019cutmix} has demonstrated significant generalisation improvements for 2D data, when the annotated training set is small~\cite{chen2021-CPS}, so we adopt 3D-CutMix to further improve our overall performance.}
In summary, our contributions are: 
\begin{enumerate}
    \item A novel co-training strategy based on translation consistency that is designed to promote cross-model prediction consistency by alleviating the overfitting impact promoted by the surrounding `background organs'; and
    \item A new entropy-based CRC loss that takes only the most confident positive and negative pseudo labels to co-train the two models with the goal of improving training convergence and keeping the robustness to pseudo-labelling mistakes.
\end{enumerate}
Experiments are conducted on three public 3D semi-supervised segmentation benchmarks, including the Left Atrium (LA)~\cite{xiong2021global}, Pancreas-CT (Pancreas)~\cite{clark2013cancer} and BraTS19~\cite{menze2014multimodal}.
To further show the generalisation of  TraCoCo, we extend it to the 2D semi-supervised segmentation benchmark of Automated Cardiac Diagnosis Challenge (ACDC)~\cite{bernard2018deep}. Notably, our TraCoCo outperforms previous state-of-the-art (SOTA) methods in all benchmarks above, demonstrating our model's effectiveness and generalisability. 

\section{Related work}
\subsection{Semi-supervised Learning}
Semi-supervised learning leverages limited labelled data and many unlabelled data to boost the models' performance. \textbf{Consistency learning} is the dominant idea for semi-supervised learning, consisting of a method that enforces that neighbouring feature representations must share the same labels (i.e., smoothness assumption). The $\Pi$ model~\cite{laine2016temporal} proposes the weak-strong augmentation scheme, where the pseudo labels are produced via weakly augmented data and predictions are from its strong augmentation version (i.e., noise perturbation or colour-jittering). To improve pseudo labels' stability, the $\Pi$ model variant TemporalEmbedding~\cite{laine2016temporal} proposes an exponential moving average (EMA) method to accumulate all the historical results. Even though TemporalEmbedding improved stability, the hardware costs of storing the historical results are expensive. Mean Teacher (MT)~\cite{tarvainen2017mean}, the most popular semi-supervised structure in current research, solves such issues by ensembling the network parameters to a `teacher' network via EMA transfer to generate pseudo labels. The main drawback of those consistency-learning methods is that the ensemble of models/predictions eventually fall to the same local minimum during the training~\cite{berthelot2019mixmatch}, where teacher and student models will have the same behaviour for many complex data patterns. 
In this work, we adopt the Co-training framework~\cite{blum1998combining} that incorporates two different models to produce pseudo labels for unlabelled data and supervise each other simultaneously. Those two models share the same architecture, but their parameters are differently initialised, which brings more diversity when analysing complex input data patterns from challenging medical image segmentation tasks.

\vspace{-7pt}
\subsection{Semi-supervised Medical Image Segmentation}

Semi-supervised medical segmentation has received much attention in recent years, where most papers focus on estimating uncertain regions within pseudo labels. 
\cite{yu2019uncertainty, wang2021tripled, xia2020uncertainty} propose uncertainty-aware mechanisms that estimate the uncertainty regions for the pseudo labels via Monte Carlo dropout.
\cite{luo2020deep, luo2021urpc} learn the unsupervised samples progressively via prediction confidence or the distance between the predictions from different scales. 
Wang et al.\cite{wang2022uncertainty} estimate the uncertainty via a threshold-based entropy map. The methods in~\cite{mehrtash2020confidence, xu2023dual, xiang2022fussnet} measure  uncertainty by the calibration of multiple networks' predictions. 
ReFixMatch~\cite{nguyen2023boosting} selectively adopts high-confident results with `hard' cross-entropy loss and low-confident results with `soft' Kullback–Leibler (KL) divergence loss based on dual thresholds, Tri-U-MT~\cite{wang2021tripled} integrates reconstruction and signed distance field (SDF) prediction tasks to enhance its understanding of geometry information, and
AC-MT~\cite{xu2023ambiguity} estimates ambiguous regions between teacher and student models within the intermediate feature map and proposes a novel consistency loss to constrain their distance.

While exploring unlabelled data, these approaches aim to avoid the negative impact of trying to learn from potentially noisy regions. However, by doing that, they  have inadvertently neglected the learning of potentially correct pseudo-labels, resulting in insufficient convergence, particularly when dealing with complex input data patterns. 
This issue has been partly handled by \cite{ishida2017learning}, who proposed a complementary learning method to train from the inverse label information based on negative entropy. \cite{kim2019nlnl, kim2021joint, rizve2021defense} successfully explore negative and positive learning techniques to achieve a good balance between avoiding the learning from potentially noisy regions and mitigating the insufficient convergence issue.
In our paper, we address both issues of learning from noisy regions and insufficient convergence with \textbf{weighted} negative and positive learning method that restricts its attention to regions of high classification confidence for foreground and background objects. Besides, those methods~\cite{bai2023bidirectional, yu2019uncertainty, basak2023pseudo} are based on the Mean Teacher~\cite{tarvainen2017mean} framework, which can decrease their diversity for predicting the voxel-wise unlabelled data, resulting in a reduction of the model's generalisation capability. Co-training approaches~\cite{chen2021-CPS, wu2023compete, miao2023caussl, gao2023correlation, nguyen2023cross, chen2023decoupled, wang2023dhc, zhong2023semi, wu2022mutual, wu2021semi} have become a dominant approach in semi-supervised medical segmentation in the last few years. Some co-training approaches~\cite{wu2023compete, xia20203d} incorporate multiple branches to enhance the pseudo-label diversity, while other methods~\cite{chen2023decoupled, zhong2023semi, wu2022mutual} build those branches upon a shared decoder to improve the consistency effectiveness for the perturbation results. Another set of methods~\cite{luo2022semi, wang2022cnn} adopt co-training between transformer and CNN architectures to improve the network perturbation~\cite{chen2021-CPS}. Despite their promising results, those methods usually require increased computation cost for training the model. The closest approach to our method is DMD~\cite{xie2023deep} that explores mutual knowledge from predictions through a two-way Dice loss.
Another similar method is Cross-ALD~\cite{nguyen2023cross} which estimates the distribution of virtual adversarial noise~\cite{liu2021perturbed, verma2019interpolation} among different branches with the aim of formulating the most effective perturbation.  Despite the recently published method DHC~\cite{wang2023dhc}, which applies a weighted loss to address the imbalanced-distributed foreground categories, these previous methods have overlooked the importance of overfitting to the ``context" data in the background. 
The natural existence of confirmation bias in the co-training strategy results in the production of overconfident pseudo labels for background categories. 
Our translation consistency (TraCo) learning promotes the model to become robust to background patterns via consistency learning, alleviating such `memorisation' issue.

To achieve good generalisation, some papers~\cite{zheng2019semi, li2020shape, wang2023cat, lei2022semi} utilise adversarial learning, and others~\cite{liu2021perturbed,ouali2020semi,wu2021semi,wu2022mutual} rely on a perturbation across multiple networks to improve consistency. Recently, \cite{wang2022semi, wang2022efficient} have explored the benefits of multi-task learning to achieve better generalisation.  MTCL~\cite{wang2022efficient} includes feature regression and target detection tasks to improve pixel-level consistency.
Despite their success in recognising unlabelled patterns, these papers have overlooked the significance of in-context perturbations in input data, which are essential for encouraging the model to differentiate spatial patterns.  
\cite{lai2021semi} have proposed a contrastive learning framework to address such spatial information for urban driving scenes' segmentation. However, their contrastive learning framework depends on potentially mistaken segmentation predictions, which can lead to confirmation bias in the pixel-wise positive/negative sampling strategy, resulting in sub-optimal accuracy when dealing with complex medical images. Also, their framework does not explore network perturbations, which reduces its generalisation ability. In contrast, our TraCoCo successfully promotes cross-model prediction consistency under spatial input context variation, enabling our method to yield promising results across  2D and 3D medical image segmentation benchmarks.

\begin{figure*}[t!]
    \centering
    \hspace{-2.5cm}\includegraphics[width=\textwidth]{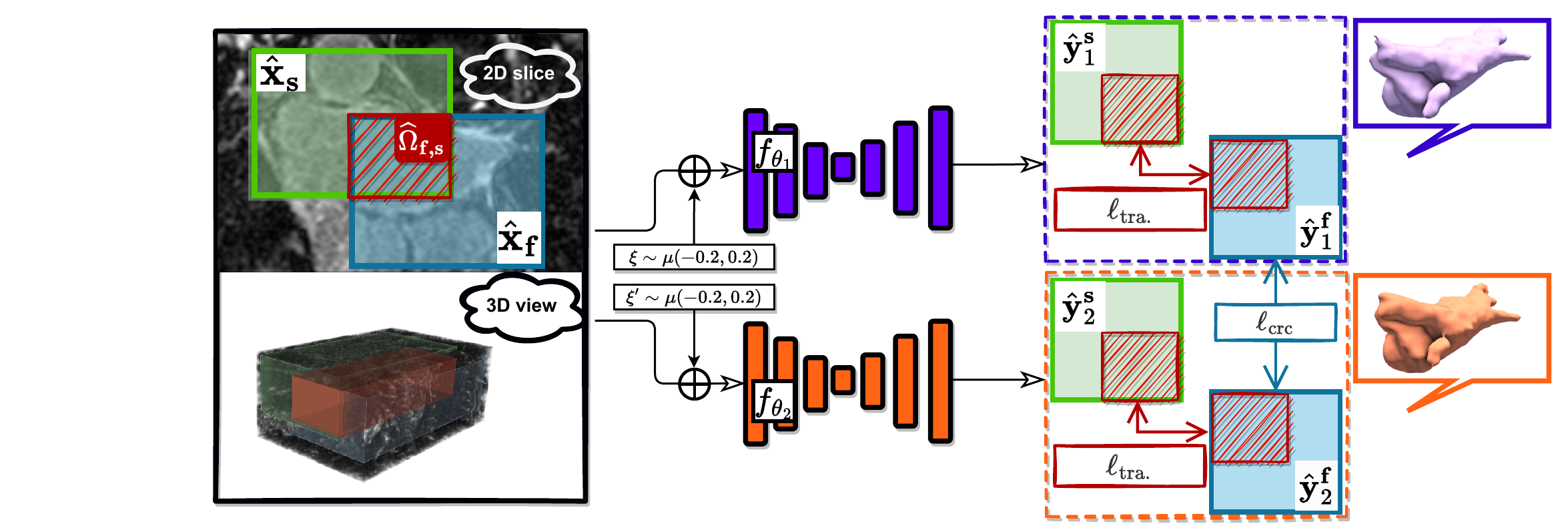}
    \caption{\textbf{TraCoCo's learning with unlabelled data.} The input volume $\mathbf{x}$ is randomly cropped into sub-volumes $\hat{\mathbf{x}}_{\mathbf{f}}$ in  \textcolor{blue}{blue} (extracted from image lattice $\widehat{\Omega}_{\mathbf{f}}$) and the translated volume $\hat{\mathbf{x}}_{\mathbf{s}}$ in  \textcolor{green}{green} (extracted from image lattice $\widehat{\Omega}_{\mathbf{s}}$), where these sub-volumes have a non-empty intersection denoted by the lattice $\widehat{\Omega}_{\mathbf{f,s}}= \widehat{\Omega}_{\mathbf{f}} \cap \widehat{\Omega}_{\mathbf{s}} \neq \emptyset$, represented in \textcolor{red}{red}. This pair of sub-volumes are perturbed with uniform noise $\xi$ before being used to produce the segmentation results \{$\hat{\mathbf{y}}_1^{\mathbf{f}}$, $\hat{\mathbf{y}}_1^{\mathbf{s}}$\}, \{$\hat{\mathbf{y}}_2^{\mathbf{f}}$, $\hat{\mathbf{y}}_2^{\mathbf{s}}$\} from the networks $f_{\theta_1}(.)$ and $f_{\theta_2}(.)$, respectively. The translation consistency loss $\ell_{\text{tra}}(.)$ in~\eqref{eq:ell_tra} will penalise differences in the responses within $\widehat{\Omega}_{\mathbf{f,s}}$, and the outputs $\hat{\mathbf{y}}_1^{\mathbf{f}}$,  $\hat{\mathbf{y}}_2^{\mathbf{f}}$ are used to minimize  $\ell_{crc}(.)$ in~\eqref{eq:ell_crc}. 
    On the right-hand side, we show the 3D segmentations  produced by the models (purple and orange volumes). 
    }
    \label{fig:main_strut}
    \vspace{-10pt}
\end{figure*}
\section{Method}
\label{sec:method}
\begin{table}[ht!]
\centering
\renewcommand{\arraystretch}{1.25}
\caption{\textbf{Symbol table.} The important symbols with their description in the Methodology section.} 
\label{tab:symbol}
\resizebox{.49\textwidth}{!}{\begin{tabular}{!{\vrule width 1pt}c|c!{\vrule width 1pt}}
\specialrule{1pt}{0pt}{0pt}
symbol & description            \\
\hline
$\mathbf{x}$     &      The input volumetric scan      \\
$\mathbf{y}$      &     The ground truth of the input scan     \\
$\mathbf{\Omega}$ &     The voxel lattice of the input scan \\
\hline
$\mathbf{\hat{x}_f}$, $\mathbf{\hat{x}_s}$ &    The cropped two sub-volumes from input         \\
$\mathbf{\Hat{y}^f}$, $\mathbf{\Hat{y}^s}$ &     The pseudo label of the cropped volume                       \\
$\mathbf{\Hat{\Omega}_f}$, $\mathbf{\Hat{\Omega}_s}$ & The voxel lattice of the cropped sub-volumes \\
\hline
$\mathbf{\hat{x}_{f,s}}$ &      The intersection input between two cropped volumes           \\
$\mathbf{\Hat{\Omega}_{f,s}}$ &      The voxel lattice of the  intersection region           \\
\hline
 $\theta_1$, $\theta_2$&  Two differently initialised network parameters\\
$\mathcal{D}_L$, $\mathcal{D}_U$ & The labelled and unlabelled datasets \\
\hline
$\ell_{dice}$ &  The Dice loss \\
$\ell_{ce}$ & The Cross-Entropy loss \\
$\ell_{kl}$ & The loss based on Kullback-Leibler divergence ($\mathbb{KL}$) \\
$\ell_{reg}$ & The regularisation loss based on Negentropy ($\mathbb{H}$) \\
$\ell_{crc}$ & Our proposed Confident Regional Cross-Entropy (CRC) loss \\
\specialrule{1pt}{0pt}{0pt}
\end{tabular}}
\end{table}
\vspace{-5pt}

For the 3D semi-supervised semantic segmentation, we have a small labelled set $\mathcal{D}_L = \{ (\mathbf{x},\mathbf{y})_i\}_{i=1}^{|\mathcal{D}_L|}$, where $\mathbf{x} \in \mathcal{X} \subset \mathbb{R}^{H \times W \times C}$ represents the input volume with size $H \times W$ and $C$ slices, and $\mathbf{y}  \in \mathcal{Y} \subset \{0,1\}^{H \times W \times C \times S}$ denotes the segmentation ground truth with $S$ categories, or binary segmentation when $S=1$.
We also have a large unlabelled set $\mathcal{D}_U = \{ \mathbf{x}_i\}_{i=1}^{|\mathcal{D}_U|}$ , where $|\mathcal{D}_L| << |\mathcal{D}_U|$. 
As depicted in Fig.~\ref{fig:main_strut}, our approach is a co-training framework~\cite{qiao2018deep} that contains two differently initialised networks.
Both segmentation models work with an input sub-volume $\widehat{\mathcal{X}} \subset \mathbb{R}^{\widehat{H} \times \widehat{W} \times \widehat{C}}$, where $\widehat{H} < H$, $\widehat{W} < W$, and $\widehat{C} < C$.
These sub-volumes are extracted from the original volume lattice $\Omega$ using the sub-lattice $\widehat{\Omega}_{\mathbf{f}} \subset \Omega$ of size $\widehat{H} \times \widehat{W} \times \widehat{C}$ and centred at volume index $\mathbf{f} \in \mathbb{N}^{3}$. 
The extracted two sub-volumes from volume $\mathbf{x}$ are identified as $\hat{\mathbf{x}}_\mathbf{f} = \mathbf{x}(\widehat{\Omega}\mathbf{_f})$ and $\mathbf{\hat{x}_s} = \mathbf{x}(\widehat{\Omega}_\mathbf{s})$, where their intersection region is $\mathbf{\hat{x}_{f,s}} = \mathbf{x}(\widehat{\Omega}_\mathbf{f} \cap \widehat{\Omega}_\mathbf{s}) = \mathbf{x}(\widehat{\Omega}_\mathbf{{f,s}})$
The models are represented by $f_{\theta}:\widehat{\mathcal{X}} \to [0,1]^{\widehat{H} \times \widehat{W} \times \widehat{C}}$, where $\theta_1,\theta_2 \in \Theta \subset \mathbb{R}^P$ denote the $P-$dimensional parameters of the two models. To enhance the accessibility of the paper, we have provided the description of the symbols in Tab.~\ref{tab:symbol}.

\vspace{-7pt}
\subsection{Translation Consistent Co-training (TraCoCo)}
The proposed translation consistent co-training (TraCoCo) optimisation is based on the following loss function minimization:
\begin{equation}
\begin{split}
    \theta^*_1,  \theta^*_2 = &\arg\underset{\theta_1,\theta_2}{\min} \:\: \ell_{sup}(\mathcal{D}_L,\theta_1,\theta_2)  \\
    & + \lambda \Big{(}\ell_{sem}(\mathcal{D}_U,\theta_1,\theta_2)+\ell_{tra}(\mathcal{D}_L \cup \mathcal{D}_U, \theta_1,\theta_2)\Big{)},
\end{split}
    \label{eq:main_loss}
\end{equation}
where the supervised learning loss is defined by
\begin{equation}
\begin{split}
    \ell_{sup}(\mathcal{D}_{L},\theta_1,\theta_2) = \sum_{(\mathbf{x},\mathbf{y}) \in \mathcal{D}_L} & \ell_{ce}(\mathbf{y}^{\mathbf{f}},\hat{\mathbf{y}}^{\mathbf{f}}_1) + \ell_{ce}(\mathbf{y}^{\mathbf{f}},\hat{\mathbf{y}}^{\mathbf{f}}_2)  \\
    & + \ell_{dice}(\mathbf{y}^{\mathbf{f}},\hat{\mathbf{y}}^{\mathbf{f}}_1) + 
    \ell_{dice}(\mathbf{y}^{\mathbf{f}},\hat{\mathbf{y}}^{\mathbf{f}}_2),
\end{split}    
    \label{eq:ell_sup}
\end{equation}
with $\ell_{ce}(.)$ denoting the voxel-wise cross-entropy loss, $\ell_{dice}(.)$ representing the volume-wise linearised Dice loss~\cite{yu2019uncertainty},  
$\hat{\mathbf{y}}_1^{\mathbf{f}} = f_{\theta_{1}}(\hat{\mathbf{x}}_{\mathbf{f}} + \xi)$ (similarly for  $\hat{\mathbf{y}}_2^{\mathbf{f}}$, with
$\xi \sim \mu(-0.2,0.2)$ denoting 
a sample from a uniform distribution in the range $[-0.2,0.2]$), and
$\mathbf{y}^{\mathbf{f}} = \mathbf{y}(\widehat{\Omega}_{\mathbf{f}})$ representing the sub-volume label. 
Also in~\eqref{eq:main_loss},   
$\ell_{sem}(.)$, defined below in~\eqref{eq:ell_sem}, represents the semi-supervised loss that uses the pseudo labels from the two models, $\ell_{tra}(.)$, defined below in~\eqref{eq:ell_tra}, denotes the translation consistency loss which enforces the cross-model consistency between the segmentation of two randomly-cropped sub-regions (containing varying spatial input contexts) of the original volume, and $\lambda$ in~\eqref{eq:main_loss} is a cosine ramp-up function that controls the trade-off between the losses.

\subsubsection{Translation Consistency Loss} 
We take the training volume $\mathbf{x} \in \mathcal{D}_L \cup \mathcal{D}_U$, and extract two sub-volumes centred at $\mathbf{f},\mathbf{s} \in \mathbb{N}^3$, with $\mathbf{f} \neq \mathbf{s} $ and denoted by $\hat{\mathbf{x}}_{\mathbf{f}}$ and $\hat{\mathbf{x}}_{\mathbf{s}}$, where their respective lattices have a non-empty intersection, i.e., $\widehat{\Omega}_{\mathbf{f,s}}= \widehat{\Omega}_{\mathbf{f}} \cap \widehat{\Omega}_{\mathbf{s}} \neq \emptyset$.
Our proposed translation consistency loss from~\eqref{eq:main_loss} is defined by
\begin{equation}
\begin{split}
\ell_{tra}(\mathcal{D}_L \cup \mathcal{D}_U,\theta_1,\theta_2) =  \sum_{\substack{\mathbf{x} \in \mathcal{D}_L \cup \mathcal{D}_U }}&
\alpha_1 \ell_{kl}(\hat{\mathbf{x}}_\textbf{f,s},\theta_1,\theta_2)  \\
& + \alpha_2\ell_{reg}(\hat{\mathbf{x}}_\textbf{f,s},\theta_1,\theta_2),
\end{split}
\label{eq:ell_tra}
\end{equation}
where the centres of the two sub-volumes denoted by $\mathbf{f,s}$ are uniformly sampled from the original volume lattice $\Omega$, 
$\alpha_1$ and $\alpha_2$ are hyper-parameters to balance the two loss functions ($\alpha_1=1.0$ and $\alpha_2=0.1$ in the experiments),
\begin{equation}
\begin{split}
    \ell_{kl}(\mathbf{x}_\textbf{f,s},\theta_1,\theta_2) =
    \sum_{\omega \in \widehat{\Omega}_{\mathbf{f},\mathbf{s}}} 
    & \mathbb{KL}(\hat{\mathbf{y}}_1^{\mathbf{f}}(\omega), \hat{\mathbf{y}}_1^{\mathbf{s}}(\omega))  \\
    & + \mathbb{KL}(\hat{\mathbf{y}}_2^{\mathbf{f}}(\omega), \hat{\mathbf{y}}_2^{\mathbf{s}}(\omega))
\end{split}    
    \label{eq:ell_spt}
\end{equation}
computes the Kullback-Leibler (KL) divergence between the segmentation outputs of the two models in the intersection region of two translated input sub-volumes represented by $\widehat{\Omega}_{\mathbf{f},\mathbf{s}}$,
and $\hat{\mathbf{y}}_1^{\mathbf{f}}$, $\hat{\mathbf{y}}_1^{\mathbf{s}}$, $\hat{\mathbf{y}}_2^{\mathbf{f}}$, and $\hat{\mathbf{y}}_2^{\mathbf{s}}$ are defined in~\eqref{eq:ell_sup}.
Also in~\eqref{eq:ell_tra}, we have
\begin{equation}
\begin{split}
    \ell_{reg}(\mathbf{x}_\textbf{f,s},\theta_1,\theta_2) =
    -\sum_{\omega \in \widehat{\Omega}_{\mathbf{f},\mathbf{s}}}
    & \mathbb{H}(\hat{\mathbf{y}}_1^{\mathbf{f}}(\omega)) + \mathbb{H}(\hat{\mathbf{y}}_2^{\mathbf{f}}(\omega)) \\ 
    &+ 
    \mathbb{H}(\hat{\mathbf{y}}_1^{\mathbf{s}}(\omega)) + \mathbb{H}(\hat{\mathbf{y}}_2^{\mathbf{s}}(\omega))
\end{split}
    \label{eq:ell_reg}
\end{equation}
that aims to balance the foreground and background classes in the training voxels,
where $\mathbb{H}(.)$ represents Shannon's entropy. 
\subsubsection{ Confident Regional Cross-Entropy (CRC) Loss} 
The semi-supervised loss $\ell_{sem}(.)$ in~\eqref{eq:main_loss} enforces the consistency between the segmentation results by the two models, as follows:
\begin{equation}
\begin{split}
    \ell_{sem}(\mathcal{D}_U, \theta_1, \theta_2) = -  \sum_{\mathbf{x} \in \mathcal{D}_U} \sum_{\omega \in \widehat{\Omega}_{\mathbf{f}}} &   \ell_{crc}\left (\hat{\mathbf{y}}_{1}^{\mathbf{f}}(\omega) ,\hat{\mathbf{y}}_{2}^{\mathbf{f}}(\omega) \right ) \\
    & + 
    \ell_{crc}\left (\hat{\mathbf{y}}_{2}^{\mathbf{f}}(\omega) ,\hat{\mathbf{y}}_{1}^{\mathbf{f}}(\omega) \right ),
\end{split}    
    \label{eq:ell_sem}
\end{equation}
where $\hat{\mathbf{y}}_{1}^{\mathbf{f}}(\omega) \in [0,1]^2$ represents the background ($\hat{\mathbf{y}}_{1}^{\mathbf{f}}(\omega)[0]$) and foreground ($\hat{\mathbf{y}}_{1}^{\mathbf{f}}(\omega)[1]$) segmentation probabilities in voxel $\omega \in \widehat{\Omega}_{\mathbf{f}}$ obtained from the softmax activation function of model $1$ (similarly for model $2$),
\begin{equation}
\begin{split}
  \ell_{crc}(\mathbf{a},\mathbf{b}) =  &  \mathbb{I}(\mathbf{a} > \gamma) \times  \ell_{ce}(\mathsf{OneHot}(\mathbf{a}), \mathbf{b}) \\
  & + \mathbb{I}(\mathbf{a} < \beta) \times  \ell_{ce}(1-\mathsf{OneHot}(\mathbf{a}), 1-\mathbf{b})),
\end{split}  
    \label{eq:ell_crc}
\end{equation}
with 
\begin{equation}
    \begin{split}
        \mathbb{I}(\mathbf{a} > \gamma) & = 
    \begin{cases}
\max(\mathbf{a})\hspace{0.2cm} \text{, if } \max(\mathbf{a}) > \gamma\\
0\hspace{0.2cm} \hspace{.85cm} \text{, otherwise }
\end{cases} \\
\mathbb{I}(\mathbf{a} < \beta) & = 
    \begin{cases}
1-\min(\mathbf{a})\hspace{0.2cm} \text{, if } \min(\mathbf{a}) < \beta\\
0\hspace{0.2cm} \hspace{1.35cm} \text{, otherwise }
\end{cases},
    \end{split}
\end{equation}
$\mathbf{a},\mathbf{b}$ being a output probability distribution from branches of voxel $\omega$, and $\gamma,\beta$ representing hyper-parameters to balance the foreground and background losses. 
Note that our proposed $\ell_{crc}(.)$ replaces the more common MSE loss~\cite{yu2019uncertainty,li2020shape,hang2020local,wang2021tripled,li2021hierarchical,wu2021semi,wu2022mutual} used in semi-supervised learning methods, where our goal is to maintain the MSE robustness to the pseudo-label mistakes and improve training convergence.
Particularly, in~\eqref{eq:ell_crc}, we select confidently classified foreground and background voxels by the first model to train both the positive and negative cross-entropy loss for the second model, and vice-versa.Other kinds of consistency loss, such as Kullback–Leibler (KL) and Jensen–Shannon (JS) divergences, could have been used, but our CRC loss is advantageous mostly because it does not exhibit the typical numerical instabilities that affect KL and JS divergences. To further demonstrate the effectiveness of our CRC loss compared with other loss functions, we show an ablation study in Tab.~\ref{tab:loss_ablation}.
\vspace{-7pt}
\subsection{3D CutMix} 

\new{To improve training generalisation, we adopt 3D CutMix~\cite{french2019semi,chen2021-CPS} on top of our method.}
This is achieved by randomly generating a 3D binary mask $\mathbf{m} \in \{0,1\}^{\Tilde{H} \times \Tilde{W} \times \Tilde{C}}$  containing a box of ones in a background of zeros, where the box location and size are randomly defined, and $\Tilde{H}$, $\Tilde{W}$ and $\Tilde{C}$ represent the mask size with $\Tilde{H} < H$, $\Tilde{W} < W$ and $\Tilde{C} < C$. We thus apply the binary mask to both the unlabelled data and the pseudo label from the second model (i.e., $\theta_2$), as defined below:
\begin{equation}
    \begin{aligned}
        \nu^{\mathbf{m}}_{ij} &\ = (1-\mathbf{m})\odot\hat{\mathbf{x}}_{\mathbf{i}}^{\mathbf{f}} + \mathbf{m} \odot \hat{\mathbf{x}}_{\mathbf{j}}^{\mathbf{f}}, \\
        \hat{\mathbf{y}}_{2, \mathbf{ij}}^{\mathbf{f}} &\ = (1-\mathbf{m})\odot\hat{\mathbf{y}}_{2,\mathbf{i}}^{\mathbf{f}} + \mathbf{m} \odot \hat{\mathbf{y}}_{2,\mathbf{j}}^{\mathbf{f}}, \\
    \end{aligned}
\end{equation}
where $\mathbf{i}, \mathbf{j}$ denotes different cases from same mini-batch with $\mathbf{i} \neq \mathbf{j}$.
To account for the 3D CutMix, the semi-supervised loss in~\eqref{eq:ell_sem} is re-defined with
\begin{equation}
\resizebox{.97\hsize}{!}{$
\begin{split}
    \ell_{sem}(\mathcal{D}_U, \theta_1, \theta_2) = \sum_{\mathbf{x}_i,\mathbf{x}_j \in \mathcal{D}_U} \sum_{\omega \in \widehat{\Omega}_{\mathbf{f}}} & 
    \ell_{crc}\left (f_{\theta_{1}}(\nu^{\mathbf{m}}_{ij}(\omega) + \xi),\hat{\mathbf{y}}_{2,ij}^{\mathbf{f}}(\omega) \right ) \\
    & + 
    \ell_{crc} \left (f_{\theta_{2}}(\nu^{\mathbf{m}}_{ij}(\omega) + \xi),\hat{\mathbf{y}}_{1,ij}^{\mathbf{f}}(\omega) \right ).
\end{split}    
$}
    \label{eq:ell_sem_redefine}
\end{equation}

\section{Experiments}
\subsection{Datasets} 
In order to evaluate the effectiveness of our approach, we conduct experiments on the following publicly available semi-supervised medical image datasets: the Left Atrium (LA)~\cite{xiong2021global}, Pancreas-CT (Pancreas)~\cite{clark2013cancer}, Brain Tumor Segmentation 2019 (BraTS19)~\cite{menze2014multimodal}, and Automated Cardiac Diagnosis Challenge (ACDC)~\cite{bernard2018deep}. \\
The \textbf{LA} dataset~\cite{xiong2021global} is from the \textit{Atrial Segmentation Challenge}\footnote{\hyperref[http://atriaseg2018.cardiacatlas.org/]{http://atriaseg2018.cardiacatlas.org/}} that contains $100$ 3D MRI volumes  with an isotropic resolution of 0.625 × 0.625 × 0.625 mm.
Following previous papers~\cite{yu2019uncertainty,li2020shape}, we crop at the centre of the heart region in the pre-processing stage and use 80 volumes for training and 20 for testing (we use the same training and testing sample IDs as in~\cite{yu2019uncertainty,li2020shape}). The \textbf{Pancreas} dataset~\cite{clark2013cancer} consists of 82 3D contrast-enhanced CT scans collected from $53$ male and $27$ female subjects at the National Institutes of Health Clinical Center. Each slice in the dataset has a resolution of 512 x 512, but the thickness ranges from 1.5 to 2.5 mm. In the pre-processing stage, we followed the approaches proposed in~\cite{wu2022mutual,luo2021dtc} by clipping voxel values into the range of [-125, 275] Hounsfield Units and re-sampling the data to an isotropic resolution of 1.0 x 1.0 x 1.0 mm. For our experiments, we used 62 samples for training and 20 samples for testing, following the same split sample IDs in~\cite{wu2022mutual} for fair comparisons.
The \textbf{BraTS19} dataset~\cite{menze2014multimodal} is from the \textit{Brain Tumor Segmentation Challenge}\footnote{\hyperref[https://www.med.upenn.edu/cbica/brats-2019/]{https://www.med.upenn.edu/cbica/brats-2019/}} and contains $335$ brain MRIs with tumour segmentation labels. Each sample in the dataset consists of four MRI scans of the brain: T1-weighted (T1), T1-weighted with contrast enhancement (T1-ce), T2-weighted (T2), and T2 fluid-attenuated inversion recovery (FLAIR). The resolution of each scan is 240 x 240 x 155, and they have been aligned into the same space (1.0 x 1.0 x 1.0 mm). For pre-processing, we follow the approach described in~\cite{chen2019multi} and use only the FLAIR sequences to simplify the task to binary classification. Similar to~\cite{yu2019uncertainty}, we centre crop each volume in BraTS19 with random offsets.
In the experiments, we use $250$ volumes for training, $25$ volumes for validation and $60$ for testing. The \textbf{ACDC}~\cite{bernard2018deep}\footnote{\hyperref[https://www.creatis.insa-lyon.fr/Challenge/acdc/]{https://www.creatis.insa-lyon.fr/Challenge/acdc/}} dataset contains three foreground segmentation classes, including the left ventricle cavity (LVen), right ventricle cavity (RVen), and myocardium (Myoc). The voxel spatial resolution ranges from $0.7 \times 1.92 \times 5$ to $0.7 \times 1.92 \times 10$ mm, and the volume sizes of the MRI volumes range from $154 \times 154 \times 6$ to $428 \times 512 \times 18$ voxels. The dataset contains 100 cardiac MRI scans, where the training, validation and test data contain 70, 10, and 20 scans, respectively.

\begin{table*}[t!]
\renewcommand{\arraystretch}{1.1}
\caption{\textbf{Evaluation on the Left Atrium dataset using VNet architecture} based on the partition protocols of $8$ and $16$ labelled data. The column \textbf{best} denotes the best checkpoint evaluation protocol and  \textbf{complexity} is measured during the inference phase following~\cite{wu2022mutual}.  Our results are highlighted in \textcolor{DarkCyan}{cyan}, with the best results shown in \textbf{bold}.}\label{tab:la_results}
\centering
\resizebox{.92\textwidth}{!}{\begin{tabular}{!{\vrule width 1.5 pt}r|c|c|c!{\vrule width 1.5pt}c|c|c|c|c|c!{\vrule width 1.5pt}} 
\specialrule{1.5pt}{0pt}{0pt}
\multicolumn{0}{!{\vrule width 1.5 pt}c|}{\multirow{2}{*}{Left Atrium}} & \multirow{2}{*}{best} &  \multicolumn{2}{c!{\vrule width 1.5pt}}{\# scan used} & \multicolumn{4}{c|}{measures}  &  \multicolumn{2}{c!{\vrule width 1.5pt}}{complexity}                                                             \\ 
\cline{3-10}
                       & & labelled & unlabelled             & Dice(\%)  & Jaccard(\%)  & ASD(Voxel)    & 95HD(Voxel)  & Param.(M) & Macs(G)   \\ 
\specialrule{1.5pt}{0pt}{0pt}
UA-MT~\cite{yu2019uncertainty} & \textcolor{darkred}{\xmark}  & 8        & 72 & 86.28 	& 76.11  	& 4.63 	& 18.71   & 9.44 & 47.02\\
SASSNet~\cite{li2020shape} & \textcolor{darkred}{\xmark}  & 8        & 72 & 85.22 	& 75.09  	& 2.89  	& 11.18   & 9.44 & 47.05\\
   
LG-ER~\cite{hang2020local}   & \textcolor{darkred}{\xmark} & 8        & 72  & 85.54 	& 75.12 	& 3.77 	& 13.29  & 9.44 & 47.02 \\
   
DUWM~\cite{wang2020double} & \textcolor{darkred}{\xmark}  & 8        & 72 & 85.91 	& 75.75 	& 3.31 	& 12.67    & 9.44 & 47.02\\

URPC~\cite{luo2021urpc} & \textcolor{darkred}{\xmark}           & 8       & 72                   & 85.01                     & 	74.36                     & 3.96     & 15.37               & 5.88 & 69.43       \\
CAC~\cite{lai2021semi} & \textcolor{darkred}{\xmark}               & 8       & 72                    & 87.61                     & 78.76                        & 2.93           & 9.65        & 9.44 & 47.02    \\
PS-MT~\cite{liu2021perturbed} & \textcolor{darkred}{\xmark}               & 8       & 72                    & 88.73                     & 79.02                        & 2.79           & 8.11        & 9.44 & 47.02    \\

\rowcolor{LightCyan}
TraCoCo {\small \textbf{(ours)}}                  & \textcolor{darkred}{\xmark}   & 8        & 72          & 89.29                               &  80.82                             &     2.28           &   6.92              & 9.44 & 47.02  \\ 
\hline
DTC~\cite{luo2021dtc}             & \textcolor{darkred}{\cmark}           & 8       & 72                   & 87.51                     & 	78.17                        & 2.36           & 8.23      & 9.44 & 47.05       \\
MC-Net~\cite{wu2021semi}  & \textcolor{darkred}{\cmark}  & 8       & 72           & 87.50   & 77.98                        & 2.30           & 11.28        &12.35 &95.15      \\ 
MC-Net+~\cite{wu2022mutual}  & \textcolor{darkred}{\cmark}  & 8       & 72          & 88.96   & 80.25                        & 1.86           & 7.93        & 9.44 & 47.02       \\ 

\textcolor{black}{UniMatch~\cite{yang2023revisiting}}  & \textcolor{darkred}{\cmark}  & 8       & 72 & 89.04 & 80.95  & 2.18 & 7.26  & 9.44 & 47.02 \\

\textcolor{black}{BCP~\cite{bai2023bidirectional}}  & \textcolor{darkred}{\cmark}  & 8       & 72 & 89.62 & 81.31  & \textbf{1.76} & 6.81  & 9.44 & 47.02 \\
\textcolor{black}{CAML~\cite{gao2023correlation}}  & \textcolor{darkred}{\cmark}  & 8       & 72 & 89.62 & 81.28 & 2.02 & 8.76 & 9.44 & 47.02 \\

\rowcolor{LightCyan}
TraCoCo {\small \textbf{(ours)}}            & \textcolor{darkred}{\cmark}         & 8        & 72           & \textbf{89.86}                               &  \textbf{81.70}                             &  2.01               &  \textbf{6.81}           & \textbf{9.44} & \textbf{47.02}   \\
\specialrule{1.5pt}{0pt}{0pt}
UA-MT~\cite{yu2019uncertainty} & \textcolor{darkred}{\xmark}  & 16        & 64 & 88.74 	& 79.94  	& 2.32 	& 8.39   & 9.44 & 47.02\\

SASSNet~\cite{li2020shape} & \textcolor{darkred}{\xmark}  & 16        & 64 & 89.16 	& 80.60  	& 2.26  	& 8.95    & 9.44 & 47.05\\
LG-ER~\cite{hang2020local}   & \textcolor{darkred}{\xmark} & 16        & 64 & 89.62 	& 81.31 	& 2.06 	& 7.16   & 9.44 & 47.02 \\
DUWM~\cite{wang2020double} & \textcolor{darkred}{\xmark} & 16        & 64   & 89.65  & 81.35 	& 2.03 	& 7.04   & 9.44 & 47.02 \\
URPC~\cite{luo2021urpc} & \textcolor{darkred}{\xmark}           & 16       & 64                   & 88.74                     & 	79.93                        & 3.66           & 12.73      & 5.88 & 69.43       \\
CAC~\cite{lai2021semi} & \textcolor{darkred}{\xmark}               & 16       & 64                    & 89.77                     & 81.53                        & 2.04           & 6.92        & 9.44 & 47.02    \\
PS-MT~\cite{liu2021perturbed} & \textcolor{darkred}{\xmark}               & 16       & 64                    & 90.02                     & 81.89                        & 1.92           & 6.74        & 9.44 & 47.02    \\

SCC~\cite{liu2022contrastive}        & \textcolor{darkred}{\xmark}               & 16       & 64                    & 89.81                     & 81.64                        & 1.82           & 7.15        & 12.35 & 95.15     \\

\rowcolor{LightCyan}
TraCoCo {\small \textbf{(ours)}}                  & \textcolor{darkred}{\xmark}   & 16       & 64   & 90.94            & 83.47        &1.79 & 5.49    & 9.44 & 47.02 \\

\hline
DTC~\cite{luo2021dtc}             & \textcolor{darkred}{\cmark}           & 16       & 64                   & 89.52                     & 	81.22                        & 1.96           & 7.07      & 9.44 & 47.05     \\
MC-Net~\cite{wu2021semi}    & \textcolor{darkred}{\cmark} & 16       & 64                  & 90.12                     & 82.12                        & 1.99           & 8.07         &12.35 &95.15      \\ 

MC-Net+~\cite{wu2022mutual}  & \textcolor{darkred}{\cmark} & 16       & 64                   & 91.07                     & 83.67                        & \textbf{1.67}          & 5.84       & 9.44 & 47.02      \\ 
\textcolor{black}{UniMatch~\cite{yang2023revisiting}}  & \textcolor{darkred}{\cmark}  & 16       & 64 & 90.99 & 83.57  & 1.98 & 6.07  & 9.44 & 47.02 \\
\textcolor{black}{BCP~\cite{bai2023bidirectional}}  & \textcolor{darkred}{\cmark}  & 16       & 64 & 91.26 & 84.01  & 1.86 & 5.76  & 9.44 & 47.02 \\
\textcolor{black}{CAML~\cite{gao2023correlation}}  & \textcolor{darkred}{\cmark}  & 16       & 64 & 90.78 & 83.19 & 1.68 & 6.11 & 9.44 & 47.02 \\
\rowcolor{LightCyan}
TraCoCo {\small \textbf{(ours)}}          & \textcolor{darkred}{\cmark}            & 16       & 64  & \textbf{91.51}            & \textbf{84.40}        & 1.79 & \textbf{5.63}  & \textbf{9.44} & \textbf{47.02} \\

\specialrule{1.5pt}{0pt}{0pt}
\end{tabular}}
\end{table*}
The LA and Pancreas datasets are commonly used benchmarks for testing semi-supervised 3D medical segmentation methods, so the results from competing methods are from the original papers. In the case of the BraTS19 dataset, the results of the current state-of-the-art (SOTAs) methods are based on our re-implementation. For ACDC, we utilise 2D slices of the 3D scans~\cite{bai2023bidirectional, basak2023pseudo, miao2023caussl} and perform the evaluation via slice-wise segmentation.

\vspace{-7pt}

\subsection{Implementation Details} 

Following~\cite{yu2019uncertainty,luo2021dtc}, we employ VNet~\cite{milletari2016v} for the experiments in the LA and Pancreas datasets, and we utilise 3D-UNet~\cite{cciccek20163d} for the BraTS19 dataset, following~\cite{chen2019multi}. We set the initial learning rate to be $5e^{-2}$, and decay via the poly learning rate scheduler $\left(1-\frac{\text{iter}}{\text{max\_iter}}\right )^{0.9}$, where the $\text{max\_iter}$ is $15,000$ in LA and Pancreas datasets and $30,000$ in BraTS19 dataset. 
To balance the supervised and unsupervised losses in~\eqref{eq:main_loss}, we utilize \textit{Cosine ramp-up} function with $40$ iterations, starting with $\lambda=0$. 
We adopt mini-batch SGD with momentum to train our model, where the momentum is fixed at $0.9$, and weight decay is set to $0.0005$. In the training of 3D volumes, each batch consists of $2$ voxel-wise labelled and $2$ unlabelled volumes, and they are cropped to be $112 \times 112\times 80$ on LA, and $96 \times 96 \times 96$ on BraTS19 and Pancreas. \new{We have tried various sizes of sub-volumes and achieved the best results based on the setup reported in the previous paper~\cite{bai2023bidirectional, yu2019uncertainty, wu2022mutual} across all benchmarks.} We employ the same online augmentation with random crop and flipping following~\cite{yu2019uncertainty,li2020shape}. In the training of 2D slices, we fix the batch size to $4$ for both labelled and unlabelled data and crop the slice to be $256 \times 256$ pixels~\cite{bai2019self, yang2023revisiting}. We utilise 2D UNet~\cite{zhou2018unet++} as the backbone and the input data is processed with the strong augmentations introduced in~\cite{yang2022self}.
The semi-supervised learning experimental setup partitions the original training set with 10\% and 20\% labelled samples and  the rest of samples remain unlabelled. 

\noindent\textbf{Hardware Requirements.} We use one NVIDIA GeForce RTX 3090 GPU to implement our algorithm for the LA and ACDC datasets, and one 32GB V100 for the BraTS19 and Pancreas datasets because of the larger input volume.

\vspace{-7pt}
\subsection{Evaluation Details}

\noindent\textbf{Inference.} During testing, we only use one network (i.e., $f_{\theta_1}(.)$) to produce our results. The final segmentation is obtained via the sliding window strategy, where the strides are 18x18x4 for the LA dataset, and 16x16x16 for the Pancreas and BraTS19 datasets, following previous works~\cite{yu2019uncertainty,li2020shape,luo2021dtc,wu2021semi,wu2022mutual}. The evaluation of ACDC dataset is based on the inference of each slice following~\cite{bai2019self, yang2023revisiting, basak2023pseudo}, where the final 3D prediction is obtained via concatenation of those results. We report last epoch results in BraTS19 datasets and the best results for the LA, Pancreas and ACDC datasets.

\noindent\textbf{Measurements.} We use four measures to quantitatively evaluate our method, which are Dice, Jaccard, the average surface distance (ASD), and the 95\% Hausdorff Distance (95HD). The Dice and Jaccard are measured in percentage, while ASD and 95HD are measured in voxels. Additionally, we measure complexity with the number of model parameters and multiply-accumulate (MAC) during the inference~\cite{wu2022mutual}.

\vspace{-7pt}

\subsection{Comparison with SOTA approaches}
\label{sec:compare_with_sotas}

\begin{table*}[t!]
\caption{\textbf{Evaluation on the Pancreas-CT dataset using VNet architecture} based on the partition protocols of $6$ and $12$ labelled data. The \textbf{scales} column denotes the methods that utilise multi-scale consistency, and \textbf{complexity} is measured during the inference phase, following~\cite{wu2022mutual}.  Our results are highlighted in \textcolor{DarkCyan}{cyan}, where $\dagger$ represents the best evaluation protocol and the best results are shown in \textbf{bold}.}\label{tab:pan_results}
\centering
\resizebox{.92\textwidth}{!}{\begin{tabular}{!{\vrule width 1.5 pt}r|c|c|c!{\vrule width 1.5pt}c|c|c|c|c|c!{\vrule width 1.5pt}} 
\specialrule{1.5pt}{0pt}{0pt}
\multicolumn{1}{!{\vrule width 1.5 pt}c|}{\multirow{2}{*}{Pancreas-CT}} & \multirow{2}{*}{scales} &  \multicolumn{2}{c!{\vrule width 1.5pt}}{\# scan used} & \multicolumn{4}{c|}{measures}  &  \multicolumn{2}{c!{\vrule width 1.5pt}}{complexity}                                                             \\ 
\cline{3-10}
                       & & labelled & unlabelled             & Dice(\%)  & Jaccard(\%)  & ASD(Voxel)    & 95HD(Voxel)  & Param.(M) & Macs(G)   \\ 
\specialrule{1.5pt}{0pt}{0pt}
UA-MT~\cite{yu2019uncertainty} & \textcolor{darkred}{\xmark}  &6       & 56 & 66.44 	& 52.02  	& 3.03 	& 17.04   & 9.44 & 41.45 \\
    
SASSNet~\cite{li2020shape} & \textcolor{darkred}{\xmark}  &6       & 56 & 68.97 	& 54.29  	&  \textbf{1.96}  	& 18.83   & 9.44 & 41.48 \\

DTC~\cite{luo2021dtc}             & \textcolor{darkred}{\xmark}           &6      & 56                   & 66.58                     & 	51.79                        &  4.16           & 15.46      & 9.44 & 41.48       \\
     
MC-Net~\cite{wu2021semi}  & \textcolor{darkred}{\xmark}  &6      & 56           & 69.07   & 54.36                        &  2.28         & 14.53         &12.35 &83.88      \\ 
URPC~\cite{luo2021urpc} & \textcolor{darkred}{\cmark}           &6      & 56                   & 73.53                     & 	59.44                     & 7.85     & 22.57               & 5.88 & 61.21       \\
 
MC-Net+~\cite{wu2022mutual}  & \textcolor{darkred}{\xmark}  &6      & 56          & 70.00   & 55.66                        & 3.87           & 16.03        & 9.44 & 41.45       \\ 
\textcolor{black}{MCCauSSL~\cite{miao2023caussl}} & \textcolor{darkred}{\xmark} & 6      & 56                  & 72.89                       & 58.06                        & 4.37          & 14.19       & 18.88 & 83.34      \\ 
PS-MT~\cite{liu2021perturbed}  & \textcolor{darkred}{\xmark}  &6      & 56          & 76.94   & 62.37                        & 3.66          & 13.12        & 5.88 & 61.21       \\

MC-Net+~\cite{wu2022mutual}  & \textcolor{darkred}{\cmark}  &6      & 56          & 74.01   & 60.02                        & 3.34          & 12.59        & 5.88 & 61.21       \\


\rowcolor{LightCyan}
TraCoCo {\small \textbf{(ours)}}                   & \textcolor{darkred}{\xmark}   &6       & 56          & \textbf{79.22}                               &  \textbf{66.04}                             &     2.57           &   \textbf{8.46}              & \textbf{9.44} & \textbf{41.45}  \\ 
\specialrule{1.5pt}{0pt}{0pt}
UA-MT~\cite{yu2019uncertainty} & \textcolor{darkred}{\xmark}  &12       & 50 & 76.10 	& 62.62  	& 2.43 	& 10.84   & 9.44 & 41.45 \\
SASSNet~\cite{li2020shape} & \textcolor{darkred}{\xmark}  &12       & 50 & 76.39 	& 63.17  	&  \textbf{1.42}  	& 11.06   & 9.44 & 41.48 \\
DTC~\cite{luo2021dtc}             & \textcolor{darkred}{\xmark}           & 12      & 50                  & 76.27                     & 	62.82                        & 2.20           & 8.70      & 9.44 & 41.48     \\
     
MC-Net~\cite{wu2021semi}    & \textcolor{darkred}{\xmark} & 12      & 50                 & 78.17                     & 65.22                        & 1.55           & 6.90         &12.35 &83.80      \\ 
URPC~\cite{luo2021urpc} & \textcolor{darkred}{\cmark}           &12      & 50                   & 80.02                     & 	67.30                     & 1.98     & 8.51               & 5.88 & 61.21       \\
MC-Net+~\cite{wu2022mutual}  & \textcolor{darkred}{\xmark} & 12      & 50                  &79.37                     &66.83                        & 1.72         &8.52       & 9.44 & 41.45      \\ 
\textcolor{black}{MCCauSSL~\cite{miao2023caussl}} & \textcolor{darkred}{\xmark} & 12      & 50                  & 80.92 & 68.26 & 1.53 &  8.11     & 18.88 & 83.34      \\ 

PS-MT~\cite{liu2021perturbed}  & \textcolor{darkred}{\xmark}  &12      & 50          & 80.74   & 68.15                        & 2.06          & 7.41        & 5.88 & 61.21       \\
MC-Net+~\cite{wu2022mutual}  & \textcolor{darkred}{\cmark} & 12      & 50                  & 80.59                     & 68.08                        & 1.74          & 6.47       & 5.88 & 61.21      \\ 

\rowcolor{LightCyan}
TraCoCo {\small \textbf{(ours)}}          & \textcolor{darkred}{\xmark}            & 12      & 50 & 81.80            & 69.56        & 1.49 & \textbf{5.70} & 9.44 & 41.45 \\
\hline
\textcolor{black}{BCP~\cite{bai2023bidirectional}}$\dagger$ & \textcolor{darkred}{\xmark} & 12      & 50 & 82.91 & 70.97& 2.25& 6.43 & 9.44 & 41.45 \\
\textcolor{black}{UniMatch~\cite{basak2023pseudo}}$\dagger$ & \textcolor{darkred}{\xmark} & 12 & 50 & 82.35 & 70.58  & 2.63 & 7.66   & 9.44 & 41.45\\

\rowcolor{LightCyan}
TraCoCo {\small \textbf{(ours)}}$\dagger$      & \textcolor{darkred}{\xmark}            & 12      & 50 & \textbf{83.36}            & \textbf{71.70}        & 1.74 & 7.34 & \textbf{9.44} & \textbf{41.45} \\


\specialrule{1.5pt}{0pt}{0pt}
\end{tabular}}
\end{table*}
\begin{table*}[t!]
\centering
\renewcommand{\arraystretch}{1.1}
\caption{\textbf{Evaluation on the BraTS19 dataset using 3D-UNet} based on different partition protocols of $25$ and $50$ labeled data. We replicate the SOTA results based on the published code. The \textbf{complexity} is measured during the inference phase, following~\cite{wu2022mutual}.  Our results are highlighted in \textcolor{DarkCyan}{cyan}, with the best results shown in \textbf{bold}. }\label{tab:brats_results}
\resizebox{.95\textwidth}{!}{
\begin{tabular}{!{\vrule width 1.5 pt}r|c|c!{\vrule width 1.5 pt}c|c|c|c|c|c!{\vrule width 1.5 pt}} 

\specialrule{1.5pt}{0pt}{0pt}
\multicolumn{0}{!{\vrule width 1.5 pt}c|}{\multirow{2}{*}{BraTS2019}} & \multicolumn{2}{c!{\vrule width 1.5 pt}}{\# scan used} & \multicolumn{4}{c|}{measures} & \multicolumn{2}{c!{\vrule width 1pt}}{complexity}                                                                \\ 
\cline{2-9}
                        & labelled & unlabelled             & Dice(\%)  & Jaccard(\%)  & ASD(Voxel)    & 95HD(Voxel) & Param.(M) &Macs(G)     \\ 

\specialrule{1.5pt}{0pt}{0pt}
UA-MT~\cite{yu2019uncertainty}       & 25       & 225     & 84.64   & 74.76  & 2.36   & 10.47  & 5.88 & 61.14  \\
ICT~\cite{verma2019interpolation}         & 25       & 225     & 83.71   & 73.62  & 2.65   & 12.09  & 5.88 & 61.14 \\
SASSNet~\cite{li2020shape}     & 25       & 225      & 84.73   & 74.89  & 2.44  & 9.88  & 5.88 & 61.17  \\ 
LG-ER~\cite{hang2020local} & 25        & 225  & 84.75 & 74.97  & 2.21 & 9.56  & 5.88 & 61.14 \\

URPC~\cite{luo2021urpc}         &25      & 225                   & 84.53                     & 	74.60                     & 2.55     & 9.79               & 5.88 & 61.21  \\ 
CAC~\cite{lai2021semi}             & 25       & 225                    & 83.97                     & 73.99                        & 2.93           & 9.60        & 5.88 & 61.14  \\ 

PS-MT~\cite{liu2021perturbed}              & 25       & 225                    & 84.88                     & 75.01                        & 2.49           & 9.93        & 5.88 & 61.14  \\ 
MC-Net+~\cite{wu2022mutual} & 25      & 225                  &84.96                     &75.14                        & 2.36         &9.45   & 5.88 & 61.14\\ 
\textcolor{black}{MCCauSSL~\cite{miao2023caussl}} & 25      & 225                 &83.54  & 73.46 & \textbf{1.98} & 12.53    & 11.76 & 128.28\\ 
\textcolor{black}{UniMatch~\cite{yang2023revisiting}}  & 25       & 225 & 85.03 & 75.91  & 2.42 & 9.50  & 5.88 & 61.14 \\
\textcolor{black}{BCP~\cite{bai2023bidirectional}}  &  25       & 225 & 85.14 & 76.01  & 2.88 & 9.89  & 5.88 & 61.14 \\

\rowcolor{LightCyan}
TraCoCo {\small \textbf{(ours)}}                     & 25        & 225             & \textbf{85.71} & \textbf{76.39}  & 2.27 & \textbf{9.20}   & 5.88 & 61.14 \\ 
\specialrule{1.5pt}{0pt}{0pt}
UA-MT~\cite{yu2019uncertainty}      & 50       & 200    & 85.32    & 75.93  & 1.98  & 8.68    & 5.88 & 61.14   \\ 
ICT~\cite{verma2019interpolation}        & 50       & 200    & 84.85    & 75.34  & 2.13  & 9.13  & 5.88 & 61.14    \\
SASSNet~\cite{li2020shape}    & 50       & 200    & 85.64   & 76.33  & 2.04  & 9.17  & 5.88 & 61.17    \\ 
LG-ER~\cite{hang2020local} & 50        & 200  & 85.67 & 76.36 & 1.99 & 8.92  & 5.88 & 61.14  \\

URPC~\cite{luo2021urpc}         &50      & 200                   & 85.38                     & 	76.14                     & \textbf{1.87}     & 8.36               & 5.88 & 61.21  \\ 
CAC~\cite{lai2021semi}             & 50       & 200                    & 84.96                     & 75.47                        & 2.18           & 9.02        & 5.88 & 61.14  \\ 
PS-MT~\cite{liu2021perturbed}              & 50       & 200                    & 85.91                     & 76.82                        &1.95           & 8.63         & 5.88 & 61.14  \\ 
MC-Net+~\cite{wu2022mutual} & 50      & 200                  & 86.02                     & 76.98                        & 1.98         &8.74     & 5.88 & 61.14  \\ 

\textcolor{black}{UniMatch~\cite{yang2023revisiting}}  & 50       & 200 & 85.94 & 76.93  & 1.97 & 8.68  & 5.88 & 61.14 \\
\textcolor{black}{BCP~\cite{bai2023bidirectional}}  &  50       & 200 & 86.13 & 77.24  & 2.06 & 8.99  & 5.88 & 61.14 \\

\rowcolor{LightCyan}
TraCoCo {\small \textbf{(ours)}}       & 50       & 200      & \textbf{86.69}  & \textbf{77.69}     & 1.93 & \textbf{8.04}  & 5.88 & 61.14 \\

\specialrule{1.5pt}{0pt}{0pt}
\end{tabular}}
\vspace{-10pt}
\end{table*}
\begin{table*}[t!]
\renewcommand{\arraystretch}{1.1}
\caption{\textbf{Evaluation on the ACDC dataset using the UNet architecture} based on the partition protocols of $3$, $7$ and $14$ labelled data. The measures are done with the best checkpoint evaluation protocol, and \textbf{Parameter(M)} is measured during the inference.  Our results are highlighted in \textcolor{DarkCyan}{cyan}, with the best results shown in \textbf{bold}.}\label{tab:acdc_results}
\centering
\resizebox{.88\textwidth}{!}{\begin{tabular}{!{\vrule width 1.5 pt}r|c|c!{\vrule width 1.5pt}c|c|c|c|c!{\vrule width 1.5pt}} 
\specialrule{1.5pt}{0pt}{0pt}
\multicolumn{0}{!{\vrule width 1.5 pt}c|}{\multirow{2}{*}{ACDC}}  &  \multicolumn{2}{c!{\vrule width 1.5pt}}{\# scan used} & \multicolumn{4}{c|}{measures}  &  \multirow{2}{*}{Param.(M)}                                                             \\ 
\cline{2-7}
                       & labelled & unlabelled             & Dice(\%)  & Jaccard(\%)  & ASD(Voxel)    & 95HD(Voxel)  &     \\ 
\specialrule{1.5pt}{0pt}{0pt}
UNet~\cite{ronneberger2015u}  {\scriptsize \textcolor{gray}{(fully supervised)}} & 70        & 0 & 92.10 & 85.45 & 0.43 & 1.72  & 1.81  \\
\hline
UA-MT~\cite{yu2019uncertainty} {\scriptsize \textcolor{gray}{(MICCAI'19)}}  & 3        & 67 & 46.04 & 35.97 & 7.75 & 20.08  & 1.81  \\
SASSNet~\cite{li2020shape} {\scriptsize \textcolor{gray}{(MICCAI'20)}}  & 3        & 67  & 57.77 & 46.14 & 6.06 & 20.05    &  1.81  \\
DTC~\cite{luo2021dtc} {\scriptsize \textcolor{gray}{(AAAI'21)}}      & 3        & 67                  & 56.90 & 45.67 & 7.39  & 23.36 &   1.81    \\
URPC~\cite{luo2021urpc} {\scriptsize \textcolor{gray}{(MedIA'21)}}   & 3       & 67                   &  55.87 & 44.64 & 3.74 & 13.60    &  1.81 \\
MC-Net~\cite{wu2021semi} {\scriptsize \textcolor{gray}{(MICCAI'21)}}   & 3       & 67           & 62.85 & 52.29 & 2.33 & 7.62   &  2.58  \\ 
Tri-U-Net~\cite{wang2021tripled} {\scriptsize \textcolor{gray}{(MICCAI'21)}} & 3       & 67           & 73.25 & 64.04 & 2.97 & 7.14     & 1.81 \\
SS-Net~\cite{wu2022exploring} {\scriptsize \textcolor{gray}{(MICCAI'22)}} & 3       & 67           & 65.83 & 55.38 & 2.28 & 6.67    & 1.83 \\
PS-MT~\cite{liu2021perturbed} {\scriptsize \textcolor{gray}{(CVPR'22)}}  & 3       & 67           & 86.94 & 77.90 & 2.18 & 4.65    & 1.81  \\
\textcolor{black}{BCP~\cite{bai2023bidirectional}} {\scriptsize \textcolor{gray}{(CVPR'23)}} & 3       & 67           & 87.59   & 78.67  &  \textbf{0.67}  & \textbf{1.90}    & 1.81 \\
\textcolor{black}{UniMatch~\cite{yang2023revisiting}} {\scriptsize \textcolor{gray}{(CVPR'23)}} & 3 & 67 & 87.61 & 78.68 & 1.97 & 4.13  & 1.81 \\
\textcolor{black}{DCNet~\cite{chen2023decoupled}} {\scriptsize \textcolor{gray}{(MICCAI'23)}} & 3 & 67 & 70.36 & 60.78 & 0.86 & 3.94   & 1.81 \\
\textcolor{black}{Cross-ALD~\cite{nguyen2023cross}} {\scriptsize \textcolor{gray}{(MICCAI'23)}} & 3 & 67 & 80.60 & 69.08 & 1.90 & 5.96   & 1.83  \\
\textcolor{black}{DMD~\cite{xie2023deep}} {\scriptsize \textcolor{gray}{(MICCAI'23)}} & 3 & 67 & 66.23 & 55.84 & 2.40  & 8.66 & 1.81 \\
\rowcolor{LightCyan}
TraCoCo {\small \textbf{(ours)}}                    & 3        & 67           & \textbf{89.46}                               &    \textbf{81.51}                        &        0.71      &    2.43      &  \textbf{1.81}     \\

\specialrule{1.5pt}{0pt}{0pt}
UA-MT~\cite{yu2019uncertainty} {\scriptsize \textcolor{gray}{(MICCAI'19)}}  & 7        & 63 & 81.65 & 70.64 & 2.02 & 6.88   & 1.81  \\
SASSNet~\cite{li2020shape} {\scriptsize \textcolor{gray}{(MICCAI'20)}}  & 7        & 63  & 84.50 & 74.34 & 1.86 &  5.42    & 1.81  \\
DTC~\cite{luo2021dtc} {\scriptsize \textcolor{gray}{(AAAI'21)}}     & 7        & 63   & 84.29 & 73.92 & 4.01 &  12.81 &  1.81    \\
URPC~\cite{luo2021urpc} {\scriptsize \textcolor{gray}{(MedIA'21)}}   & 7       & 63    &  83.10 & 72.41 & 1.53 & 4.84    &  1.81 \\
MC-Net~\cite{wu2021semi}  {\scriptsize \textcolor{gray}{(MICCAI'21)}}  & 7       & 63     & 86.44 & 77.04 & 1.84 &  5.50  & 2.58   \\ 
Tri-U-Net~\cite{wang2021tripled} {\scriptsize \textcolor{gray}{(MICCAI'21)}} & 7       & 63           & 86.93 & 77.92 & 2.06 & 5.94     & 1.81 \\
SS-Net~\cite{wu2022exploring} {\scriptsize \textcolor{gray}{(MICCAI'22)}} & 7       & 63           & 86.78 & 77.67 & 1.40 & 6.07     & 1.83 \\
PS-MT~\cite{liu2021perturbed} {\scriptsize \textcolor{gray}{(CVPR'22)}}  & 7       & 63           & 88.91 & 80.79 & 1.83 & 4.96    & 1.81  \\

\textcolor{black}{BCP~\cite{bai2023bidirectional}} {\scriptsize \textcolor{gray}{(CVPR'23)}}  & 7       & 63           & 88.84 & 80.62 & 1.17 & 3.98    & 1.81  \\
\textcolor{black}{PatchCL~\cite{basak2023pseudo}} {\scriptsize \textcolor{gray}{(CVPR'23)}}  & 7 & 63 & 89.10 & -  & 1.80 & 4.98   & 1.81 \\
\textcolor{black}{UniMatch~\cite{yang2023revisiting}} {\scriptsize \textcolor{gray}{(CVPR'23)}}  & 7 & 63 & 89.92 & 81.97 & 0.98 & 3.75  & 1.81 \\
\textcolor{black}{BCPCauSSL~\cite{miao2023caussl}} {\scriptsize \textcolor{gray}{(ICCV'23)}}  & 7       & 63           & 89.66 & 81.79 & 0.93 & 3.67     & 3.62 \\
\textcolor{black}{Cross-ALD~\cite{nguyen2023cross}} {\scriptsize \textcolor{gray}{(MICCAI'23)}} & 7 & 63 & 87.52 & 78.62 & 1.60 & 4.81    & 1.83 \\
\rowcolor{LightCyan}
TraCoCo {\small \textbf{(ours)}}                     & 7       & 63  &  \textbf{90.44}    & \textbf{83.01} & \textbf{0.42}  &  \textbf{1.41} & \textbf{1.81} \\

\specialrule{1.5pt}{0pt}{0pt}
UA-MT~\cite{yu2019uncertainty} {\scriptsize \textcolor{gray}{(MICCAI'19)}}  & 14        & 56 & 85.16 & 75.49  & 1.79 & 5.91   & 1.81 \\
SASSNet~\cite{li2020shape} {\scriptsize \textcolor{gray}{(MICCAI'20)}} & 14        & 56  & 86.45 & 77.20  & 1.98 & 6.63   &  1.81 \\
DTC~\cite{luo2021dtc} {\scriptsize \textcolor{gray}{(AAAI'21)}}     & 14        & 56   & 87.10& 78.15 & 1.99 & 6.76 &    1.81  \\
URPC~\cite{luo2021urpc} {\scriptsize \textcolor{gray}{(MedIA'21)}}   & 14       & 56    &  85.44 & 76.36 & 1.70  & 5.93    &  1.81 \\
MC-Net+~\cite{wu2022mutual} {\scriptsize \textcolor{gray}{(MedIA'22)}}  & 14       & 56     & 87.10 & 78.21  & 1.56 & 5.04 & 2.58   \\ 
PS-MT~\cite{liu2021perturbed} {\scriptsize \textcolor{gray}{(CVPR'22)}}  & 14       & 56           & 89.94 & 81.90 & 1.15 & 4.01    & 1.81  \\

\textcolor{black}{BCP~\cite{bai2023bidirectional}} {\scriptsize \textcolor{gray}{(CVPR'23)}} &14       & 56           & 89.52 & 81.62 & 1.03 & 3.69     & 1.81  \\
\textcolor{black}{UniMatch~\cite{yang2023revisiting}} {\scriptsize \textcolor{gray}{(CVPR'23)}}  & 14 & 56 & 90.47 &  82.96 & 0.99 & 2.24  & 1.81 \\
\textcolor{black}{PatchCL~\cite{basak2023pseudo}} {\scriptsize \textcolor{gray}{(CVPR'23)}} & 14 & 56 & 91.20  & -  & 1.49 & 3.82  & 1.81 \\
\textcolor{black}{BCPCauSSL~\cite{miao2023caussl}} {\scriptsize \textcolor{gray}{(ICCV'23)}}  & 14       & 56           & 89.99 & 82.34 & 0.88 & 3.60      & 3.62 \\
\rowcolor{LightCyan}
TraCoCo {\small \textbf{(ours)}}                     & 14       & 56  &   \textbf{91.24}  & \textbf{84.32} & \textbf{0.36}   & \textbf{1.29} & \textbf{1.81}  \\
\specialrule{1.5pt}{0pt}{0pt}
\end{tabular}}
\end{table*}

Our experimental results improve the SOTA for all datasets considered in this paper, namely LA~\cite{xiong2021global}, Pancreas~\cite{clark2013cancer} and BraTS19~\cite{menze2014multimodal} datasets. 
In the \textbf{LA dataset} (shown in Tab.~\ref{tab:la_results}), our approach yields promising performance in the last-result (last iteration checkpoint) and best-result (best checkpoint~\cite{wu2022mutual,wu2021semi,luo2021dtc}) evaluation protocols across all the partition protocols. 
For example, for the best-result evaluation, our results outperform the previous SOTA MC-Net+~\cite{wu2022mutual} by 0.9\% and 0.44\% in Dice measurement, for the partition protocols of 8 and 16 labelled data, respectively. 
For these same partition protocols, our Jaccard results are  1.45\% and 0.73\% better than the previous SOTA MC-Net+, and our 95HD results are 1.12 and 0.21-voxel better than the previous SOTA MC-Net+. 
The slightly larger ASD results from TraCoCo, compared with the current SOTA, can be explained by our approach being more sensitive to the presence of foreground objects without relying too much on background information, which may generate occasional noisy predictions for hard inliers. We provide more details in Section~\ref{sec:loss_discuss}.

\noindent For the \textbf{Pancreas dataset} in Tab.~\ref{tab:pan_results}, our results are substantially better than competing methods. For instance, our results are 5.21\% and 6.02\% better than MC-Net+~\cite{wu2022mutual} for Dice and Jaccard measurements, respectively, in the 6-label partition protocol. 
Similarly, our approach yields the best performance in 95HD measurements, where it outperforms MCNet+~\cite{wu2022mutual} and URPC~\cite{luo2021urpc} by 4.13 and 14.11 voxels, respectively. 
For the 12-label partition protocol, our TraCoCo is 1.21\% and 0.77 better than the previous SOTA MC-Net+ in terms of Dice and 95HD, respectively.
This demonstrates that our approach produces stable segmentation results when dealing with a limited amount of labeled data. Notably, our approach does not require the multi-scales consistency proposed in~\cite{luo2021dtc,wu2022mutual}, which involves a trade-off between higher computational costs and improved accuracy.

\begin{figure}[t!]
    \centering
    \includegraphics[width=.49\linewidth]{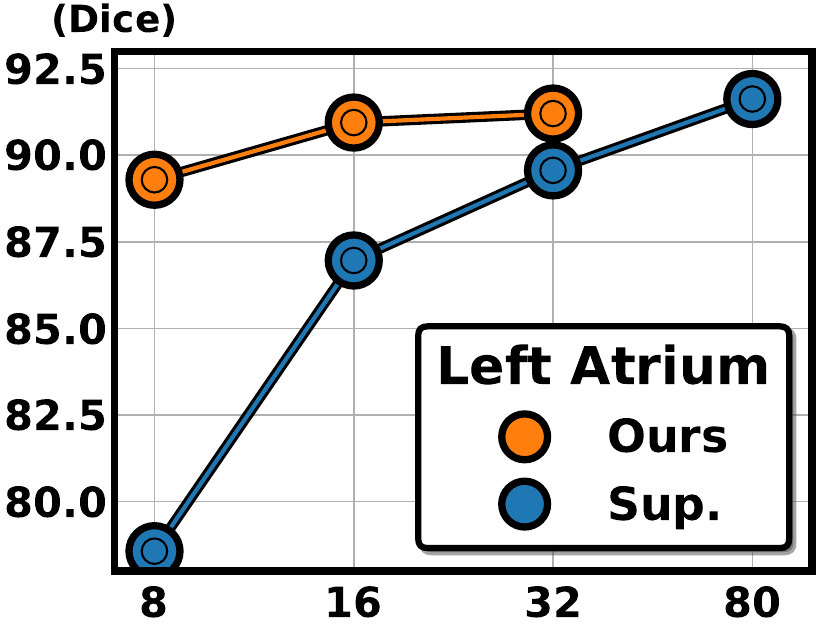}
     \includegraphics[width=.49\linewidth]{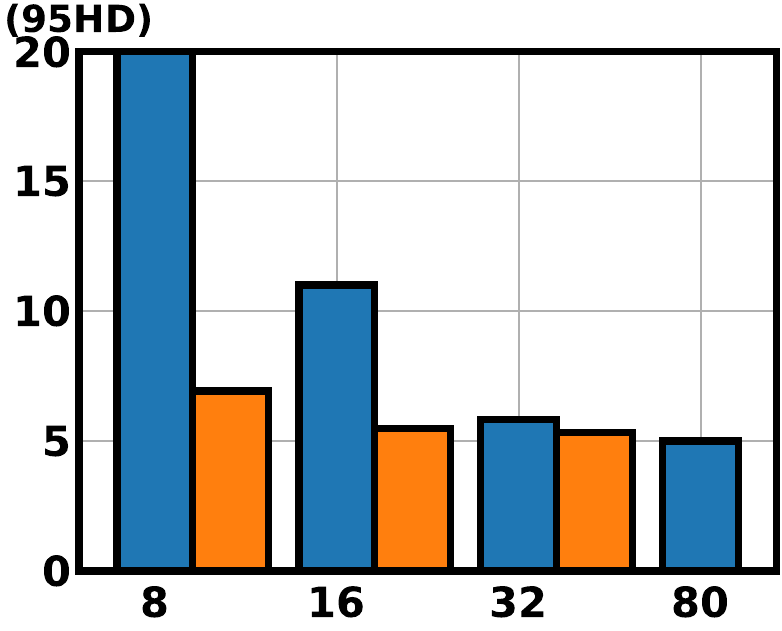}
     \caption{Dice (left) and 95HD (right) comparison between TraCoCo (Ours) and fully supervised (Sup.) trained with $\textbf{8, 16, 32}$ partition protocols and fully labelled data on LA dataset with VNet.\label{fig: sup_sub_la}}
    \vspace{-10pt}
\end{figure}

\begin{figure}[t!]
         \centering    
         \includegraphics[width=.49\linewidth]{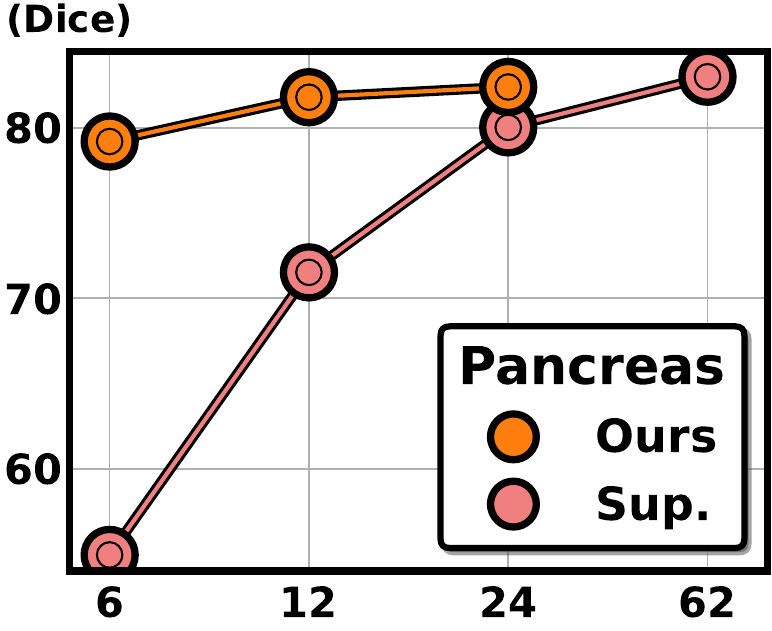}
         \includegraphics[width=.49\linewidth]{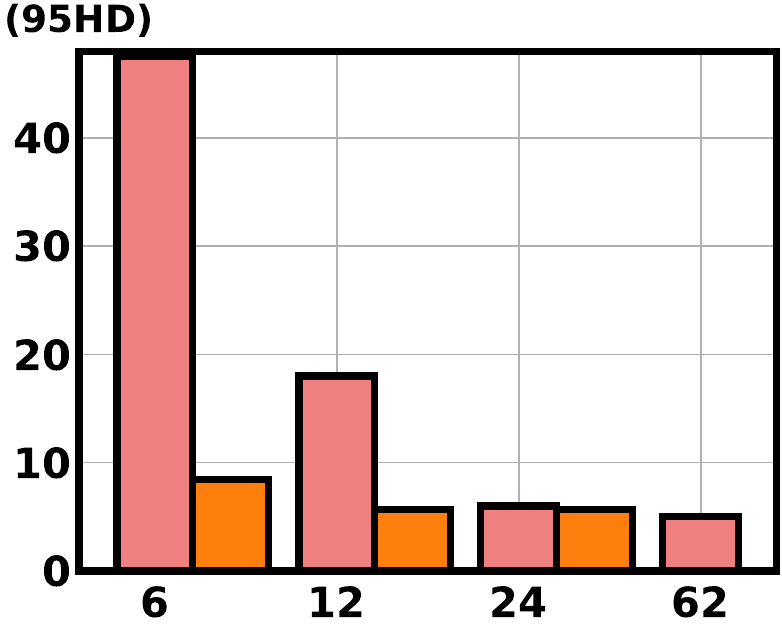}
         \caption{Dice (left) and 95HD (right) comparison between TraCoCo (Ours) and fully supervised (Sup.) trained with $\textbf{6, 12, 24}$ partition protocols and fully labelled data on Pancreas dataset with VNet.\label{fig: sup_sub_pan}} 
         
    \vspace{-10pt}
\end{figure}
\begin{figure}[t!]
         \centering    
         \includegraphics[width=.49\linewidth]{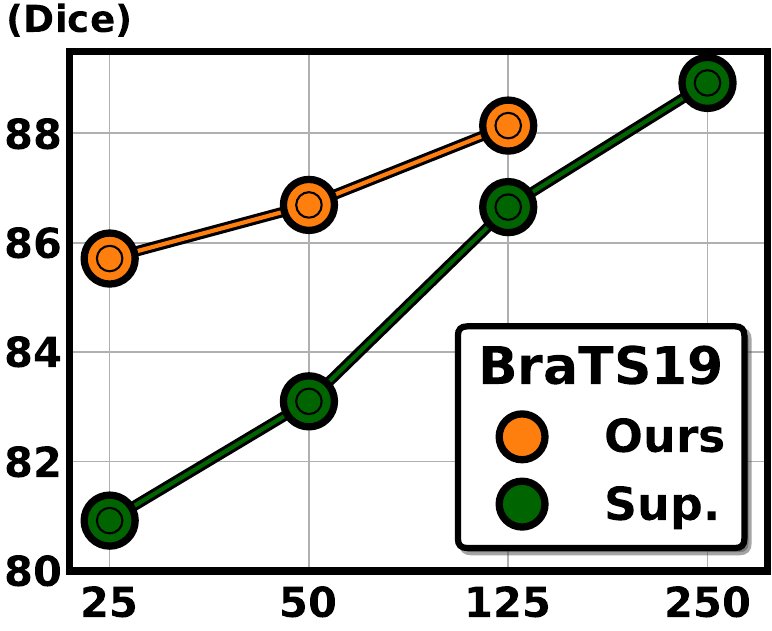}
         \includegraphics[width=.49\linewidth]{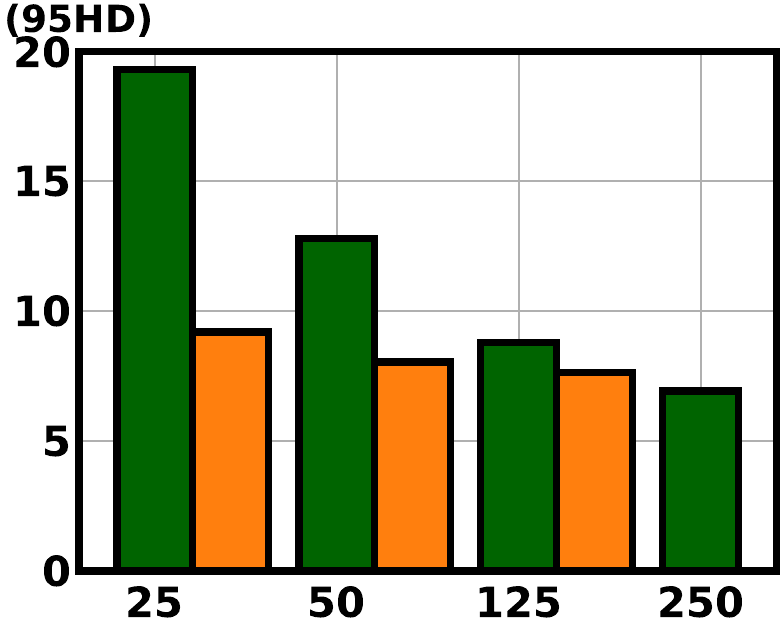}
         \caption{Dice (left) and 95HD (right) comparison between TraCoCo (Ours) and fully supervised (Sup.) trained with $\textbf{25, 50, 125}$ partition protocols and fully labelled data on BraTS19 dataset with VNet. \label{fig: sup_sub_brats}}          
    \vspace{-15pt}
\end{figure}
\noindent For the \textbf{BraTS19 dataset} in Tab.~\ref{tab:brats_results}, we replicate the SOTA methods using the code from published papers, but with small modifications to adapt them to the 3D-UNet backbone. Similarly to the LA and Pancreas datasets, our approach is substantially better than other approaches in terms of Dice and Jaccard measurements. For example, our approach increases Dice by $1.09\%$ and $1.02\%$, compared with SASSNet~\cite{li2020shape} and LG-ER-MT~\cite{hang2020local} under the $50$-sample partition protocol. Similarly, we increase Jaccard by almost 2\% over URPC~\cite{luo2021urpc} for the $25$-sample partition protocol, and 1.31\% for the $50$-sample partition protocol. \\

\noindent For the \textbf{ACDC dataset} in Tab.~\ref{tab:acdc_results}, we 
use 2D slices to train the U-Net~\cite{zhou2018unet++} architecture under the semi-supervised setup~\cite{bai2019self, yang2023revisiting}. Notably, our approach achieves SOTA results across all the partition protocols. For instance, it yields 0.54\% and 2.34\% improvements over UniMatch~\cite{yang2023revisiting} for the Dice and 95HD measurements.
Although Tri-U-MT~\cite{wang2021tripled} benefits from the advantages of multi-task consistency learning, the weak penalty from the MSE loss hinders its performance for segmenting the foreground objects, leading to a gap when compared with previous SOTA methods.

\subsection{Improvements Over Fully Supervised Learning}

We compare our results with the fully supervised learning (trained with the same labelled set indicated by the partition protocol) using VNet on LA in Fig.~\ref{fig: sup_sub_la}, VNet on Pancreas in Fig.~\ref{fig: sup_sub_pan}, and 3D-UNet on BraTS19 in Fig.~\ref{fig: sup_sub_brats}. Those three figures demonstrate that our approach successfully leverages unlabelled data and brings a dramatic performance boost. Particularly in the few-labelled data setup, our approach shows a $8.26\%$ Dice improvement for $8$ labelled samples on LA, $24.28\%$ for $6$ labelled samples on Pancreas, and $4.1\%$ for $25$ labelled samples on BraTS19. Our approach yields consistent improvements for more labelled data, such as $3.59\%$ and $2.41\%$ 
for $16$ and $32$ volumes, respectively, on LA, and $2.85\%$ and $1.49\%$ 
for $50$ and $125$ volumes, respectively, on BraTS19. Such improvements are achieved without bringing any additional computational costs for our approach during the inference phase, unlike the methods that utilise extra networks~\cite{wu2021semi,liu2022contrastive} or multi-scales~\cite{luo2021urpc, wu2022mutual} to explore the consistency learning.

\begin{table*}[ht!]
\renewcommand{\arraystretch}{1.1}
\centering
\caption{\new{\textbf{Ablation study of the main components of TraCoCo on (16-label protocol) LA~\cite{xiong2021global} and (12-label protocol) Pancreas~\cite{clark2013cancer} datasets.} 
The ablation study is based on the co-training architecture, where the \textcolor{gray}{gray} row indicates that we replace our proposed $\ell_{crc}(.)$ in the semi-supervised loss (Eq.~\ref{eq:ell_sem}) by a cross-entropy loss~\cite{chen2021-CPS} with 3D CutMix in Eq.~\ref{eq:ell_sem_redefine}.  TraCo represents the translation consistency loss $\ell_{tra}(.)$ in (Eq.~\ref{eq:ell_tra}), CRC denotes our proposed loss (Eq.~\ref{eq:ell_sem}) with $\ell_{crc}(.)$ in (Eq.~\ref{eq:ell_crc}). Our final results are in \textcolor{DarkCyan}{cyan} and the best results are marked in \textbf{bold}.}}\label{tab:comp_ablation_new_rebuttal_2}
\resizebox{.99\textwidth}{!}{\begin{tabular}{!{\vrule width 1pt}c|c!{\vrule width 1pt}c|c|c|c!{\vrule width 1pt}c|c|c|c!{\vrule width 1pt}}
\specialrule{1pt}{0pt}{0pt}
\multirow{2}{*}{TraCo}  & \multirow{2}{*}{CRC} & \multicolumn{4}{c!{\vrule width 1pt}}{LA~\cite{xiong2021global}} & \multicolumn{4}{c!{\vrule width 1pt}}{Pancreas~\cite{clark2013cancer}} \\
\cline{3-10}
\rule{0pt}{2ex}
&   & Dice $\uparrow$ & Jaccard $\uparrow$ & ASD $\downarrow$ & 95HD $\downarrow$     & Dice $\uparrow$ & Jaccard $\uparrow$ & ASD $\downarrow$ & 95HD $\downarrow$                \\ 
\specialrule{1pt}{0pt}{0pt}
\rowcolor{LightGray}
                 \xmark           &        \xmark                                         & 89.67 & 81.20  & 1.99 & 6.57   & 79.68 & 67.04 & 2.58  &  7.22    \\
\textcolor{black}{\cmark}                    &                                              & 90.50  & 82.71  & 1.84 & 6.74   & 81.26 & 69.07 & 1.64 & 5.81 \\ 
\specialrule{1pt}{0pt}{0pt}
\rowcolor{LightCyan}
\textcolor{red}{\cmark}                    & \textcolor{red}{\cmark}                                     & \textbf{90.94}  & \textbf{83.47} & \textbf{1.79}   & \textbf{5.49}    & \textbf{81.80}            & \textbf{69.56}        &\textbf{1.49} &  \textbf{5.70}  \\
 
\specialrule{1pt}{0pt}{0pt}
\end{tabular}}
\end{table*}

\begin{table*}[ht!]
\centering
\caption{\textbf{Loss ablation studies on (16-label protocol) LA and (12-label protocol) Pancreas datasets}. We study our TraCoCo with all components, but using different losses in Eq.~\eqref{eq:ell_sem}, including MSE, KL, CE and our CRC from Eq.~\eqref{eq:ell_crc} .}\label{tab:loss_ablation}
\resizebox{\textwidth}{!}{\begin{tabular}{!{\vrule width 1pt}c|c|c|c|c!{\vrule width 1pt}c|c|c|c!{\vrule width 1pt}c|c|c|c!{\vrule width 1pt}} 
\specialrule{1pt}{0pt}{0pt}
\multirow{2}{*}{$\ell_{sup}$} & \multicolumn{4}{c!{\vrule width 1pt}}{$\ell_{sem}$}  & \multicolumn{4}{c!{\vrule width 1pt}}{LA~\cite{xiong2021global}} & \multicolumn{4}{c!{\vrule width 1pt}}{Pancreas~\cite{clark2013cancer}}   \\ 
\cline{2-13}    
\rule{0pt}{2ex}
& MSE    &KL  & CE   & CRC   & Dice $\uparrow$ & Jaccard $\uparrow$ & ASD $\downarrow$ & 95HD $\downarrow$  & Dice $\uparrow$ & Jaccard $\uparrow$ & ASD $\downarrow$ & 95HD $\downarrow$                      \\ 
\specialrule{1pt}{0pt}{0pt}
$\surd$                          & $\surd$ &                  &         &                     &  89.77  & 81.49 & \textbf{1.73}   & 6.80  & 80.29 & 67.71 & \textbf{1.21} & 6.10 \\
$\surd$                          &         & $\surd$          &         &                     &  90.01  & 81.88 & 1.86   & 7.83  & 80.94 & 68.33 & 1.78 & 6.97    \\
$\surd$                          &         &                  & $\surd$        &               &  90.50  & 82.71 & 1.84 &  6.74  & 81.26 & 69.07 & 1.64 & 5.81        \\

$\mathbf{\surd}$                         &         &                  &       & $\mathbf{\surd}$             & \textbf{90.94}  & \textbf{83.47} & 1.79   & \textbf{5.49}   & \textbf{81.80}            & \textbf{69.56}        & 1.49 & \textbf{5.70}       \\
\specialrule{1pt}{0pt}{0pt}
\end{tabular}}
\vspace{-5pt}
\end{table*}

\subsection{Ablation Study of  Components}

We first present the ablation study of the different components of our approach on  LA~\cite{xiong2021global} and Pancreas~\cite{clark2013cancer} datasets in Tab.~\ref{tab:comp_ablation_new_rebuttal_2}. We use as baseline the co-training approach CPS~\cite{chen2021-CPS}, with Cross-entropy loss  replacing our proposed $\ell_{crc}(.)$ in the semi-supervised loss in Eq.~\ref{eq:ell_sem} (\textcolor{gray}{the first row}) and 3D CutMix in Eq.~\ref{eq:ell_sem_redefine}. Notably, the introduction of the translation consistency loss $\ell_{tra}(.)$ from Eq.~\ref{eq:ell_tra} is shown to improve generalisation with  Dice improvements of $0.83\%$ and $1.58\%$ on LA and Pancreas datasets, respectively.  Replacing the Cross-entropy loss in the semi-supervised loss Eq.~\ref{eq:ell_sem} by our CRC loss $\ell_{crc}(.)$ brings $0.44\%$ and $0.54\%$ Dice improvements on LA and Pancreas, respectively. Such improvements demonstrate that the model becomes more sensitive to the foreground objects with the strict penalisation of our CRC loss. The lower improvement in LA dataset is likely caused by a saturation of the results that is upper-bounded by a Dice of 91.47\% produced by the fully supervised method from~\cite{bai2023bidirectional}.

\begin{figure}[t!]
\centering
\begin{subfigure}[b]{.49\linewidth}
    \centering
    \includegraphics[width=1\linewidth]{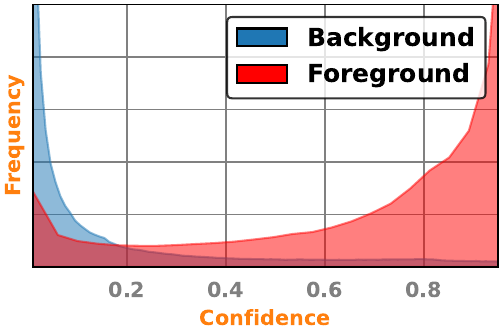}
    \caption{MSE Loss. \label{fig:sub_a_loss_dist}}    
    \end{subfigure}
\begin{subfigure}[b]{.49\linewidth}
    \centering
    \includegraphics[width=1\linewidth]{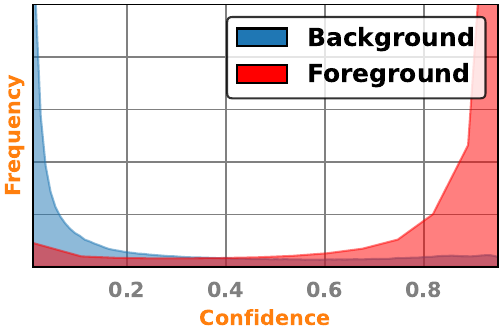}
    \caption{CRC Loss.\label{fig:sub_b_loss_dist}}    
\end{subfigure}
    \caption{Histogram of background and foreground confidence values in the Pancreas-CT dataset based on MSE (graph on the left) and our CRC (graph on the right) losses.\label{fig:loss_dist}}    
    \vspace{-15pt}
\end{figure}
\subsection{Ablation Study of Loss functions}
\label{sec:loss_discuss}

In Tab.~\ref{tab:loss_ablation}, we study different types of loss functions to be used by the semi-supervised loss $\ell_{sem}(.)$ in~\eqref{eq:ell_sem} on both LA and Pancreas datasets. Comparing with our CRC loss~\eqref{eq:ell_crc}, the KL divergence shows a $1.14\%$ worse Dice, while the CE loss, which disregards pseudo-label confidence, decreases Dice by around $0.78\%$. If we replace the CRC loss by the commonly used MSE loss~\cite{yu2019uncertainty,li2020shape,hang2020local,wang2021tripled,li2021hierarchical,wu2021semi,wu2022mutual}, we see a drop of almost $1.39\%$ Dice on LA dataset and $1.51\%$ on Pancreas dataset. 
We also observe that ASD for our CRC loss is slightly larger than that for the MSE loss, which can be explained by the fact that CRC is more sensitive than MSE to the presence of foreground objects without relying too much on background information, making it vulnerable to false positive detection due to noisy predictions from hard inliers. 
This trade-off has a negligible impact on the voxel measurement 95HD, which calculates the 95th percentile of distances between boundary points, thus discounting the negative impact caused by these rare noisy predictions. Conversely, MSE loss tends to rely too much on background features, resulting in unsatisfactory results for all other measurements, such as Dice, mIoU, and 95HD. 
We conclude this section with a demonstration of the superior capability of our CRC loss to segment foreground objects, compared with MSE loss.
We show a histogram of confidence distribution for foreground and background voxels in Fig.~\ref{fig:loss_dist} using the Pancreas-CT dataset. 
Note that the CRC loss enables a tighter clustering of foreground and background voxels around confidence values of 1 and 0, respectively, resulting a more effective classification of foreground and background voxels. 

\begin{figure}[t!]
    \centering
    \begin{subfigure}[b]{.49\linewidth}
        \captionsetup{width=\linewidth}
        \centering
        \includegraphics[width=\linewidth]{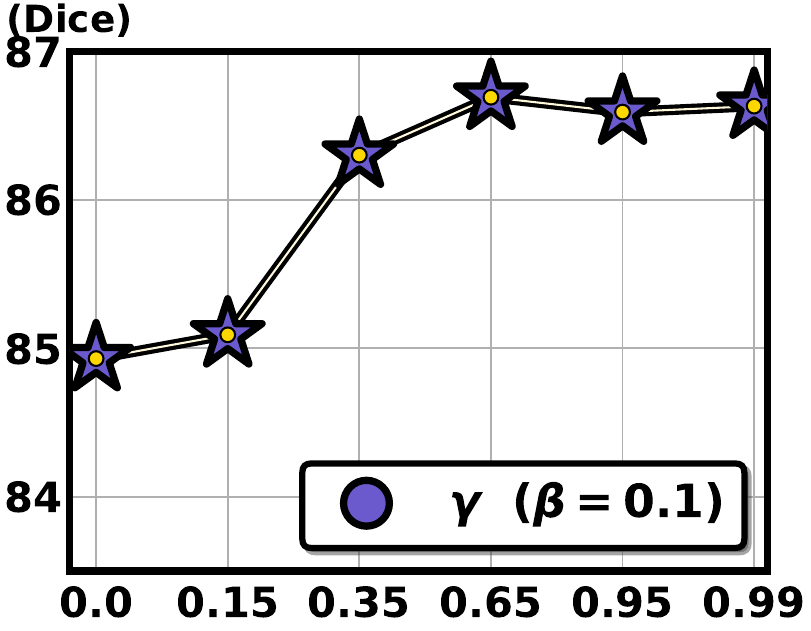} 
        \caption{Study of the positive threshold $\gamma$, while keeping the negative threshold fixed at $\beta=0.1$.\label{fig:ablation_gamma}}    
    \end{subfigure}    
    \begin{subfigure}[b]{.49\linewidth}
        \captionsetup{width=\linewidth}
        \centering
        \includegraphics[width=\linewidth]{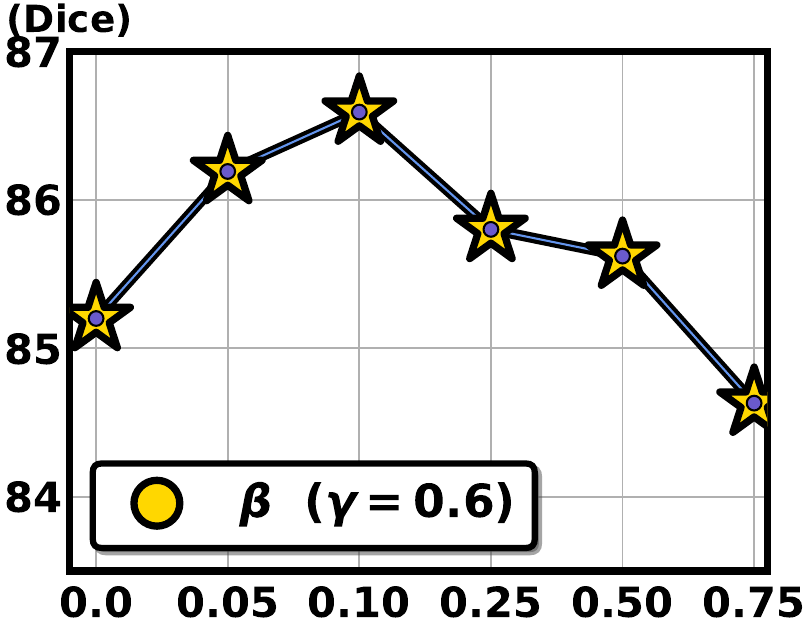} 
        \caption{Study of the negative threshold $\beta$, while keeping the positive threshold fixed at $\gamma=0.65$.\label{fig:ablation_beta}}    
    \end{subfigure}
    \caption{Ablation study of the positive ($\gamma$) and negative ($\beta$) thresholds for the CRC loss in Eq.~\eqref{eq:ell_crc} on the BraTS19 dataset.}\label{fig:ablation_gamma_beta}
    \vspace{-15pt}
\end{figure}

\begin{table*}[t!]
\renewcommand{\arraystretch}{1.1}
\caption{\textbf{Post-processing with Connected-component Thresholding (CCT)} on LA, BraTS19 and Pancreas datasets. CCT mitigates the issue of foreground noisy FP segmentation by trading off a minor computation cost increase during inference. $\dagger$ denotes the results are from best checkpoint evaluation protocol and the best results are in \textbf{bold}.}\label{tab:cct}
\resizebox{.99\textwidth}{!}{\begin{tabular}{!{\vrule width 1pt}c|c!{\vrule width 1pt}c|c|c|c!{\vrule width 1pt}c|c|c|c|c!{\vrule width 1pt}}

\specialrule{1pt}{0pt}{0pt}
\multirow{2}{*}{dataset}  & \multirow{2}{*}{Post-Process} & \multicolumn{4}{c!{\vrule width 1pt}}{10\% labelled data} & \multicolumn{4}{c|}{20\% labelled data} & \multirow{2}{*}{Time} \\
\cline{3-10}
                          &                               & Dice $\uparrow$   & Jaccard $\uparrow$  & ASD $\downarrow$  & 95HD $\downarrow$   & Dice $\uparrow$  & Jaccard $\uparrow$  & ASD $\downarrow$  & 95HD $\downarrow$  &  \\
\specialrule{1pt}{0pt}{0pt}
\multirow{4}{*}{LA}       & W/O CCT                          & 89.29    & 80.82   & 2.28 & 6.92      & 90.94    & 83.47   & 1.79 & 5.49      &   \textbf{1.386} (sec/case)   \\
                          & With CCT                           & \textbf{89.31}    & \textbf{80.83}   & \textbf{2.09} & \textbf{6.83}      & \textbf{90.96}    & \textbf{83.49}   &\textbf{1.70} & \textbf{5.45}       &   1.412 (sec/case)    \\
\cline{2-11}
                          & W/O CCT  ${\dagger}$                         & 89.86 & 81.70 & 2.01 & 6.81     &   91.51 & 84.40 & 1.79 & 5.63   &   \textbf{1.386} (sec/case)   \\
                          & With CCT ${\dagger}$                          & \textbf{89.94} &  \textbf{81.85}  & \textbf{1.72} & \textbf{6.67}     & \textbf{91.62}     & \textbf{84.59}   &  \textbf{1.45} &  \textbf{5.36}     &   1.412 (sec/case)    \\
\specialrule{1pt}{0pt}{0pt}
\multirow{2}{*}{Pancreas} & W/O CCT                           & 79.22 & 66.04  & 2.57  & 8.46     & 81.80   & 69.56 & 1.49 & 5.70       &    \textbf{9.892} (sec/case)    \\
                          & With CCT                              & \textbf{79.46} &  \textbf{66.35}          &   \textbf{1.89}       &  \textbf{7.26}    &  \textbf{81.99} &  \textbf{69.81} & \textbf{1.12} & \textbf{5.27}       & 9.975 (sec/case) \\
                          
\specialrule{1pt}{0pt}{0pt}
\multirow{2}{*}{BraTS19}  & W/O CCT                           & 85.71 & 76.39  & 2.27 & 9.20      & 86.69  & 77.69     & 1.93 & 8.04      &    \textbf{3.489} (sec/case)  \\
                          & With CCT                           & \textbf{85.79} & \textbf{76.45}  & \textbf{2.11} & \textbf{8.43}        & \textbf{86.73}  & \textbf{77.75} & \textbf{1.75} & \textbf{7.69}      &  3.587 (sec/case)     \\
\specialrule{1pt}{0pt}{0pt}
\multirow{2}{*}{ACDC}  & W/O CCT                           & 90.44 & 83.01 & 0.42 & 1.41   &      91.24  & 84.32 & 0.36   & 1.29      &    \textbf{0.037} (sec/slice)  \\
                          & With CCT                           & \textbf{90.50} & \textbf{83.09}  & \textbf{0.38} & \textbf{1.39}         & \textbf{91.29} & \textbf{84.40} & \textbf{0.33}     & \textbf{1.28}  &  0.052 (sec/slice)     \\
\specialrule{1pt}{0pt}{0pt}
\end{tabular}}
\end{table*}
\vspace{-10pt}
\subsection{Ablation Study of Hyper-parameters}
In Fig.~\ref{fig:ablation_gamma_beta}, we examine the influence of the positive threshold ($\gamma$) and negative threshold ($\beta$) for the CRC loss in Eq.~\eqref{eq:ell_crc}, where we fix $\beta=0.1$ to explore the effectiveness of $\gamma$ in Fig.~\ref{fig:ablation_gamma} and fix $\gamma=0.65$ to study $\beta$ in Fig.~\ref{fig:ablation_beta}. The results show a 1.76\% improvement when comparing $\gamma=0.65$ (86.69\% Dice) with $\gamma=0.3$ (84.93\% Dice), while the sensitivity to changes in $\gamma$ diminishes as its value increases. On the other hand, we observed that $\beta=0.1$ (86.69\% Dice) leads to a 1.39\% and 0.61\% improvement compared to $\beta=0.0$ (85.20\% Dice) and $\beta=0.05$ (86.08\% Dice), respectively. Nevertheless, a higher value of the negative threshold results in a dramatic performance drop due to the stronger penalties applied to potential noisy pseudo-labels, leading to confirmation bias.

\begin{table}[t!]
\renewcommand{\arraystretch}{1.1}
\caption{\textbf{Comparison between CAC~\cite{lai2021semi}, CAC with co-training and CutMix \textcolor{gray}{(Co)} and Translation consistency (TraCo) with co-training and CutMix \textcolor{gray}{(Co)}~\cite{chen2021-CPS} on Left Atrium~\cite{xiong2021global} and BraTS19~\cite{menze2014multimodal} datasets under different partition protocols.} The  $\Delta$ represents the improvements over the measurements between CAC and TraCo based on \textcolor{gray}{(Co)}-training architecture.}\label{tab:ablation_cac}
\centering
\resizebox{.49\textwidth}{!}{\begin{tabular}{!{\vrule width 1pt}r!{\vrule width 1pt}cccc!{\vrule width 1pt}cccc!{\vrule width 1pt}}
\specialrule{1pt}{0pt}{0pt}
\multicolumn{0}{!{\vrule width 1pt}c!{\vrule width 1pt}}{\# scan} & \multicolumn{4}{c!{\vrule width 1pt}}{Left Atrium (8)} & \multicolumn{4}{c!{\vrule width 1pt}}{Left Atrium (16)} \\
\hline
\multicolumn{0}{!{\vrule width 1pt}c!{\vrule width 1pt}}{measures}  & Dice$\uparrow$    & Jaccard$\uparrow$   & ASD$\downarrow$   & 95HD$\downarrow$  & Dice$\uparrow$    & Jaccard$\uparrow$   & ASD$\downarrow$    & 95HD$\downarrow$  \\
\hline
\rowcolor{LightGray}
CAC~\cite{lai2021semi}  & 87.61   & 78.76     & 2.93 & 9.65  & 89.77    & 81.53     & 2.04   & 6.92  \\
CAC {\scriptsize \textcolor{gray}{(Co)}}   & 88.28   & 79.88     & 2.73  & 7.74  & 90.27    & 82.89     & 2.02   & 5.84  \\
TraCo {\scriptsize \textcolor{gray}{(Co)}}  & \textbf{89.29}   & \textbf{80.82}     & \textbf{2.28}  & \textbf{6.92}  & \textbf{90.94}   & \textbf{83.47}     & \textbf{1.79}   & \textbf{5.49}  \\
$\Delta$ & \textcolor{green}{1.01$\uparrow$}  & \textcolor{green}{0.94$\uparrow$}  & \textcolor{red}{0.45$\downarrow$}  & \textcolor{red}{0.82$\downarrow$} & \textcolor{green}{0.67$\uparrow$} & \textcolor{green}{0.58$\uparrow$} & \textcolor{red}{0.23$\downarrow$} & \textcolor{red}{0.35$\downarrow$} \\
\specialrule{1pt}{0pt}{0pt}
\multicolumn{0}{!{\vrule width 1pt}c!{\vrule width 1pt}}{\# scan}  & \multicolumn{4}{c!{\vrule width 1pt}}{BraTS19 (25)}   & \multicolumn{4}{c!{\vrule width 1pt}}{BraTS19 (50)}    \\
\hline
\multicolumn{0}{!{\vrule width 1pt}c!{\vrule width 1pt}}{measures}  & Dice    & Jaccard   & ASD   & 95HD  & Dice    & Jaccard   & ASD    & 95HD  \\
\hline
\rowcolor{LightGray}
CAC~\cite{lai2021semi}  & 83.97   & 73.99     & 2.93 & 9.60  & 84.96    & 75.47     & 2.18   & 9.02  \\
CAC {\scriptsize \textcolor{gray}{(Co)}}   & 84.45   & 75.29     & 2.46  & 9.37  & 85.92   & 76.94     & 2.09   & 8.22  \\
TraCo {\scriptsize \textcolor{gray}{(Co)}} & \textbf{85.71}   & \textbf{76.39}     & \textbf{2.27}  & \textbf{9.20}  & \textbf{86.69}   & \textbf{77.69}     & \textbf{1.93}   & \textbf{8.04}  \\

$\Delta$ & \textcolor{green}{1.26$\uparrow$}  & \textcolor{green}{1.10$\uparrow$}  & \textcolor{red}{0.19$\downarrow$}  & \textcolor{red}{0.17$\downarrow$} & \textcolor{green}{0.77$\uparrow$} & \textcolor{green}{0.75$\uparrow$} & \textcolor{red}{0.16$\downarrow$} & \textcolor{red}{0.18$\downarrow$} \\
\specialrule{1pt}{0pt}{0pt}
\end{tabular}}
\end{table}
\vspace{-15pt}
\subsection{Connected Component Post-processing}

Connected component post-processing is a common approach to improve accuracy in medical image segmentation~\cite{isensee2021nnu,zhao2019multi,khanna2020deep}, where isolated false positive (FP) outlier predictions are eliminated with non-max suppression (NMS)~\cite{isensee2021nnu, li2020shape} or connected-component thresholding (CCT)~\cite{khanna2020deep}. 
In our work, we filter out the noisy regions that are smaller than $\mathbf{\frac{1}{1500}}$ of the input volume to remove isolated FP regions. 
As shown in Tab.~\ref{tab:cct}, CCT yields the best results for all the measurements with the filtering out of small isolated regions. Additionally, \textit{our results with CCT produce the SOTA results in the field in all three datasets for all measures.}

Also in Tab.~\ref{tab:cct}, we evaluate the run-time of our approach with and without CCT post-processing based on \textit{cc3d}\footnote{\url{https://github.com/seung-lab/connected-components-3d}} python package, following SASSNet~\cite{li2020shape}. For images in the LA dataset, our method without CCT, runs in $\mathbf{1.386}$ (sec/volume) on average and in $\mathbf{1.412}$ (sec/volume) with  CCT. Furthermore, our method with CCT only needs an extra $\mathbf{0.098}$ sec and $\mathbf{0.083}$ sec in BraTS19 and Pancreas, respectively.
\begin{figure}[t!]
    \centering
    \includegraphics[width=\linewidth]{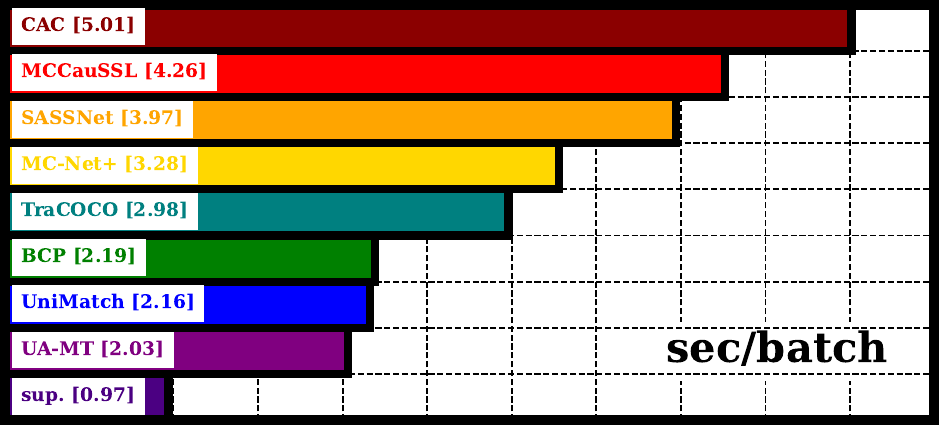}
    \caption{\textbf{Comparison of Training Time} on BraTS19 dataset, where all  approaches use the UNet3D architecture and run on a single NVIDIA V100.} \label{fig:efficiency}    
    \vspace{-15pt}
\end{figure}

\subsection{Training Efficiency}

In the Fig.~\ref{fig:efficiency}, we compare the training time of the SOTA approaches running on a single NVIDIA A6000, where we set the batch size to $8$ and the resolution to $96 \times 96 \times 96$. 
We found that \textcolor{moredarkred}{CAC~\cite{lai2021semi}} has the highest computation cost, which can be attributed to the high complexity of the voxel-wise positive and negative sample selection for the contrastive learning. \textcolor{teal}{TraCoCo} demonstrates better training efficiency, taking only half the training time under the same conditions compared to CAC. However, compared to the mean-teacher approaches (\textcolor{darkgreen}{BCP~\cite{bai2023bidirectional}}, \textcolor{blue}{UniMatch~\cite{yang2023revisiting}}, \textcolor{darkmagenta}{UA-MT~\cite{yu2019uncertainty}}), which rely on exponential moving average (EMA), \textcolor{teal}{TraCoCo} has larger training time due to the  backpropagation from the other branch.

\vspace{0.2cm}
\subsection{Visualisation Results} 

\begin{figure*}[t!]
    \centering
     \begin{subfigure}[b]{1.0\textwidth}
         \centering    
     \begin{subfigure}[b]{1.\textwidth}
         \centering    
            \begin{subfigure}[b]{.192\textwidth}
             \includegraphics[width=\textwidth]{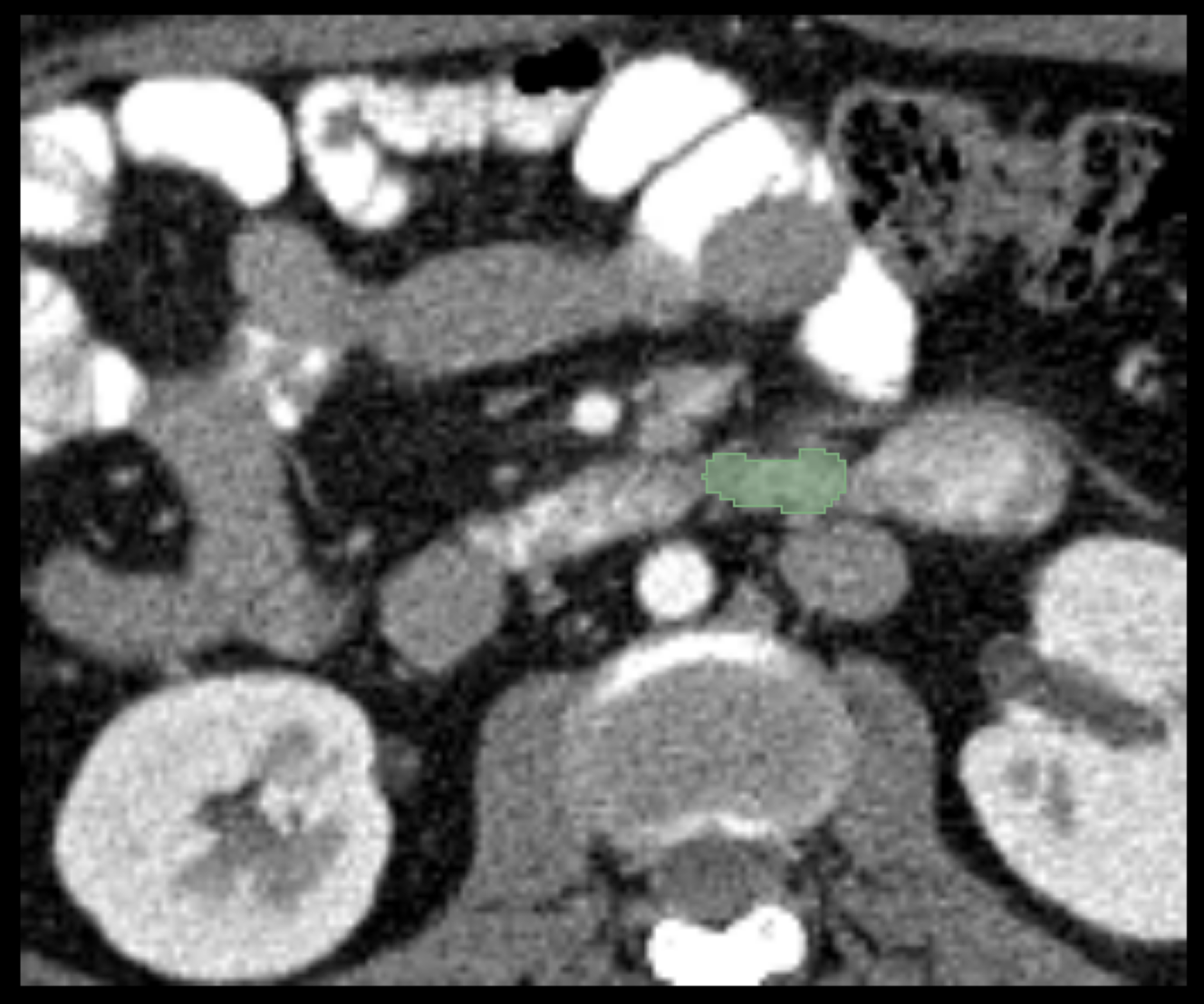}
             \includegraphics[width=\textwidth]{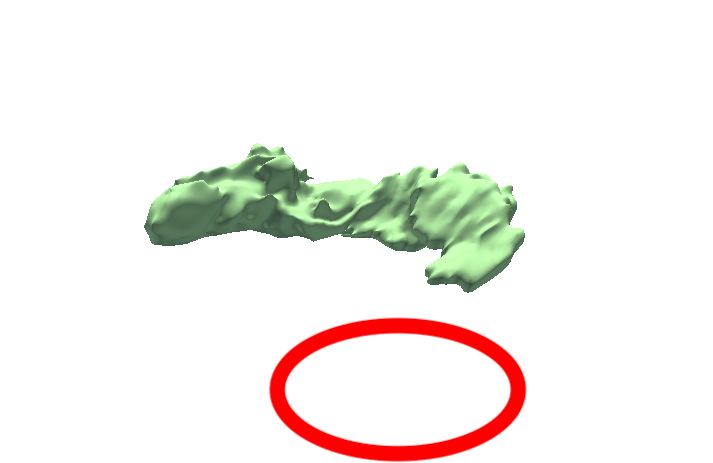}
             \caption*{Ground truth}
         \end{subfigure}
         \begin{subfigure}[b]{.192\textwidth}
             \includegraphics[width=\textwidth]{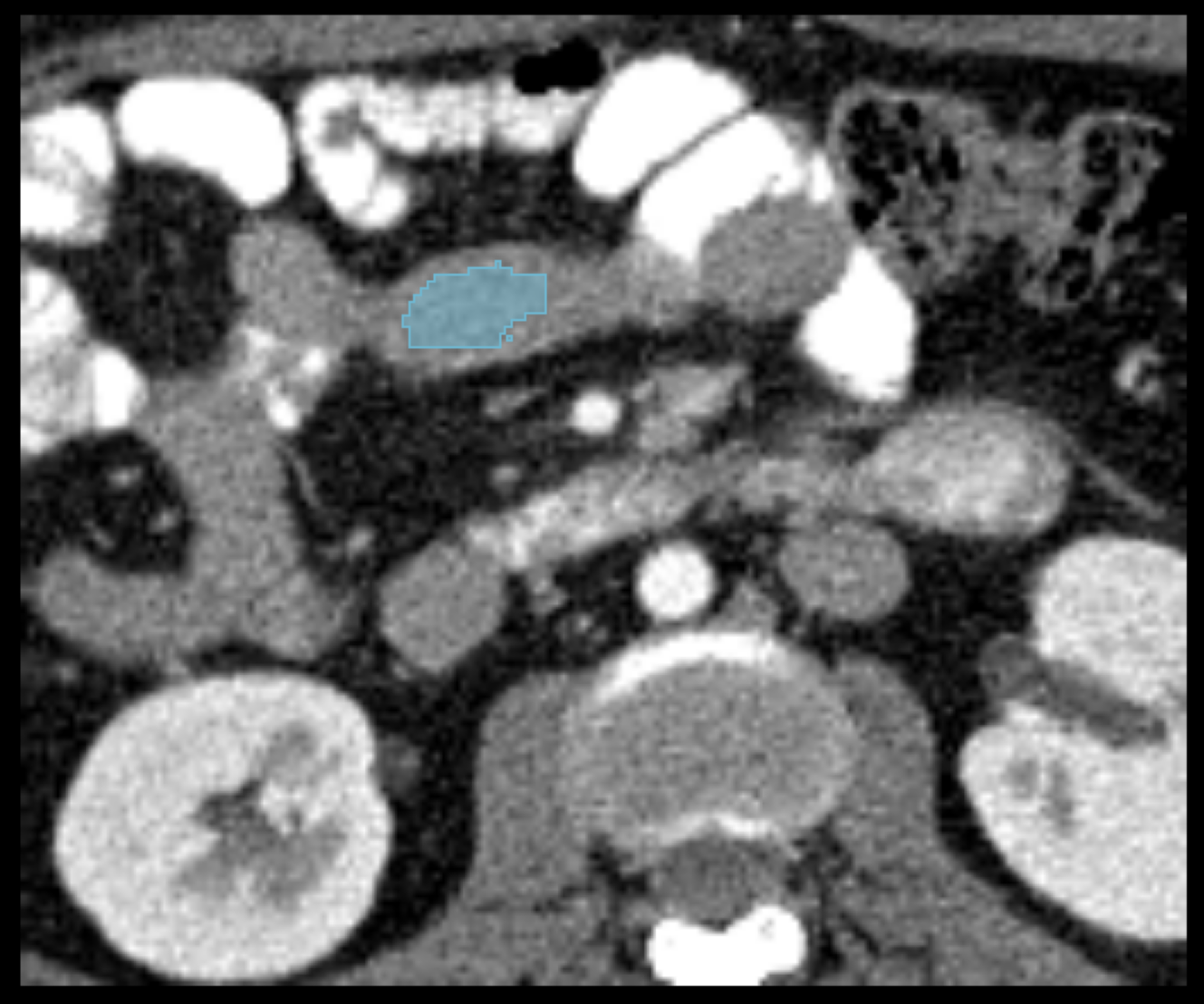}
             \includegraphics[width=\textwidth]{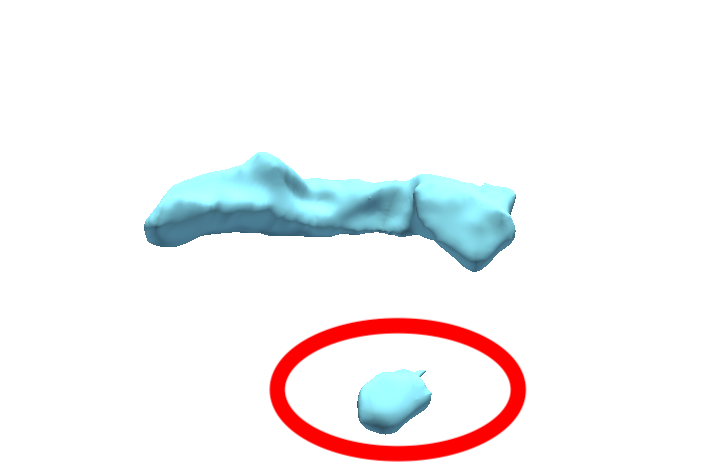}
             \caption*{SASSNet~\cite{li2020shape}}
         \end{subfigure}
         \begin{subfigure}[b]{.192\textwidth}
             \includegraphics[width=\textwidth]{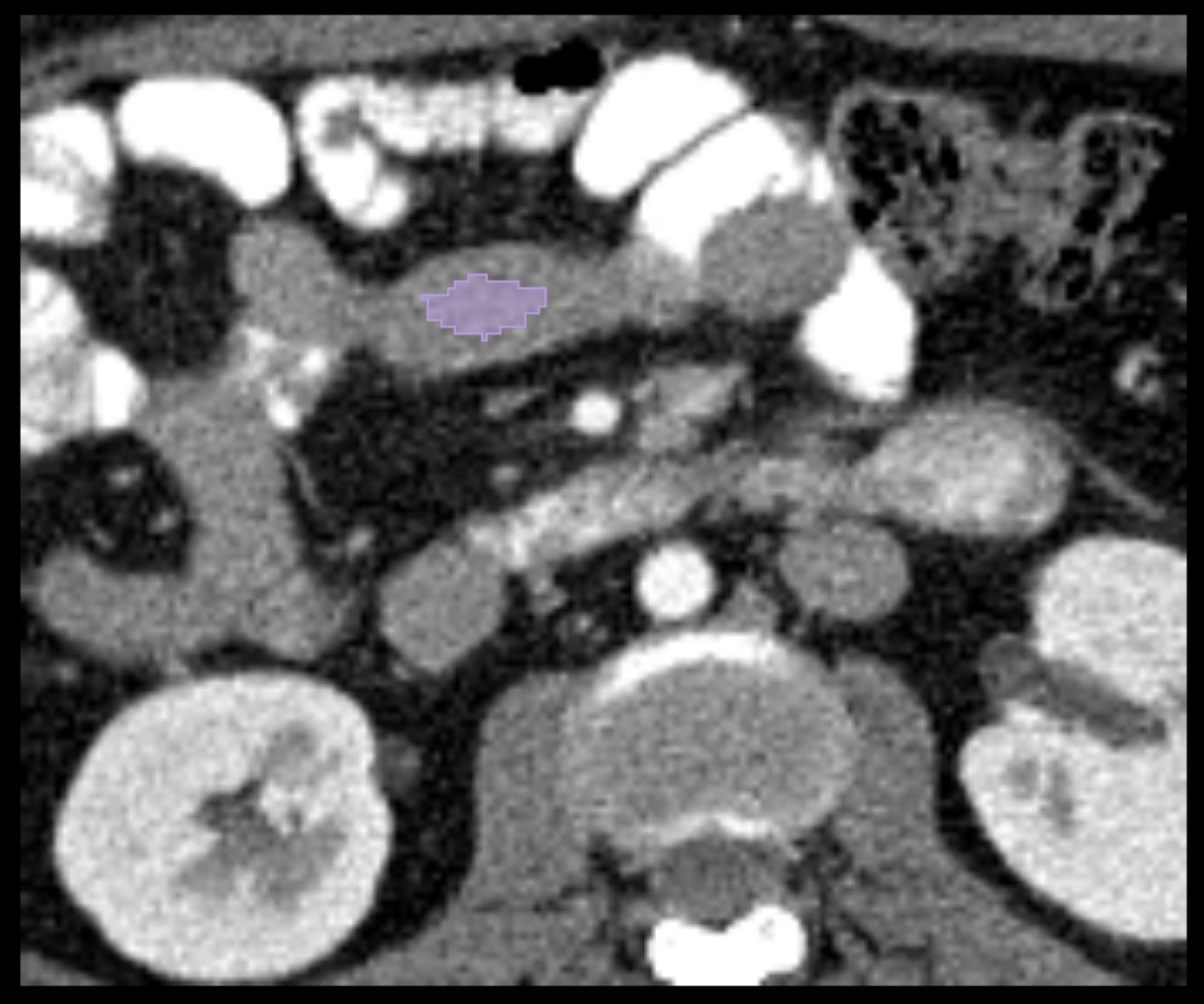}
             \includegraphics[width=\textwidth]{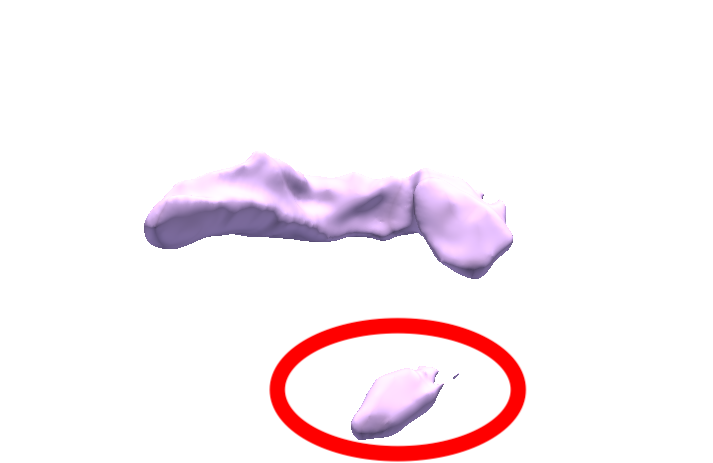}
             \caption*{UA-MT~\cite{yu2019uncertainty}}
         \end{subfigure}
         \begin{subfigure}[b]{.192\textwidth}
             \includegraphics[width=\textwidth]{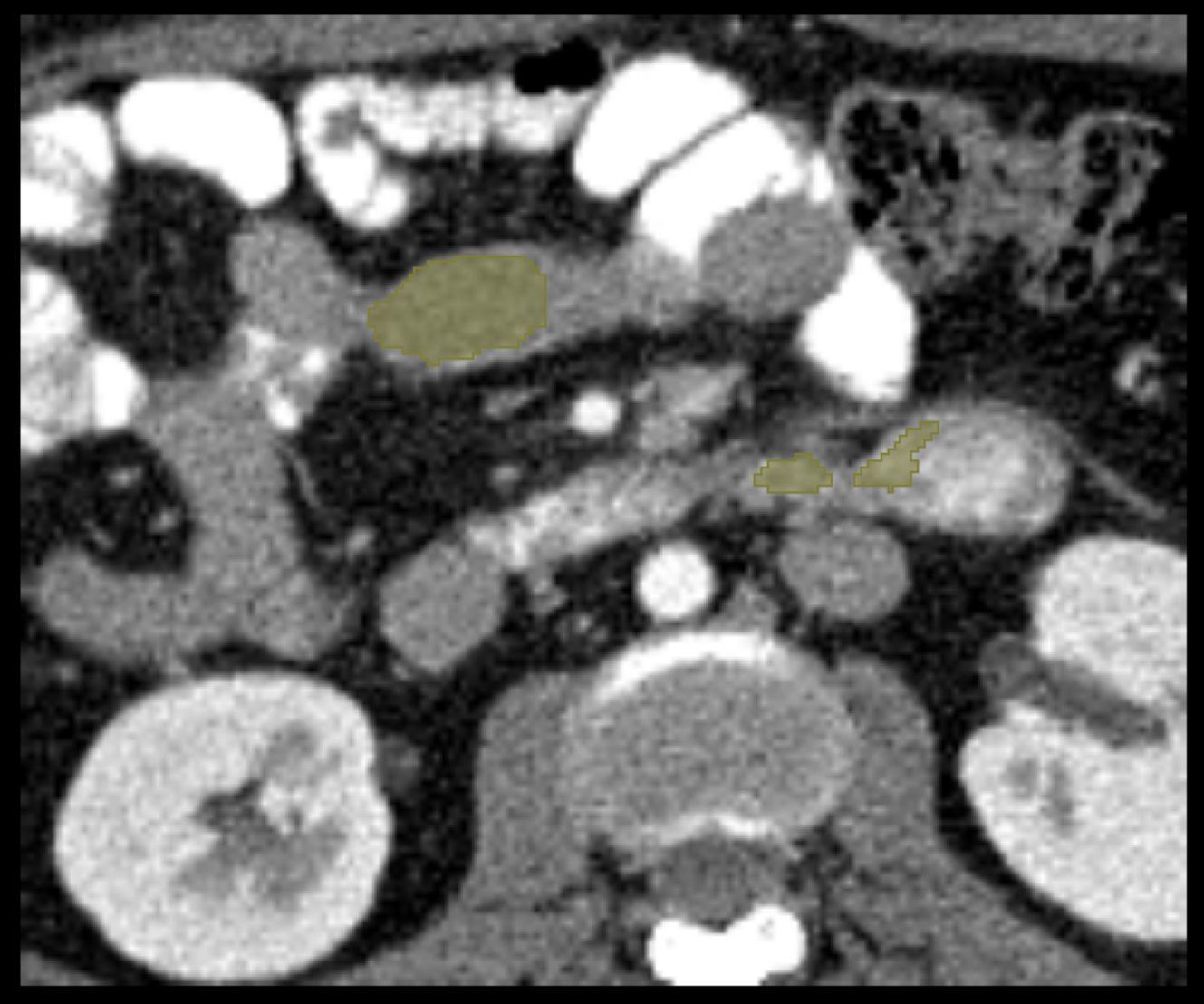}
             \includegraphics[width=\textwidth]{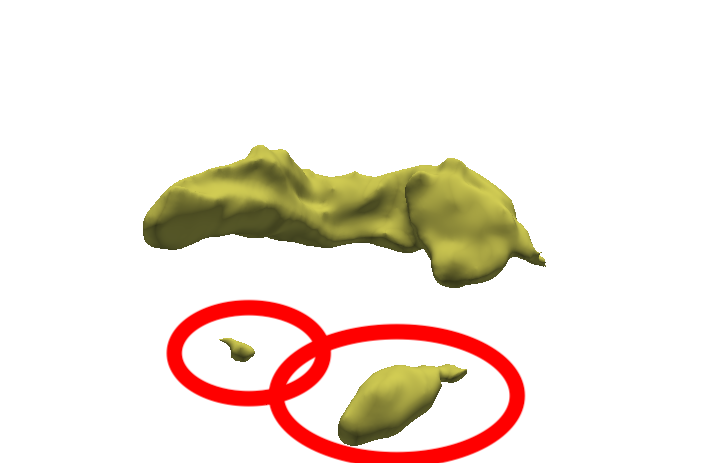}
             \caption*{Sup. Only}
         \end{subfigure}
         \begin{subfigure}[b]{.192\textwidth}
             \includegraphics[width=\textwidth]{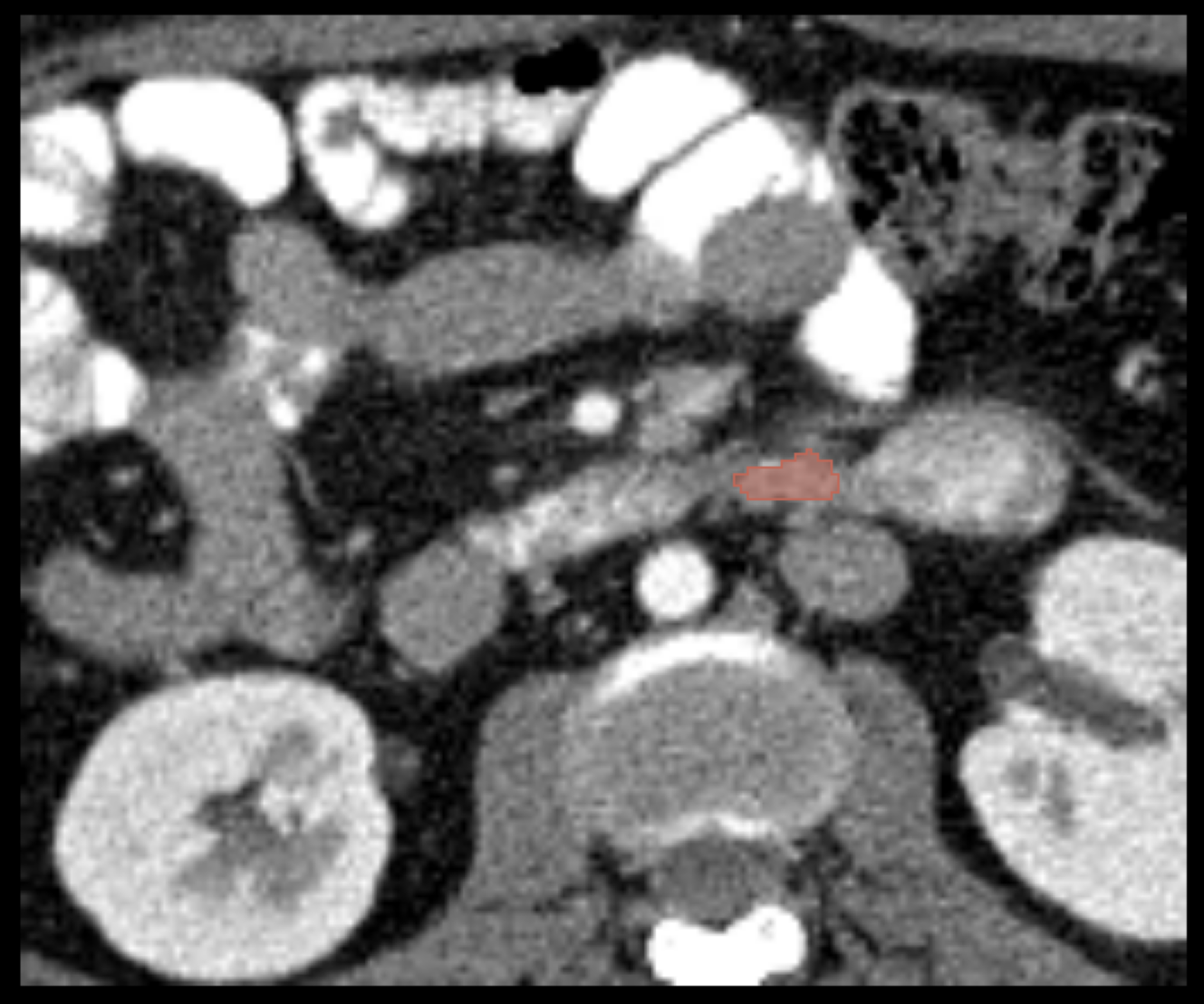}
             \includegraphics[width=\textwidth]{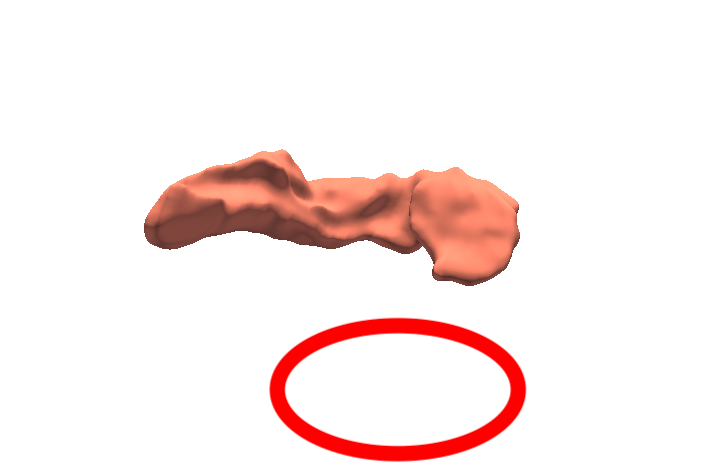}
             \caption*{Ours}
         \end{subfigure}
         \caption{Results on the $12$-labelled data partition protocol of the Pancreas-CT dataset}
     \end{subfigure}
         \begin{subfigure}[b]{.192\textwidth}
             \includegraphics[width=\textwidth]{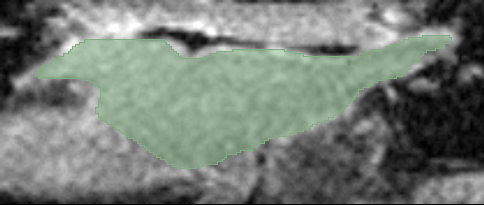}
             \includegraphics[width=\textwidth]{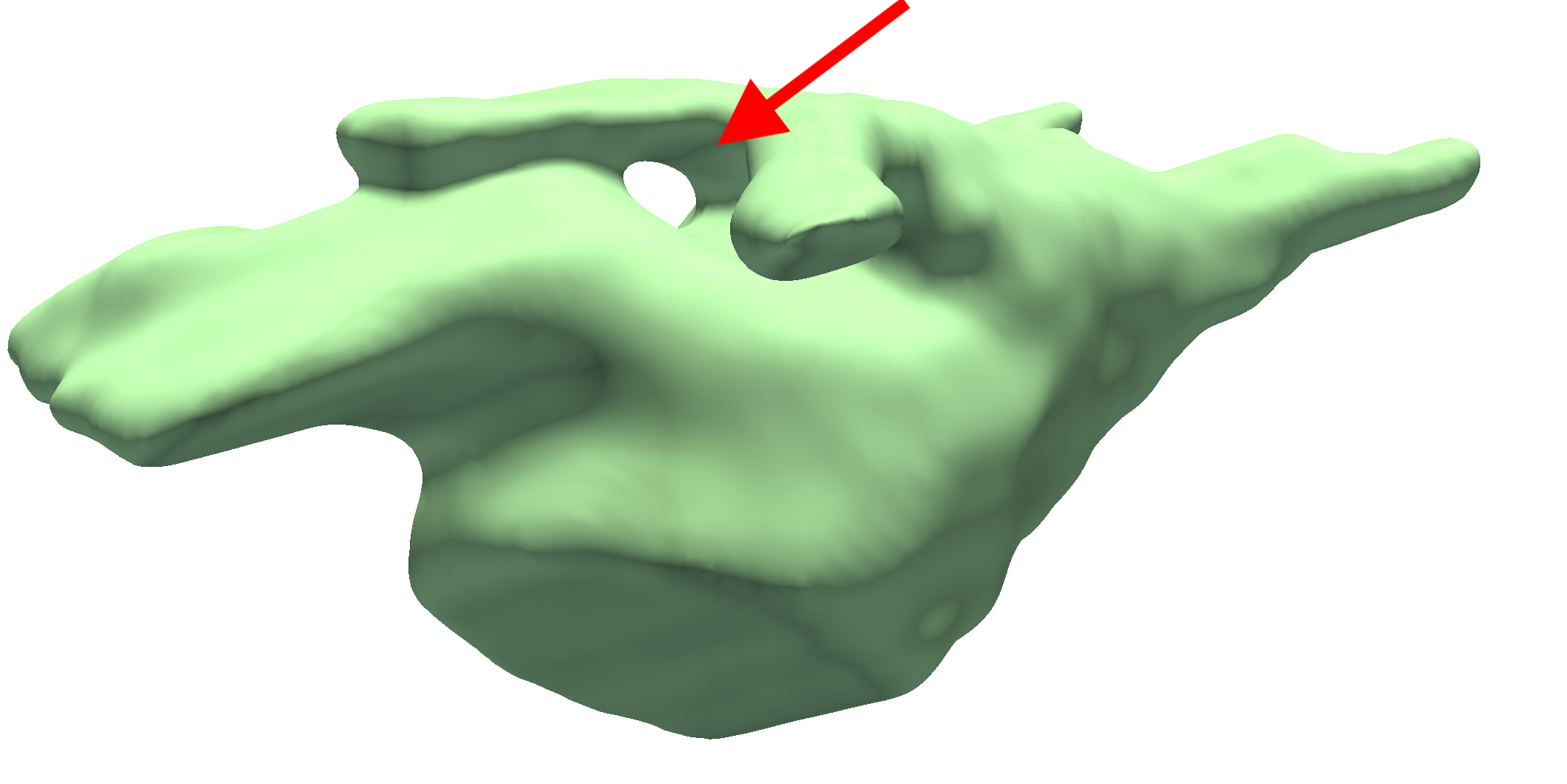}
             \caption*{Ground truth}
         \end{subfigure}
         \begin{subfigure}[b]{.192\textwidth}
             \includegraphics[width=\textwidth]{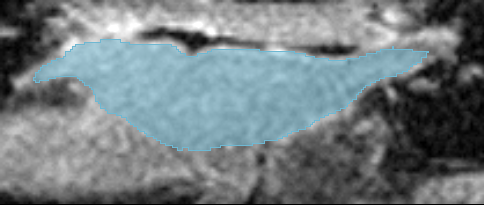}
             \includegraphics[width=\textwidth]{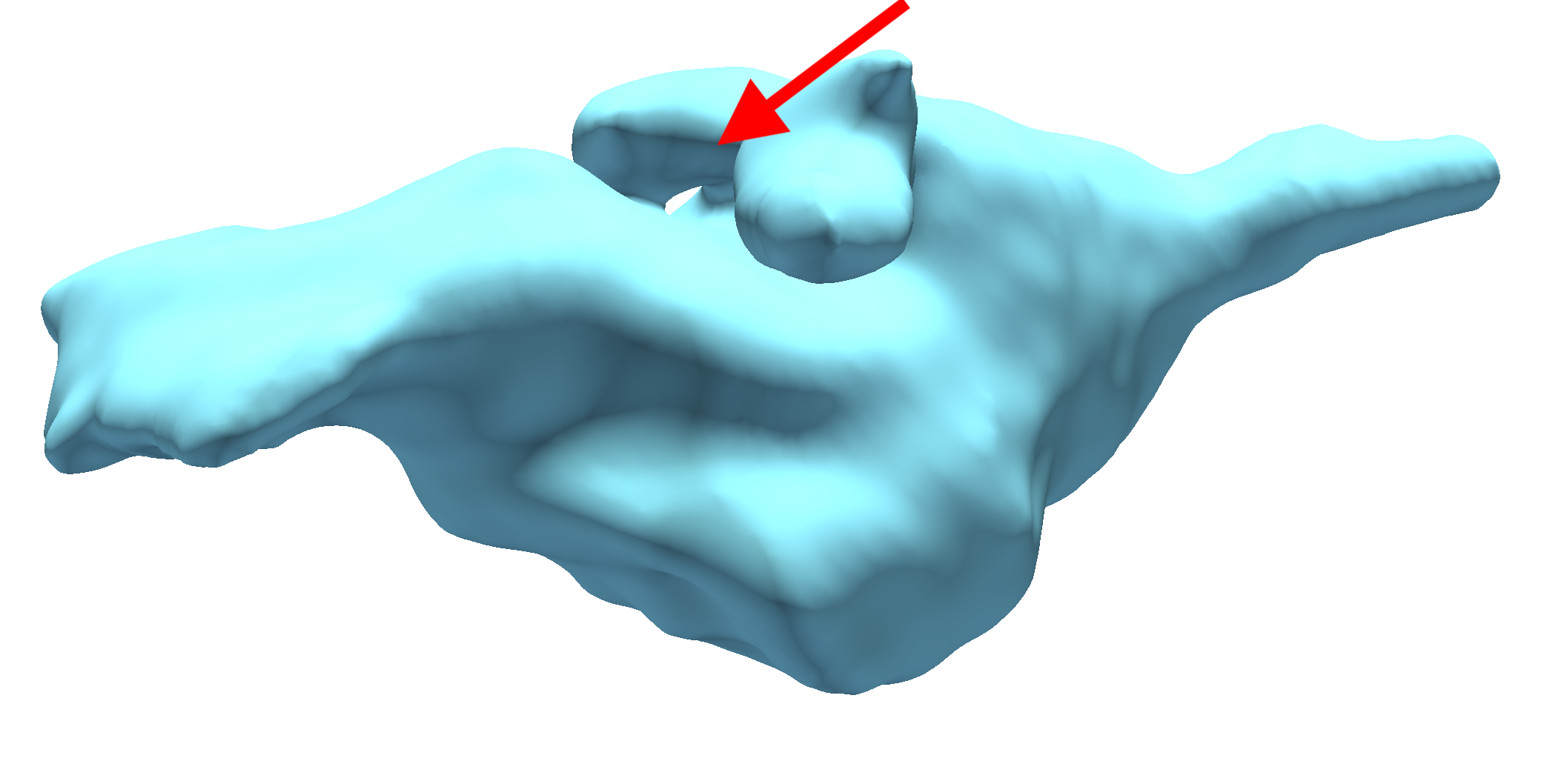}
             \caption*{SASSNet~\cite{li2020shape}}
         \end{subfigure}
         \begin{subfigure}[b]{.192\textwidth}
             \includegraphics[width=\textwidth]{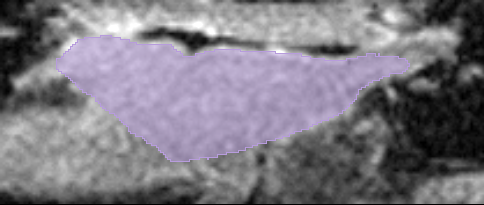}
             \includegraphics[width=\textwidth]{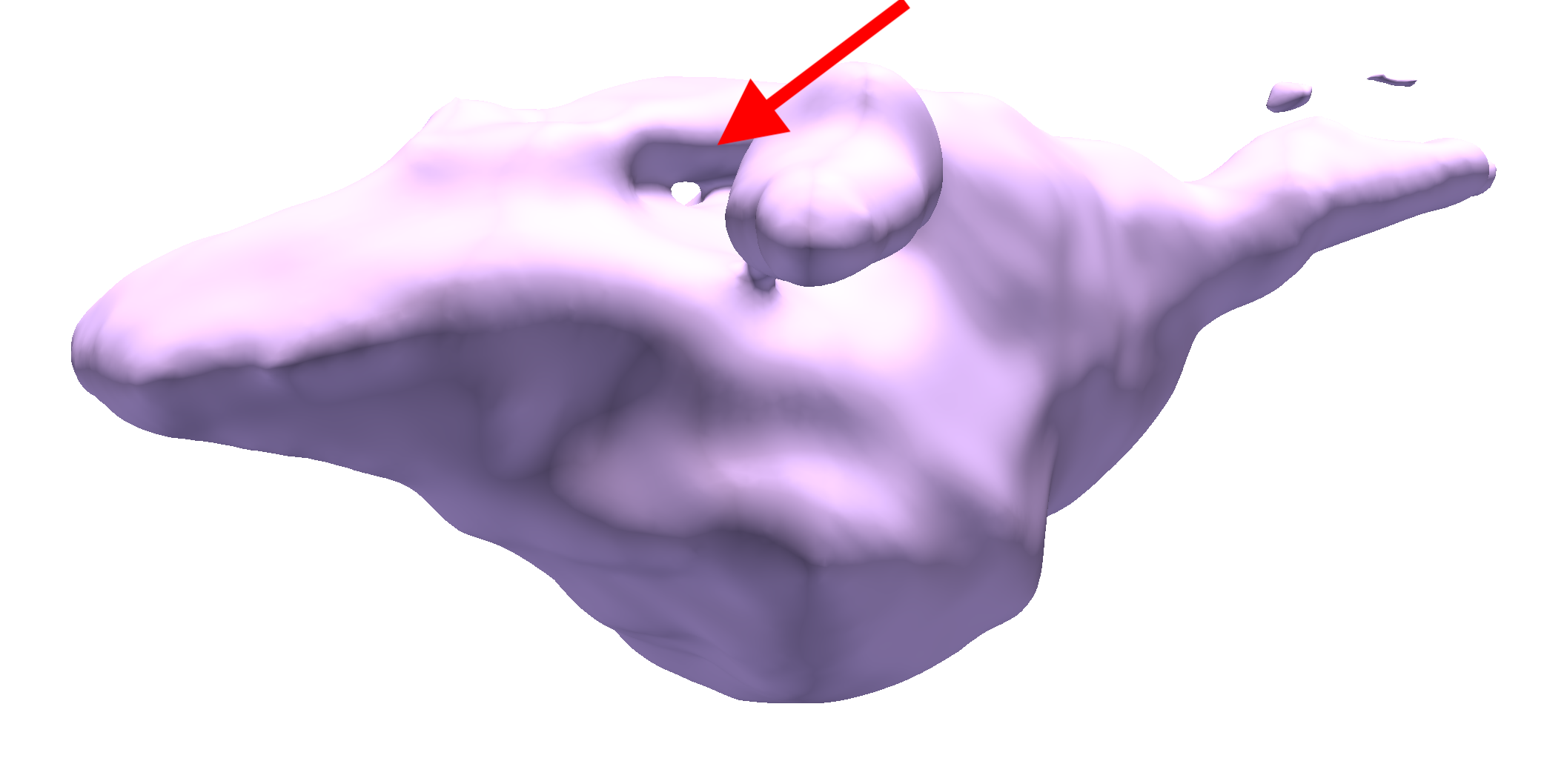}
             \caption*{UA-MT~\cite{yu2019uncertainty}}
         \end{subfigure}
         \begin{subfigure}[b]{.192\textwidth}
             \includegraphics[width=\textwidth]{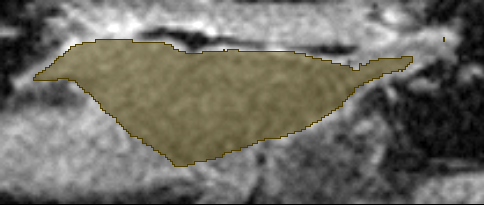}
             \includegraphics[width=\textwidth]{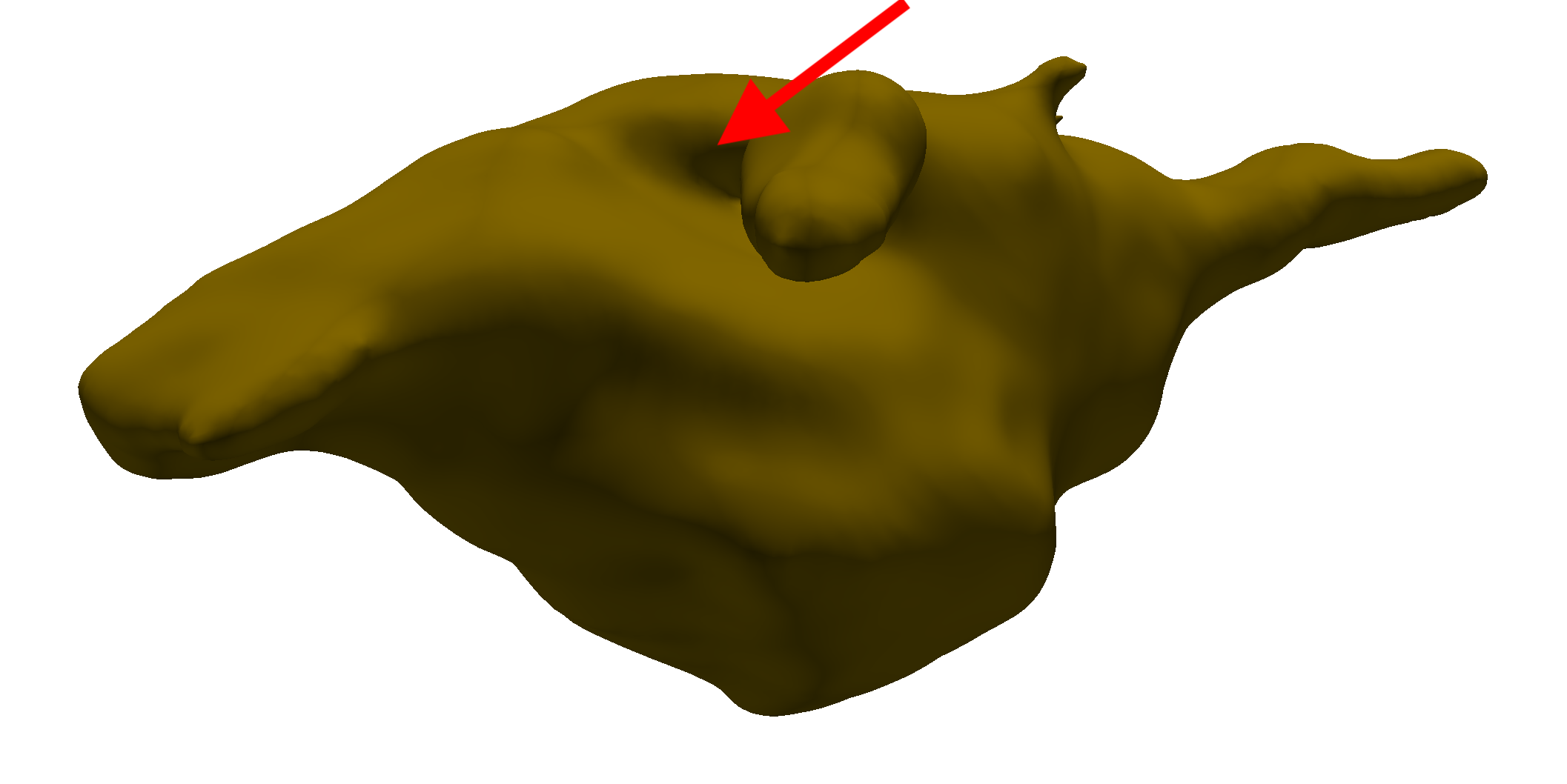}
             \caption*{Sup. Only}
         \end{subfigure}
         \begin{subfigure}[b]{.192\textwidth}
             \includegraphics[width=\textwidth]{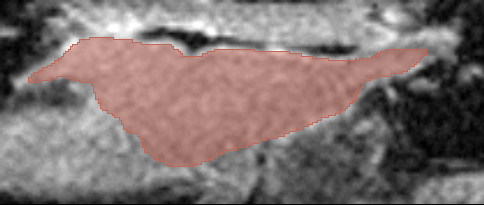}
             \includegraphics[width=\textwidth]{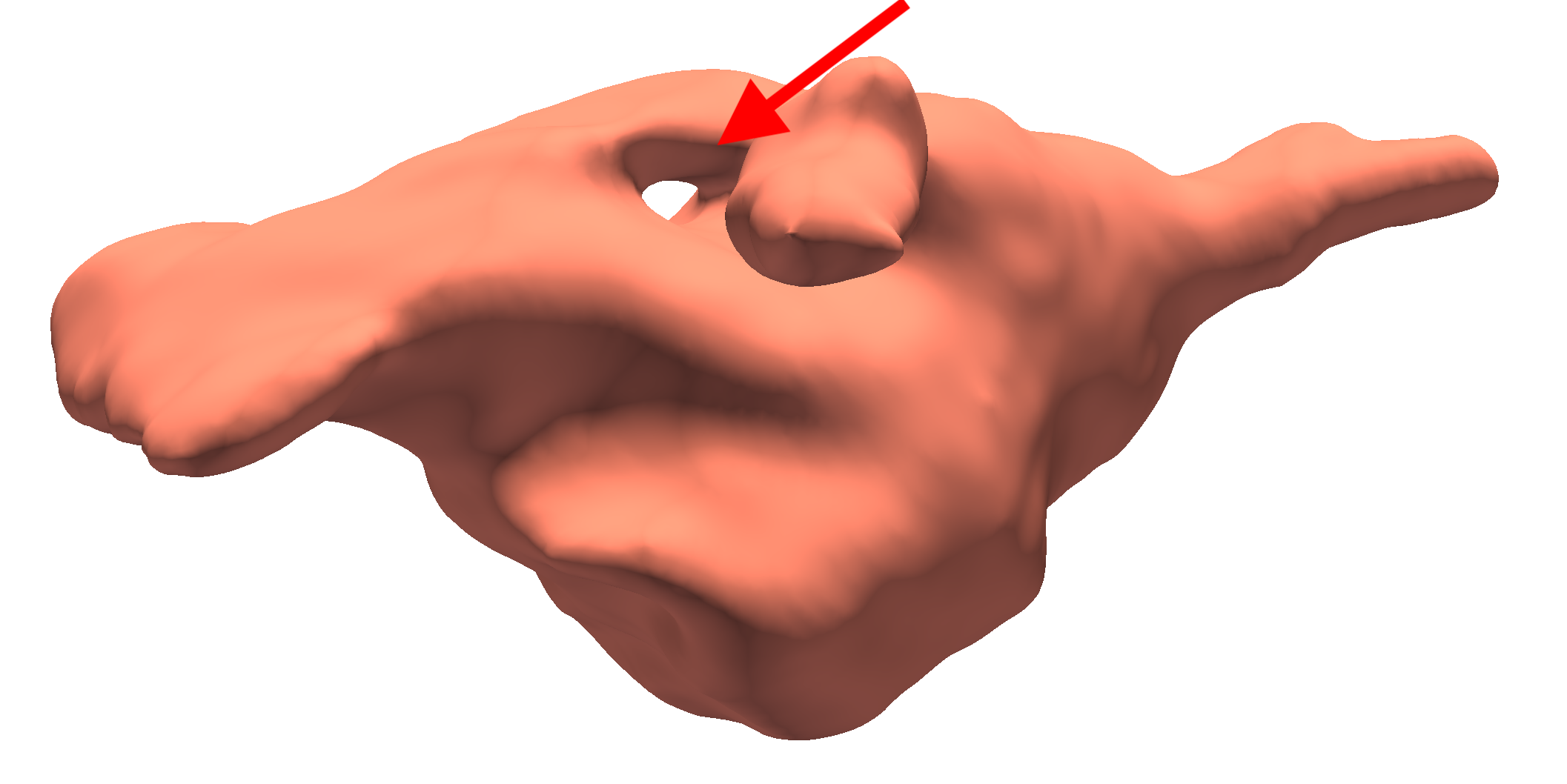}
             \caption*{Ours}
         \end{subfigure}
         \caption{Results on the $16$-labelled data partition protocol of the LA dataset}
     \end{subfigure}
     \begin{subfigure}[b]{1.\textwidth}
         \centering    
            \begin{subfigure}[b]{.192\textwidth}
             \includegraphics[width=\textwidth]{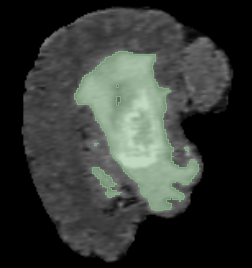}
             \includegraphics[width=\textwidth]{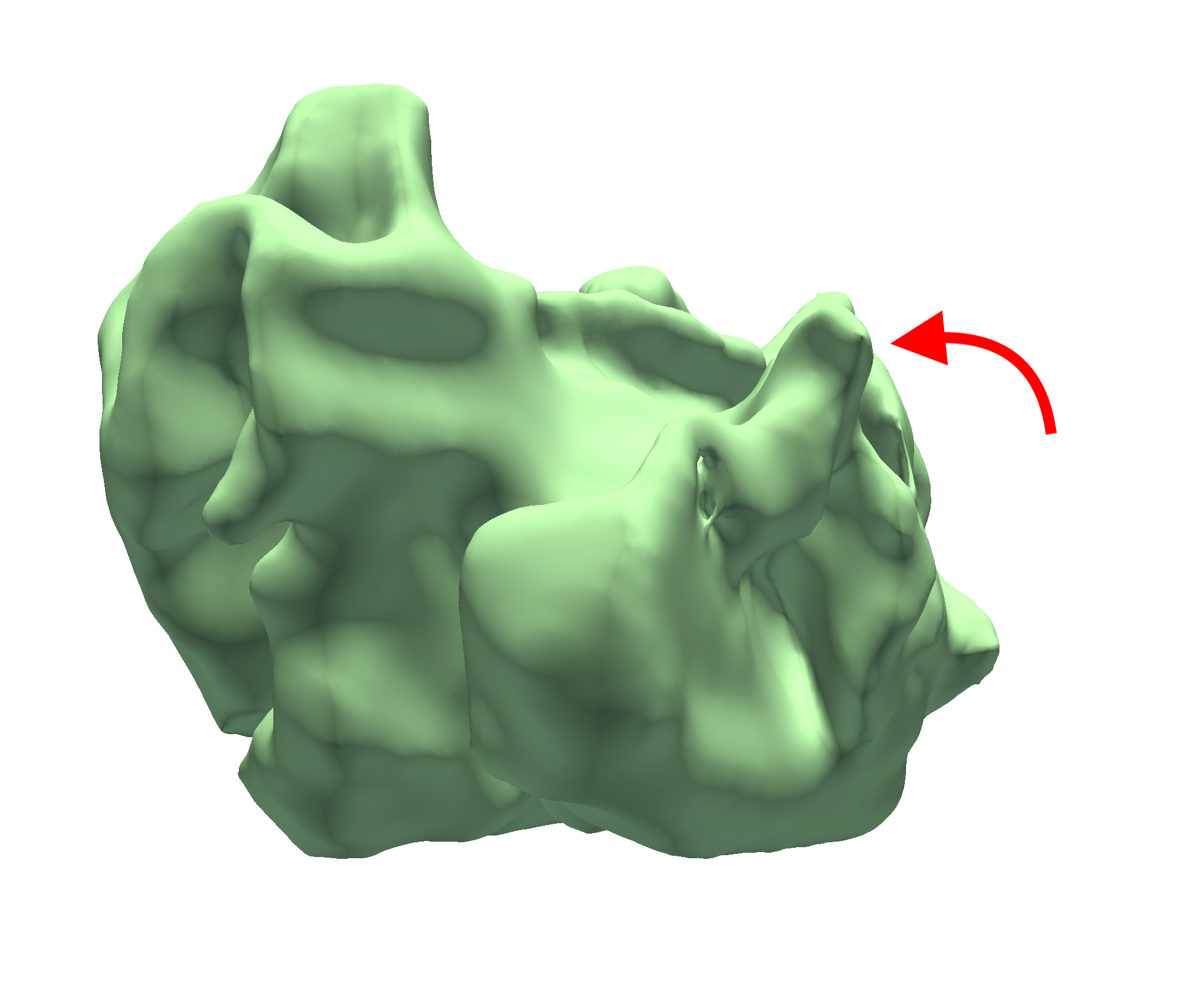}
             \caption*{Ground truth}
         \end{subfigure}
         \begin{subfigure}[b]{.192\textwidth}
             \includegraphics[width=\textwidth]{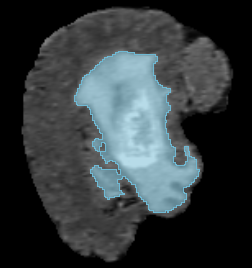}
             \includegraphics[width=\textwidth]{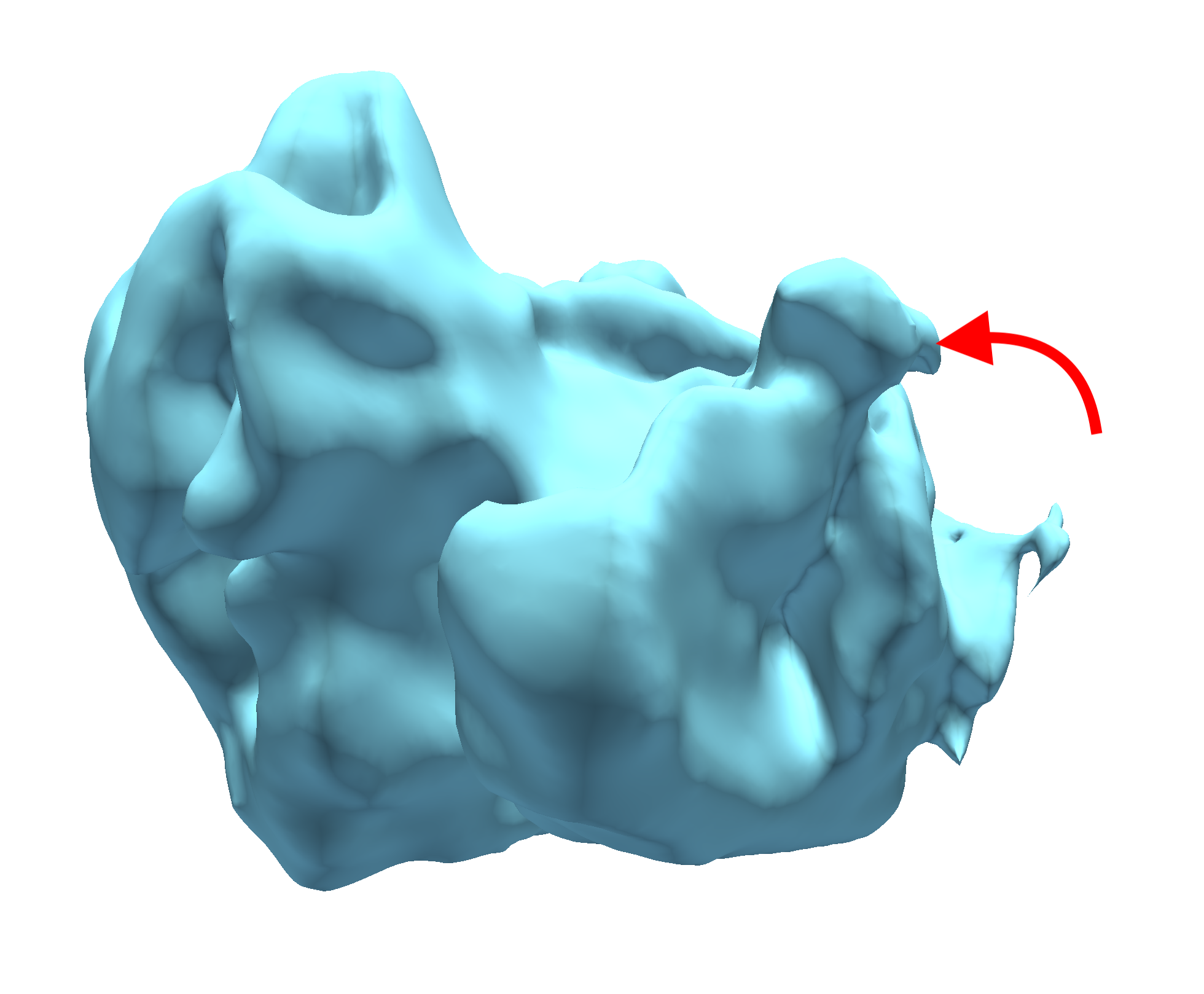}
             \caption*{SASSNet~\cite{li2020shape}}
         \end{subfigure}
         \begin{subfigure}[b]{.192\textwidth}
             \includegraphics[width=\textwidth]{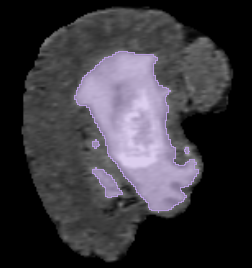}
             \includegraphics[width=\textwidth]{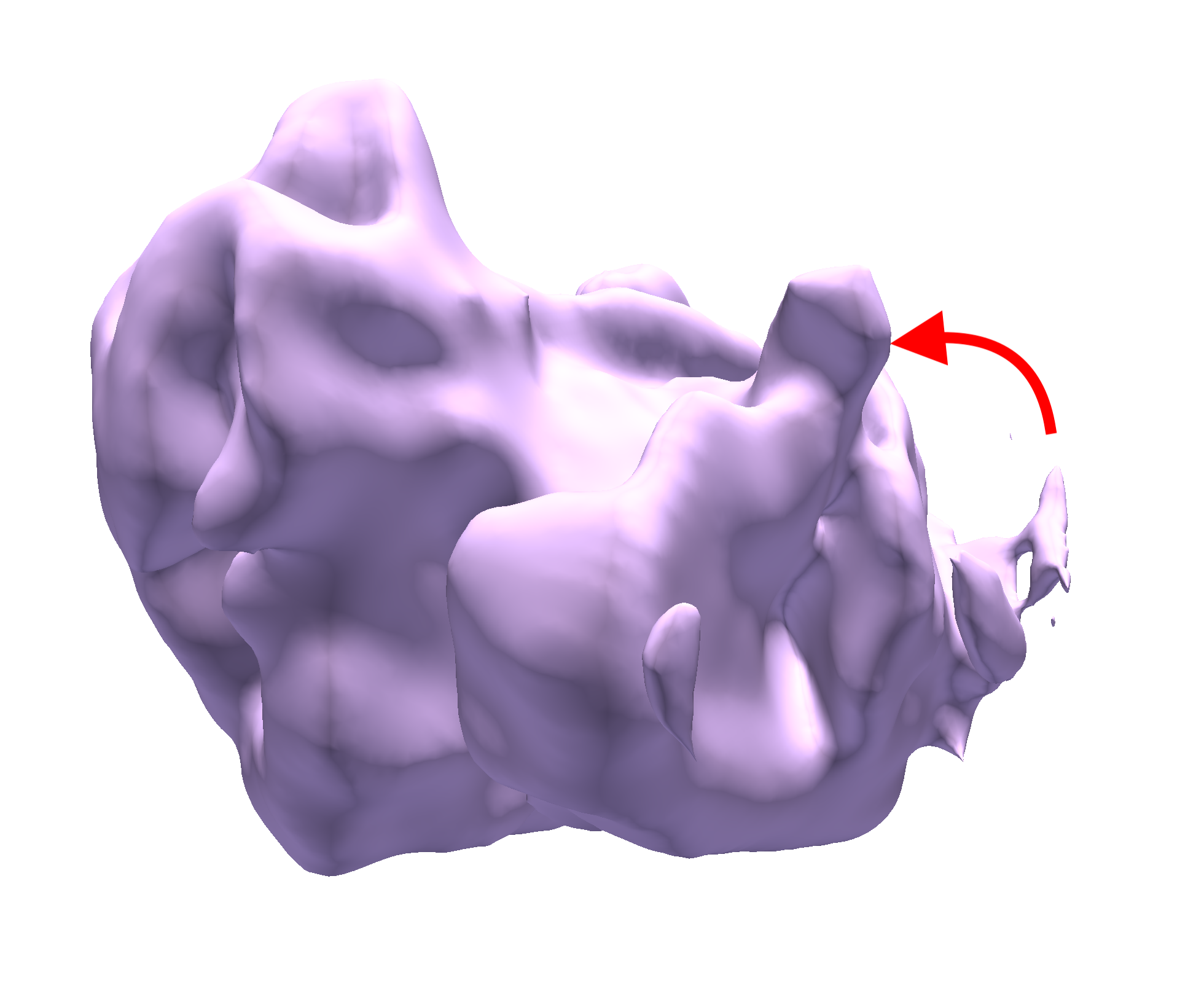}
             \caption*{UA-MT~\cite{yu2019uncertainty}}
         \end{subfigure}
         \begin{subfigure}[b]{.192\textwidth}
             \includegraphics[width=\textwidth]{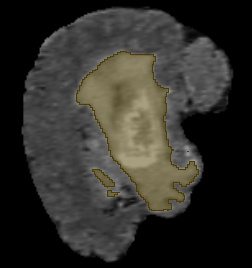}
             \includegraphics[width=\textwidth]{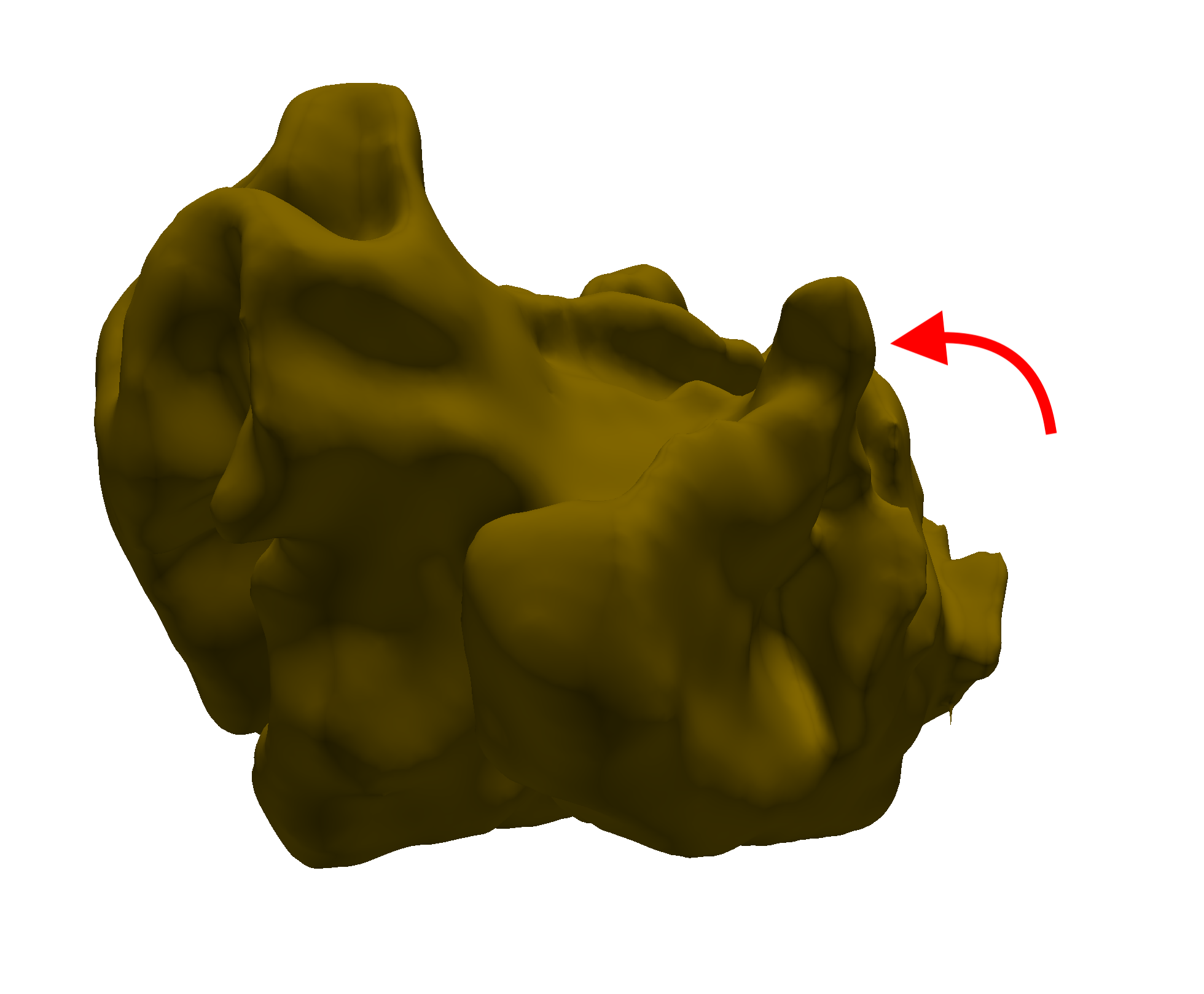}
             \caption*{Sup. Only}
         \end{subfigure}
         \begin{subfigure}[b]{.192\textwidth}
             \includegraphics[width=\textwidth]{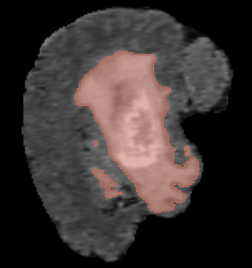}
             \includegraphics[width=\textwidth]{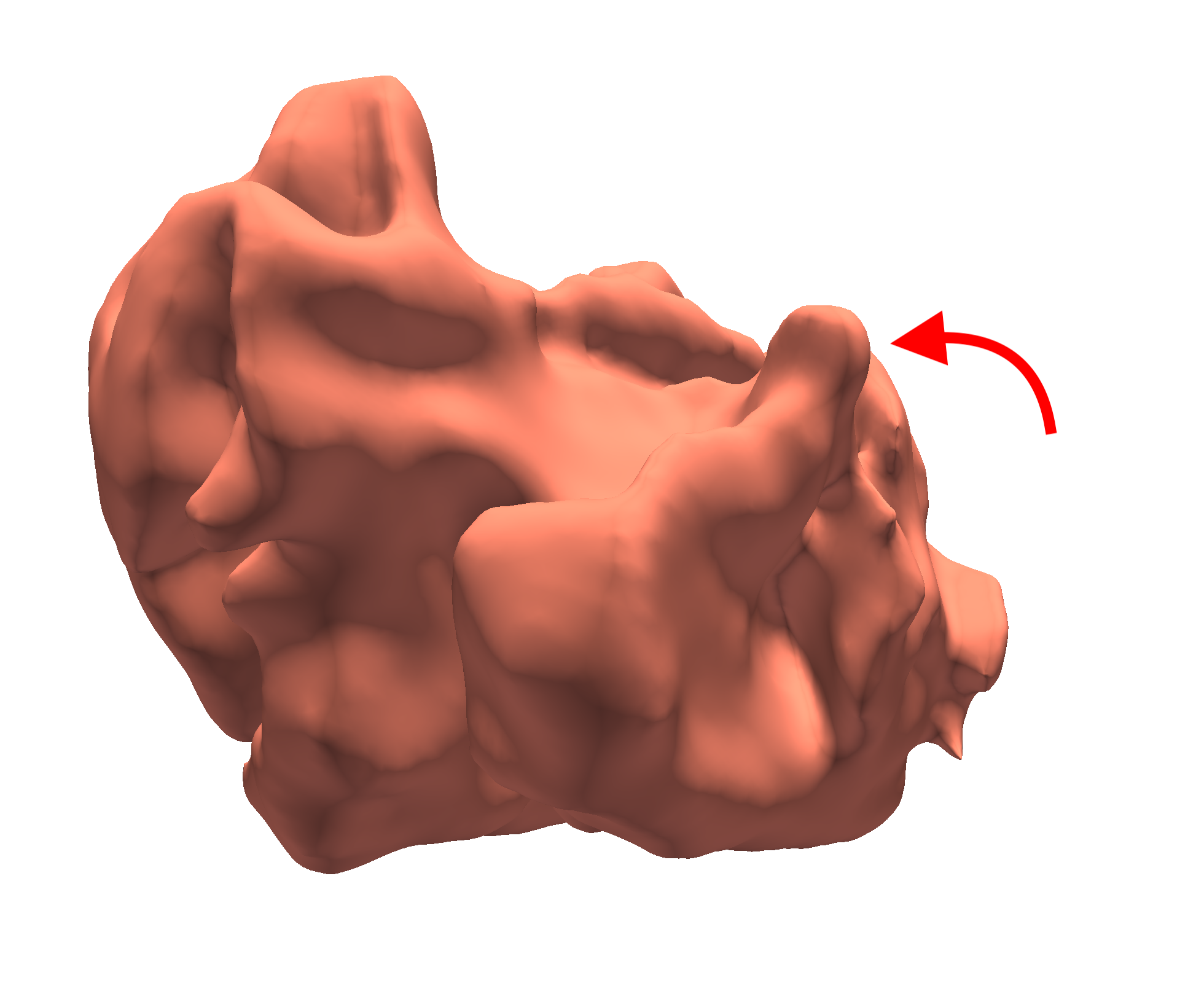}
             \caption*{Ours}
         \end{subfigure}
         \caption{Results on the $25$-labelled data partition protocol of the BraTS19 dataset}
     \end{subfigure}

    \caption{\textbf{Qualitative segmentation results.} Examples of segmentation results produced by SASSNet~\cite{li2020shape}, UA-MT~\cite{yu2019uncertainty}, fully supervised (Sup. Only), and our TraCoCo on \textit{Pancreas} (with 12-label protocol -- first row), \textit{LA} (with 16-label protocol -- second row),  and \textit{BraTS19} (with 25-label protocol -- third row). Please notice the red arrow and ellipse markers that indicate the regions to focus on in the comparison between methods. \label{fig: visual_results}}
    \vspace{-15pt}
\end{figure*}

Fig.~\ref{fig: visual_results} shows the results by SASSNet~\cite{li2020shape}, UA-MT~\cite{yu2019uncertainty}, fully supervised, and our TraCoCo on Pancreas (with 12-label protocol -- first row), LA (with 16-label protocol -- second row), and BraTS19 (with 25-label protocol -- third row). It can be noted that in all cases, the visual results from our method appears to be closer to the ground truth than the results from competing approaches, particularly when considering the regions marked with a red arrow or ellipse.

\section{Discussion}
\subsection{Improvements over Contrastive Learning Method}
Our proposed TraCoCo shares a similar motivation with the contrastive learning based method, CAC~\cite{lai2021semi}, which is proposed to mitigate the overfitting of \textbf{contextual patterns} in the natural images with the following characteristics: \textbf{1)} the contrastive learning used by CAC~\cite{lai2021semi} relies on  selected positive/negative dense embeddings, but the computationally complex sampling of such embeddings~\cite{wang2021exploring, wang2021dense} only allows CAC to work with fairly limited sets of positive/negative pairs~\cite{lai2021semi} that do not cover the entire intersection region;
\textbf{2)} such contrastive learning is performed in the feature level of the unlabelled data, where instead of optimising the entire model, only the encoder is updated from the unlabelled data, while the decoder (or classifier) remains fixed, leading to potentially sub-optimal training results; and 
\textbf{3)} the positive/negative pair selection is guided by the pseudo labels produced by the classifier, which is only optimised with the small set of labelled data that can cause it to produce noisy pseudo labels, leading to confirmation bias. 
\begin{figure}[t!]
    \begin{subfigure}[b]{.24\textwidth}
    \includegraphics[width=\textwidth]{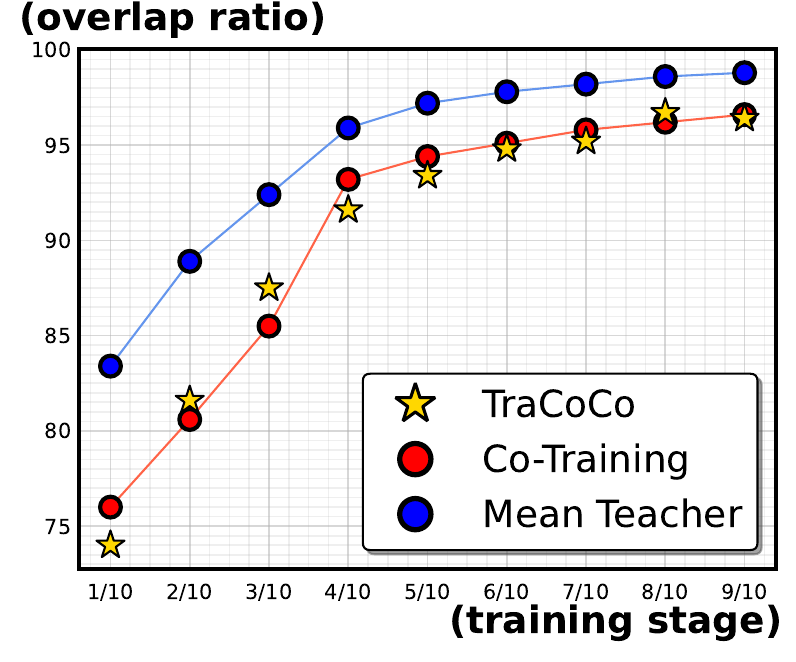}
    \caption{Overlap ratio. \label{fig:diff_over}}    
    \end{subfigure} \hfill
    \begin{subfigure}[b]{.24\textwidth}
    \includegraphics[width=\textwidth]{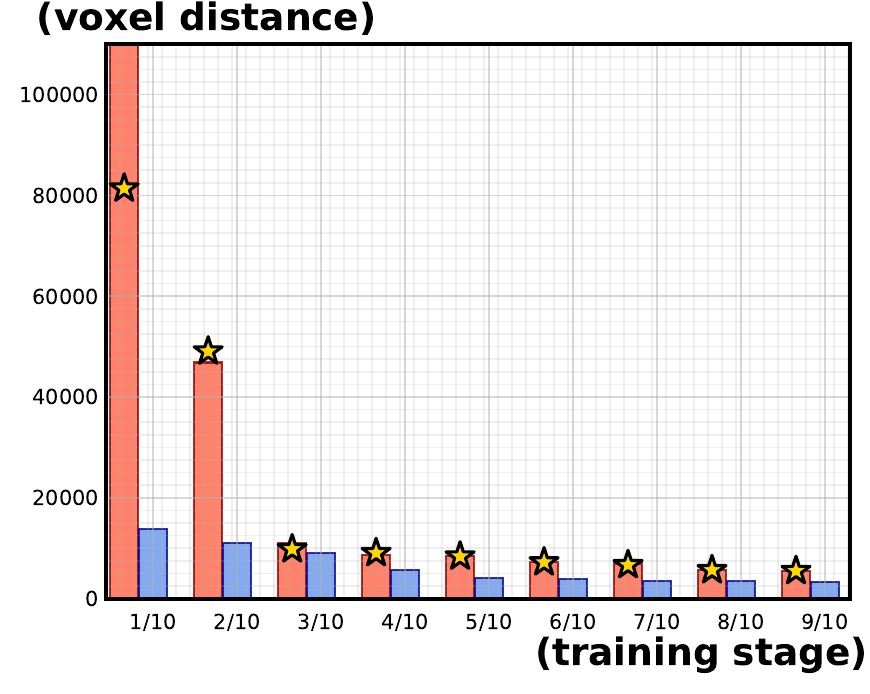}
    \caption{Voxel distance. \label{fig:diff_dist}}    
    \end{subfigure}
    \caption{\textbf{Divergence comparison} between \textcolor{darkgold!60}{TraCoCo}, \textcolor{red}{Co-Training} and \textcolor{blue}{Mean Teacher} frameworks on BraTS19 dataset, where we calculate the average overlap ratio (a) and voxel distance (b) at different training stages.}\label{fig:diff} 
    \vspace{-10pt}
\end{figure}
In our experiments, we have found that those drawbacks are more pronounced in the challenging 3D medical dataset, leading to limited generalisation. Our proposed TraCoCo addresses the issues in CAC, as follows: \textbf{1)} our consistency learning loss functions work directly in the pixel space, without requiring any type of complex sample selection process, reducing the computation cost observed in CAC and allowing our loss functions to work with the entire overlapping region that can provide a more complete optimisation; \textbf{2)} the two consistency learning losses of TraCoCo are designed to train the entire model, which better explore the training signals from the unlabelled data; and \textbf{3)} our CRC loss explores network perturbation via co-training, reducing the risk associated with confirmation bias in the semi-supervised training, where the confidence threshold for positive and negative learning further enhances the robustness to potential noisy pseudo labels. In Tab.~\ref{tab:ablation_cac}, we provide an ablation study that shows that even though co-training and 3D-CutMix introduce some improvement, the contributions from TraCoCo are also substantial. 
We begin with CAC~\cite{lai2021semi}, followed by CAC with co-training and CutMix \textcolor{gray}{(Co)}, which has been proposed in CPS~\cite{chen2021-CPS}. Then, we present the proposed TraCoCo, which is 
our Translation Consistency (TraCo) with co-training and CutMix \textcolor{gray}{(Co)}.
Notably, our TraCo \textcolor{gray}{(Co)} shows better results on both LA~\cite{xiong2021global} and BraTS19~\cite{menze2014multimodal} datasets, improving Dice by $1.01\%$ and $1.26\%$ in the 8 and 25 partition protocols, respectively. Such results confirm that TraCoCo's improvement cannot be explained solely by co-training and CutMix.

\subsection{Divergence Comparison between Co-training, TraCoCo, and Mean-teacher}
As depicted in Fig.~\ref{fig:diff}, we measure the similarities of the pseudo labels for unlabeled data, comparing the results from the models in the Co-Training, TraCoCo, and Mean Teacher frameworks on the BraTS19 dataset. 
Such measurements are averaged over an epoch and calculated at different stages through the training.  
In Fig.~\ref{fig:diff_over}, we calculate the voxel-wise overlap ratio of the foreground results from two networks, with $\text{overlap ratio} = \frac{\text{\# voxels with same segmentation results}}{\text{all voxels}}$ following~\cite{chen2021-CPS}, and we calculate the voxel-wise $L1$ distance in Fig.~\ref{fig:diff_dist}.
We find that TraCoCo has slightly lower overlap ratio and larger voxel distance compared to Co-Training,
and both Co-Training and TraCoCo exhibit greater diversity than Mean Teacher throughout the entire training process. For instance, in the late training stages (9/10), Mean Teacher achieves a 98.8\% overlap ratio and 3302 voxel distance, while Co-Training attains 96.6\% and 5453 and TraCoCo reaches 96.0\% and 5526.

\begin{table}[t!]
\centering
\renewcommand{\arraystretch}{1.2}
\caption{\textbf{Comparison of TraCoCo using different architectures} on BraTS19 dataset. The * denotes we predict 2D slices and stack them together to reconstruct the 3D prediction for evaluation, following~\cite{chen2021transunet}.}\label{tab:other_arch}
\resizebox{\linewidth}{!}{\begin{tabular}{!{\vrule width 1pt}c!{\vrule width 1pt}c|c|c|c|c|c!{\vrule width 1pt}}
\specialrule{1pt}{0pt}{0pt}
\multirow{2}{*}{backbone} & \multicolumn{2}{c|}{\# scan used} & \multicolumn{4}{c!{\vrule width 1pt}}{measures}  \\
\cline{2-7}
                          & labelled       & unlabelled      & Dice  & Jaccard & ASD  & 95HD \\
\specialrule{1pt}{0pt}{0pt}
3DUNet~\cite{cciccek20163d}                    & 25             & 225             & 85.71 & 76.39   & 2.27 & 9.20 \\
3DnnUNet~\cite{isensee2021nnu}                  & 25             & 225             & 86.94 & 77.18   & 1.96 & 7.63 \\
TransUNet*~\cite{chen2021transunet}                 & 25             & 225             & 87.13 & 77.36   & 1.57 & 6.06 \\
\specialrule{1pt}{0pt}{0pt}
\end{tabular}}
\end{table}
\vspace{-15pt}
\subsection{Results with Other Backbones}
As demonstrated in  Tab.~\ref{tab:other_arch}, our approach can be integrated into other architectures, including 3DnnUNet~\cite{isensee2021nnu} and TransUNet~\cite{chen2021transunet}. 
The superiority of 3DnnUNet and  TransUNet, compared to 3DUnet, allows Dice improvements of 1.23\% and 1.42\%, respectively.

\begin{table*}[t!]
\caption{\textbf{Analysis of the proposed CRC loss $\ell_{crc}$.} We compare the gradients of $\ell_{mse}$, $\ell_{ce}$, and $\ell_{crc}$ for different levels of confidence for the pseudo label $\tilde{\mathbf{y}}$, and agreeing and disagreeing predictions of $\hat{\mathbf{y}}$. Compared to the other two losses, $\ell_{crc}$ exhibits fast convergence for confident $\tilde{\mathbf{y}}$ and noisy pseudo-label robustness for unconfident $\tilde{\mathbf{y}}$.} \label{tab:analysis}
\centering
\renewcommand{\arraystretch}{1.5}
\resizebox{.75\textwidth}{!}{\begin{tabular}{!{\vrule width 1pt}c|c!{\vrule width 1pt}l!{\vrule width 1pt}}
\specialrule{1pt}{0pt}{0pt}
 Pseudo Label ($\tilde{\mathbf{y}}$)                         & Prediction  ($\hat{\mathbf{y}}$)              & \multicolumn{1}{c!{\vrule width 1pt}}{Gradient ($\nabla$)} \\ 
\hline
\multirow{6}{*}{\makecell{$\tilde{\mathbf{y}} = [0.55, 0.45]$\\[1ex] \textbf{\ul{Unconfident}}}}       & \multirow{3}{*}{\makecell{$\hat{\mathbf{y}}=[0.2, 0.8]$\\[1ex] \textbf{\ul{Disagree with $\tilde{\mathbf{y}}$}}}} &    $\nabla\ell_{\textbf{mse}} = -\frac{2\times (0.35 + 0.35)}{2} = -0.35$      \\
                                   &                                                       &   $\nabla\ell_{\textbf{ce}} = - \frac{1}{0.2} = - 5$       \\
                                    &                   &     $\nabla\ell_{\textbf{crc}} = 0 \times \frac{1}{0.2} - 0 \times \frac{1}{(1-0.8)} = 0$     \\
\cline{2-3}
                                    & \multirow{3}{*}{\makecell{$\hat{\mathbf{y}}=[0.8, 0.2]$\\[1ex] \textbf{\ul{Agree with $\tilde{\mathbf{y}}$}}}} &     $\nabla\ell_{\textbf{mse}} = -\frac{2\times (0.25 + 0.25)}{2} = -0.25$     \\
                                   &                                                 &   $\nabla\ell_{\textbf{ce}} = \frac{1}{0.8} =  - 1.25$       \\
                                   &                                                      &   $\nabla\ell_{\textbf{crc}} = 0 \times \frac{1}{0.2} - 0 \times \frac{1}{(1-0.8)} = 0$       \\
\cline{1-3}
                                   \multirow{6}{*}{\makecell{$\tilde{\mathbf{y}} = [0.65, 0.35]$ \\[1ex] \textbf{\ul{Mildly Confident}}}}& \multirow{3}{*}{\makecell{$\hat{\mathbf{y}}=[0.2, 0.8]$\\[1ex] \textbf{\ul{Disagree with $\tilde{\mathbf{y}}$}}}} &     $\nabla\ell_{\textbf{mse}} = -\frac{2 \times (0.45 + 0.45)}{2} = -0.45$     \\
                                   &                                                       &       $\nabla\ell_{\textbf{ce}} = -\frac{1}{0.2} = -5$    \\
                                   &                                                       &       $\nabla\ell_{\textbf{crc}} = -0.65 \times \frac{1}{0.2} - 0 \times \frac{1}{(1-0.8)} = -3.25$   \\
\cline{2-3}
                                    & \multirow{3}{*}{\makecell{$\hat{\mathbf{y}}=[0.8, 0.2]$\\[1ex] \textbf{\ul{Agree with $\tilde{\mathbf{y}}$}}}} &     $\nabla\ell_{\textbf{mse}} = -\frac{2 \times (0.15 + 0.15)}{2} = -0.15$     \\
                                   &                                                      &   $\nabla\ell_{\textbf{ce}} = -\frac{1}{0.8} =  -1.25$         \\
                                   &                                                       &   $\nabla\ell_{\textbf{crc}} = -0.65 \times \frac{1}{0.8} - 0 \times \frac{1}{(1-0.8)} =  -0.813 $       \\
\cline{1-3}
                                   \multirow{6}{*}{\makecell{$\tilde{\mathbf{y}} = [0.95, 0.05]$ \\[1ex] \textbf{\ul{Very Confident}}}}    & \multirow{3}{*}{\makecell{$\hat{\mathbf{y}}=[0.2, 0.8]$\\[1ex] \textbf{\ul{Disagree with $\tilde{\mathbf{y}}$}}}} &    $\nabla\ell_{\textbf{mse}} = -\frac{2 \times (0.7 + 0.7)}{2} = -0.7$      \\
                                   &                                                       &      $\nabla\ell_{\textbf{ce}} = -\frac{1}{0.2} = -5$     \\
                                   &                                                       &     $\nabla\ell_{\textbf{crc}} = -0.95 \times \frac{1}{0.2} - 0.95 \times \frac{1}{(1-0.8)} = - 9.5 $     \\
\cline{2-3}
                                   & \multirow{3}{*}{\makecell{$\hat{\mathbf{y}}=[0.8, 0.2]$\\[1ex] \textbf{\ul{Agree with $\tilde{\mathbf{y}}$}}}} &     $\nabla\ell_{\textbf{mse}} = -\frac{2 \times (0.1 + 0.1)}{2} = -0.1$       \\
                                   &                                                       &   $\nabla\ell_{\textbf{ce}} = -\frac{1}{0.8} =  -1.25$       \\
                                   &                                                       &   $\nabla\ell_{\textbf{crc}} = -0.95 \times \frac{1}{0.8} - 0.95 \times \frac{1}{(1-0.2)} = -2.37 $    \\
\specialrule{1pt}{0pt}{0pt}
\end{tabular}}
\end{table*}

\subsection{Analysis of the Proposed CRC Loss $\ell_{crc}(.)$}

The CRC loss $\ell_{crc}$ yields stronger penalties than the MSE loss $\ell_{mse}$ to recognise the foreground objects, but also maintain its robustness to noisy pseudo labels, compared to the cross-entropy loss $\ell_{ce}$. Below we compare the MSE, CE, and CRC losses between the pseudo label $\tilde{\mathbf{y}}$ and prediction $\hat{\mathbf{y}}$, where $\hat{\mathbf{y}}, \tilde{\mathbf{y}} \in [0,1]^C$ represent the respective classification probability distributions with $C$ being the number of classes. 

\textbf{1)} We can derive the gradient for the MSE loss function ($\ell_{mse}$)  as follows:
\begin{equation}
\begin{split}
    \ell_{mse}(\tilde{\mathbf{y}},\hat{\mathbf{y}}) = & \frac{1}{C} \sum_{c=1}^C (\tilde{\mathbf{y}}(c) - \hat{\mathbf{y}}(c))^2  \\
    \nabla \ell_{mse}(\tilde{\mathbf{y}},\hat{\mathbf{y}}) = & \frac{1}{C} \sum_{c=1}^C \frac{\partial (\tilde{\mathbf{y}}(c) - \hat{\mathbf{y}}(c))^2}{\partial \hat{\mathbf{y}}(c)} = -\frac{2}{C} \sum_{c=1}^C (\tilde{\mathbf{y}}(c) - \hat{\mathbf{y}}(c)).
\end{split}    
\end{equation}

\textbf{2)} For the cross-entropy cost function ($\ell_{ce}$), we represent the pseudo-labels as a one-hot vector, with the gradient being denoted by:
\begin{equation}
\begin{split}
    \ell_{ce}(\tilde{\mathbf{y}},\hat{\mathbf{y}}) = & - \sum_{c=1}^C \mathds{1}_{\tilde{\mathbf{y}}(c)} \log (\hat{\mathbf{y}}(c))   \\
    \nabla \ell_{ce}(\tilde{\mathbf{y}},\hat{\mathbf{y}}) = & - \sum_{c=1}^C \mathds{1}_{\tilde{\mathbf{y}}(c)}  \frac{\partial \log(\hat{\mathbf{y}}(c))}{\partial \hat{\mathbf{y}}(c)} =  - \sum_{c=1}^C \mathds{1}_{\tilde{\mathbf{y}}(c)}  \left(\frac{1}{\hat{\mathbf{y}}(c)}\right),
\end{split}    
\end{equation}
where $\mathds{1}_{\tilde{\mathbf{y}}(c)} = 1$ if $\tilde{\mathbf{y}}(c) = \max_k \tilde{\mathbf{y}}(k)$, and $\mathds{1}_{\tilde{\mathbf{y}}(c)} = 0$, otherwise.

\textbf{3)} Similarly, we derive the gradient for the CRC loss based on equation (7) and $\mathbb{I}(\cdot)$ from equation (8) in the main paper, as shown below:
\begin{equation}
\resizebox{\linewidth}{!}{$
\begin{aligned}
\begin{split}
    & \ell_{crc}(\tilde{\mathbf{y}},\hat{\mathbf{y}}) = - \sum_{c=1}^C \Big{(}\mathbb{I}_{\tilde{\mathbf{y}} > \gamma} \mathds{1}_{\tilde{\mathbf{y}}(c)} \log (\hat{\mathbf{y}}(c)) + \mathbb{I}_{\tilde{\mathbf{y}} < \beta} (1-\mathds{1}_{\tilde{\mathbf{y}}(c)}) \log (1-\hat{\mathbf{y}}(c))\Big{)} \\
    & \hspace{-8pt} \nabla\ell_{crc}(\tilde{\mathbf{y}},\hat{\mathbf{y}}) = - \sum_{c=1}^C \frac{\partial \Big{(}\mathbb{I}_{\tilde{\mathbf{y}} > \gamma} \mathds{1}_{\tilde{\mathbf{y}}(c)} \log (\hat{\mathbf{y}}(c))\Big{)}}{\partial \hat{\mathbf{y}}(c)} + \frac{\partial\Big{(}\mathbb{I}_{\tilde{\mathbf{y}} < \beta} (1-\mathds{1}_{\tilde{\mathbf{y}}(c)}) \log (1-\hat{\mathbf{y}}(c))\Big{)}}{\partial \hat{\mathbf{y}}(c)} \\
    & \hspace{16pt} = - \sum_{c=1}^C \left ( \mathbb{I}_{\tilde{\mathbf{y}} > \gamma} \mathds{1}_{\tilde{\mathbf{y}}(c)} \left(\frac{1}{\hat{\mathbf{y}}(c)}\right)  -   \mathbb{I}_{\tilde{\mathbf{y}} < \beta} (1-\mathds{1}_{\tilde{\mathbf{y}}(c)}) \left( \frac{1}{1 - \hat{\mathbf{y}}(c)}\right) \right),
\end{split}
\end{aligned}$}
\end{equation}
with
\begin{equation}
    \begin{split}
        \mathbb{I}_{\tilde{\mathbf{y}} > \gamma} & = 
    \begin{cases}
\max(\tilde{\mathbf{y}})\hspace{0.2cm} \text{, if } \max(\tilde{\mathbf{y}}) > \gamma\\
0\hspace{0.2cm} \hspace{.85cm} \text{, otherwise }
\end{cases} \\
\mathbb{I}_{\tilde{\mathbf{y}} < \beta} & = 
    \begin{cases}
1-\min(\tilde{\mathbf{y}})\hspace{0.2cm} \text{, if } \min(\tilde{\mathbf{y}}) < \beta\\
0\hspace{0.2cm} \hspace{1.35cm} \text{, otherwise }
\end{cases},
    \end{split}
\end{equation}
and $\gamma,\beta$ representing hyper-parameters. 

To further clarify our methodology, we demonstrate the gradients calculation in Tab.~\ref{tab:analysis},
where we show three situations, with pseudo label $\tilde{\mathbf{y}}$ being unconfident, mildly confident and very confident, with the assumption that unconfident pseudo labels are more likely to be noisy and confident pseudo labels become cleaner. 
In each of these situations, we show the prediction $\hat{\mathbf{y}}$ agreeing and disagreeing with $\tilde{\mathbf{y}}$. For all these cases, we show the gradients of $\ell_{mse}$, $\ell_{ce}$, and $\ell_{crc}$, assuming that $\gamma = 0.65$ and $\beta = 0.1$.
Note that the gradient of $\ell_{mse}$ is relatively small when $\tilde{\mathbf{y}}$ is not confident, and grows slightly as $\tilde{\mathbf{y}}$ becomes more confident. Also, there is a small difference between the gradient values when $\hat{\mathbf{y}}$ and $\tilde{\mathbf{y}}$ agree or disagree. In other words, $\ell_{mse}$ is robust when the pseudo label may contain noise, but it grows too slowly when the pseudo label gets confident, which can compromise convergence.
For $\ell_{ce}$, we notice that the gradient is the same independently if $\tilde{\mathbf{y}}$ is confident or not (i.e., if the pseudo label can be noisy), but when $\hat{\mathbf{y}}$ and $\tilde{\mathbf{y}}$ disagree, then the gradient can be quite large. This suggests faster convergence, but at the risk of overfitting the noisy pseudo labels.
Our $\ell_{crc}$ does not fit samples with unconfident $\tilde{\mathbf{y}}$ (i.e., gradients have zero magnitude). However, as $\tilde{\mathbf{y}}$ becomes more confident, disagreements between $\hat{\mathbf{y}}$ and $\tilde{\mathbf{y}}$ show very large gradients. Hence, our CRC loss aims to be robust to potentially noisy pseudo labels, but also to have fast convergence when $\tilde{\mathbf{y}}$ becomes confident. 

\subsection{Analysis of the Overlap Region}

\begin{table}[t!]
\centering
\renewcommand{\arraystretch}{1.8}
\caption{\textbf{Overlap region ablation study.} We study different overlap region in BraTS19 dataset with the 25-labelled data protocol. $\widehat{\Omega}_\mathbf{f,s}$ denotes the intersection region between cropped volume $\widehat{\Omega}_\mathbf{f}$ and $\widehat{\Omega}_\mathbf{s}$. The overlap volume size is marked in \textcolor{darkred}{red} and the best results are marked in \textbf{bold}.}
\label{tab:overlap}
\resizebox{.48\textwidth}{!}{\begin{tabular}{!{\vrule width 1pt}r!{\vrule width 1pt}c|c|c|c!{\vrule width 1pt}}
\specialrule{1pt}{0pt}{0pt}
\multicolumn{1}{!{\vrule width 1pt}c!{\vrule width 1pt}}{\multirow{2}{*}{Overlap Region}}   & \multicolumn{4}{c!{\vrule width 1pt}}{BraTS19 (25-labelled data protocol)} \\
\cline{2-5}
                                  & Dice       & Jaccard      & ASD       & 95HD      \\
\specialrule{1pt}{0pt}{0pt}
 $\widehat{\Omega}_\mathbf{{f,s}} \in \frac{\mathbf{\widehat{H}} \times \mathbf{\widehat{W}} \times \widehat{\mathbf{C}}}{4\times 4\times 4}$ \textcolor{darkred}{(24x24x24)}
& 84.82      & 74.99        & 2.76      & 9.88      \\
 $\widehat{\Omega}_\mathbf{{f,s}} \in \frac{\mathbf{\widehat{H}\times \widehat{W} \times \widehat{C}}}{2\times 2\times 2}$ \textcolor{darkred}{(48x48x48)}
&85.00 &75.64 &2.84 &10.01     \\
 $\widehat{\Omega}_\mathbf{{f,s}} \in \widehat{H}\times \widehat{W} \times \widehat{C}$ \textcolor{darkred}{(96x96x96)}
&84.76 &74.93 &2.36 &9.24     \\
\hline
\rowcolor{LightCyan}
 $\widehat{\Omega}_\mathbf{{f,s}} \in [\frac{\widehat{H} \times \widehat{W}\times \widehat{C}}{2\times 2\times 2}, \widehat{H} \times \widehat{W}\times \widehat{C}\big{)}$
& \textbf{85.71}      & \textbf{76.39}        & \textbf{2.27}      & \textbf{9.20}     \\
\specialrule{1pt}{0pt}{0pt}
\end{tabular}}
\end{table}
Our selection of the sub-volume ($\hat{H} \times \hat{W} \times \hat{C}$) and the entire volume (averaging $H \times W \times C$ for each dataset) sizes are as follows:
\begin{enumerate}
    \item \textbf{LA dataset}~\cite{xiong2021global}: we follow~\cite{yu2019uncertainty} and define the sub-volume to be $112 \times 112 \times 80$. The entire volume of the dataset in average is $198.8 \times 142.4 \times 88.0$.
    \item \textbf{Pancreas-CT dataset}~\cite{clark2013cancer}: we follow~\cite{bai2023bidirectional} and define the sub-volume to be $96 \times 96 \times 96$, where the entire volume in average is  $185.3\times 166.8 \times 242.9$.
    \item \textbf{BraTS19 dataset}~\cite{menze2014multimodal}: We follow~\cite{miao2023caussl} and define the sub-volume to be $96\times 96 \times 96$. The entire volume of the dataset in average is $187.0 \times 167.39 \times 151.4$.
\end{enumerate}
In our implementation, 
to avoid the empty overlap and the entire sub-volume overlap situations, we define a random overlap region with 
$\widehat{\Omega}_\mathbf{{f,s}} \in [\frac{\widehat{H} \times \widehat{W}\times \widehat{C}}{2\times 2\times 2}, \widehat{H} \times \widehat{W}\times \widehat{C}\big{)}$ for height, weight and channel sides, respectively. 
As illustrated in Tab.~\ref{tab:overlap}, we compare the performance of the overlap region size $\widehat{\Omega}_\textbf{f,s}$ with height $\widehat{H}$, width $\widehat{W}$, and number of channels $\widehat{C}$ from $\frac{\widehat{H} \times\widehat{W} \times \widehat{C}}{4\times 4\times 4}$ to $\widehat{H} \times \widehat{W} \times \widehat{C}$ on the BraTS19 dataset.  
We observe improvements of 0.89 and 0.49 for the Dice and ASD measurements, between $\frac{1}{4}$ volume size and our final results, demonstrating the effectiveness of our proposed overlap region. Please note that utilizing the entire sub-volumes $\widehat{H} \times \widehat{W} \times \widehat{C}$ is equivalent to the Co-Training framework with an additional KL divergence loss.  \\

\section{Conclusions and Future Work}

In this paper, we presented TraCoCo, which is a new type of consistency learning SSL method that perturbs the input data views by varying the input data spatial context to reduce the dependencies between segmented objects and background patterns. Moreover, our proposed CRC loss trains one model based on the other model’s confident predictions, yielding a better convergence than other SSL losses. The comparison with the SOTA shows that our approach produces the most accurate segmentation results on LA, Pancreas and BraTS19 datasets. 
In the future, in order to deal with the the running time challenge for large volumes, more attention will be drawn to reduce the small noisy predictions while keep the sensitiveness for the foreground objects.

Self-supervised learning~\cite{caron2020unsupervised, caron2018deep} is another prominent technique explored to handle unlabelled data for medical segmentation task. Current approaches~\cite{peng2021self,zhang2021self, kiyasseh2021segmentation} rely on an additional unsupervised learning strategy to produce the latent feature patterns, which often happens before the training for the segmentation task~\cite{bai2019self, yang2023voxsep, peng2021self} or can also occur as an additional task during the segmentation training~\cite{kiyasseh2021segmentation, chaitanya2023local}. 
Most of those approaches~\cite{chaitanya2023local, yan2023localized, yang2022self} utilise contrastive learning to cluster the samples that share the same semantic meaning with the InfoNCE loss based on the pseudo label's category of the unlabelled data~\cite{chen2023semi} or the augmented version of a single instance~\cite{peng2021self, ye2022desd}. Recently, 3D-ViT~\cite{wang2023dual} has enhanced the feature representation through cross-contrastive learning between convolution networks and transformers. BHPC~\cite{tang2024semi} employs two distinct pre-training strategies for both image-wise and pixel-wise contrastive learning to enhance generalisation. Since our approach is solely based on consistency learning for semi-supervised learning, exploring self-supervised contrastive loss to learn the latent feature patterns before the semi-supervised training is an interesting topic for this project in future research.

\bibliographystyle{ieeetr}
\bibliography{refs}
\end{document}